# Accelerating science with human-aware artificial intelligence

Jamshid Sourati[1], James A. Evans[1,2]*

[1]Department of Sociology, University of Chicago; Chicago, IL, 60637, the United States.

[2]Santa Fe Institute; Santa Fe, NM, 87501, the United States.

*Corresponding author. Email: jevans@uchicago.edu

**Summary**

We explore how incorporating human awareness with artificial intelligence models can accelerate science. These simulate human inferences and can predict future discoveries, but also avoid them to generate valuable complementary hypotheses.

**Abstract**

Artificial intelligence (AI) models trained on published scientific findings have been used to invent valuable materials and targeted therapies, but they typically ignore the human scientists who continually alter the landscape of discovery. Here we show that incorporating the distribution of human expertise by training unsupervised models on simulated inferences cognitively accessible to experts dramatically improves (up to 400%) AI prediction of future discoveries beyond those focused on research content alone, especially when relevant literature is sparse. These models succeed by predicting human predictions and the scientists who will make them. By tuning human-aware AI to avoid the crowd, we can generate scientifically promising "alien" hypotheses unlikely to be imagined or pursued without intervention until the distant future, which hold promise to punctuate scientific advance beyond questions currently pursued. Accelerating human discovery or probing its blind spots, human-aware AI enables us to move toward and beyond the contemporary scientific frontier.

Research across applied science and engineering, from materials discovery to drug and vaccine development, is hampered by enormous design spaces that overwhelm researchers' ability to experimentally evaluate all candidate designs[1]. To face this challenge, researchers have initialized data-driven AI models with published scientific results to create powerful prediction engines. These models have begun to assist human discovery by focusing scientific attention on the subset of discovery candidates most predicted to possess properties relevant to energy,[2] human health[3], and other economic and societal values. In this way, AI intervenes in the discovery process by proposing efficient, model-based experiments that would require much longer for unassisted human scientists to identify. Such efforts typically ignore the distribution of scientists and inventors[4], however, the human prediction engines who continuously alter the landscape of discovery and invention. As we demonstrate below, incorporating knowledge of human researchers can dramatically improve predictions of future discoveries compared with AI methods that ignore them. Our work formalizes and demonstrates the critical importance of situated human expertise, communication and collaboration for unfolding scientific advance.

Previous studies have indicated that most new scientific discoveries emerge within neighborhoods of prior findings[5,6]. Here, we take a step further and demonstrate that the collective pattern of scientific attention is sufficient to boost the precision of future discovery forecasts. This generalizes the availability heuristic—the psychological tendency for individuals to evaluate event frequency based on cognitive availability[7]. The availability heuristic is known to result in misjudgments and decision bias[8,9]. Here we consider how and when this aggregates in scenarios involving entire scientific communities[10]. The more scientists investigate a combination of topics, the more frequently other scientists from their community will observe it presented at conferences and read about it in literature. As that combination of ideas is spoken and written about, it becomes easy to imagine and consider by nearby scientists and so conditions future scientific consideration and investigation. Here, we demonstrate that the distribution of scientists who author articles and their collaboration networks across topics and time is sufficient to foresee future discoveries and their discoverers with high precision, especially when research on the topic is sparse. This distribution, which can be recovered from publication metadata, represents a critical social fact that can stably improve our inferences about whether possible scientific relationships will soon be attempted. It can also inform our understanding of whether scientific possibilities will remain unimagined and unexplored until the more distant future[11].

We define scientific knowledge discovery as the first-time report of the relationship between an existing material and a well-defined property. An example of such pairwise relationships is "vancomycin may be used to treat pneumonia", where vancomycin is the material and effective treatment of pneumonia is the property. Our approach draws on explicit measurement of the distribution of human scientists around each topic involved in candidate discoveries, using advances in unsupervised manifold learning[12–14] and drawing upon easily available publication meta-data. By programmatically incorporating information on the evolving distribution of human experts, our approach balances exploration and exploitation in experimental search that could be used to accelerate the realization of discoveries predicted to appear in future. We contrast our human-aware approach with precise replication of a recent, prominent content-only analysis[15] that trained a Word2Vec embedding model[12] over millions of abstracts from materials science publications. That study used the resulting word vectors to infer that materials closest to electrochemical properties in the embedding space will be discovered in future to possess that property. Our models yield ~100% increase in the precision of forecasts regarding future material science discoveries. We extend this approach to identify a much broader matrix of materials and their functional properties[16], demonstrating comparable increases for predicting thousands of drugs to treat more than a hundred distinct human diseases, including vaccines and therapies for COVID-19.

Using human-aware AI, we can not only accelerate science by anticipating the human crowd; we can avoid that crowd to construct insights that punctuate human discovery with complementary hypotheses unlikely to be discovered by human scientists. If we model discovery as establishing novel links among otherwise disconnected concepts[13], it cannot occur until discoverers arise with viewpoints that bridge the fields required to imagine those conceptual connections (Fig. 1a). This diversity of scientific viewpoints was implicitly drawn upon by pioneering information scientist Swanson in his heuristic approach to knowledge generation. For example, he hypothesized that if Raynaud's disorder was linked to blood viscosity in one literature, and fish oil was known to decrease blood viscosity in another, then fish oil might lessen the symptoms of Raynaud's disorder, but would unlikely be arrived at in either field because no scientist was available to infer it[14–16]. This was one of several hypotheses later experimentally demonstrated[17–19]. Expansive opportunities for discovery persist as researchers crowd around past discoveries[20], neglecting to explore regions of knowledge cognitively distant from recent findings[21] (Extended Data Fig. 1). Our human-aware approach to complementary discovery scales and makes Swanson's heuristic continuous, identifying unstudied pairs of scientific entities likely to be scientifically and technologically relevant, but unlikely imagined. This approach avoids scientific topics at the center of collective attention and generates complementary hypotheses, which are not only unlikely to be considered by unassisted human experts, but outperform published discoveries. By staging intellectual arbitrage between isolated communities, our "alien" predictions are unconstrained by the human incentive to flock together within fields. In this way, our human-aware framework provides opportunities for accelerating the normal pathway of human discovery by predicting human accessible hypotheses, and punctuating that path by predicting human inaccessible hypotheses that complement it.

## Incorporating Human Experts with Hypergraph Proximity

We model the distribution of inferences collectively and cognitively accessible to scientists by constructing a hypergraph over research publications. A hypergraph is a generalized graph where an edge connects a set of nodes rather than a node pair. Our research hypergraph is mixed, containing nodes corresponding not only to materials and properties mentioned in title or abstract, but also the researchers who investigate them (Fig. 1c, first step). Following construction of this research hypergraph, we identify cognitively accessible inferences by generating random walk sequences over it. These walks suggest paths of inference available to active human scientists, which trace mixtures of diverse expertise sufficient for contemporary discoveries. If a valuable material property (e.g., ferroelectricity—reversible electric polarization useful in sensors) is investigated by a scientist who, in prior research, worked with lead titanate ($PbTiO_3$, a ferroelectric material), that scientist is more likely to consider whether lead titanate is ferroelectric than a scientist without the research experience. If that scientist later coauthors with another who has previously worked with sodium nitrite ($NaNO_2$, another ferroelectric material), that scientist is more likely to imagine whether sodium nitrite has the property through conversation than a scientist without the personal connection. In this way, the density of random walks over our research hypergraph is proportional to the density of cognitively plausible and conversationally attainable inferences. If two literatures share no scientists, a random walk over our hypergraph will rarely bridge them, just as a scientist will rarely consider connecting a property valued only in one community with a material understood only in a disjoint one (Fig. 1a). We hypothesize that identifying topics with high human expert density around them provides us with an informative signal regarding near-future discoveries. These topics might be located far from one another in terms of the number of steps required to travel between them in the hypergraph, but a random walker—and the collective scientific mind—can easily travel between them if intermediate steps are socially dense, facilitating conversation and collaboration (Fig. 1a).

To generate each random walk sequence, our model (i) initiates the walk with a valued property (e.g., ferroelectricity) as first node in the sequence, (ii) randomly selects an article (hyperedge) having mentioned that property, (iii) randomly selects a material or author from that article as next node (end of first step), then starts the second step by randomly selecting another article with the newly selected material or author, and repeats this Markov process[5,14] a pre-specified number of times (see Fig. 1b for an example, and Supplementary Information for more details). Each random walk step can be viewed as a simulation of human actions: an author-author step mimics networking or conversation between two expert collaborators; an author-material or author-property step represents how an author is deeply familiar with the selected material/property they have studied and published on; finally, a material/property-material/property step captures the potential for the transition to be realized by human scientists through reading a collection of scientific articles. From the collaborative character of physical and biological science, author nodes in our hypergraph far outnumber materials. To compensate for this imbalance, we devise a non-uniform sampling distribution parameterized by $\alpha$, which roughly determines the fraction of material to author nodes in resulting sequences. Specifically, we define $\alpha$ when sampling a node from a paper (e.g., in step (iii) above) such that the probability of selecting a material is $\alpha$ times that of selecting an author (See Supplementary Fig. 1). Larger values of $\alpha$ result in sampling materials/properties more frequently suggesting that our simulated researcher will uncover new scientific possibilities predominantly through research and reading; smaller values result in higher frequency of author selection implying discovery through networking, conversation, and collaboration with others in the field.

Random walks over the mixed hypergraph induce meaningful proximities between nodes. The proximity of two authors suggests they share similar research interests and experiences. The proximity of a material to a scientist assesses the likelihood she is or will become familiar with that material through research experience, related reading, or social interaction. The proximity of materials to one another suggests that they may be substitutes, complements or share another more subtle relationship such as interaction or comparison. Finally, the proximity of a material to a property suggests the likelihood that the material may possess the property, but also that a scientist will discover and publish it (Extended Data Fig. 1a-b). In this way, our hypergraph-induced proximities incorporate physical and material properties latent within literature, but also the distribution of human scientists, which enables us to anticipate inferences by those scientists and predict upcoming discoveries. The distribution of human scientists is a factor available to and naturally "read" by other competitive scientists when they attend conferences and survey their fields for promising new directions.

In order to foresee the potential discovery of materials with a valued property (e.g., store energy; cure breast cancer, vaccinate against COVID-19), we utilize random walk-induced node similarity metrics to capture the relevance between the targeted property and candidate materials. These metrics, evaluated between pairs of property/material nodes, reflect the human-inferrable relatedness of corresponding nodes and are used to sort candidate materials and report those highest ranked as inferred to possess the property. A simple metric of this kind draws upon the local hypergraph structure to estimate the transition probability that a random walker travels from the property node to a material through intermediate author nodes within a fixed number of steps, denoted by $s$. We use Bayesian rules to calculate these probabilities without the need for actually running the random-walk sampler (see Supplementary Fig. 2). Here, we only consider two- and three-step transitions ($s=2$ and $s=3$). Our main choice of metric, however, is based on a popular, unsupervised neural network-based embedding algorithm (deepwalk[13]), estimated over the random walks we generate. Like previous content-only methods[15], this method also entails construction of a word embedding model[15]. Instead of abstract sentences as input, however, the embedding is constructed over our hypergraph, considering every random walk sequence a "sentence" that links materials, experts and functional properties.

Whereas a text-based embedding captures semantic relevance among words, our approach obtains word vectors while preserving hypergraph proximities among all nodes and therefore can be used to measure the human cognitive accessibility of each material with respect to a targeted property. Because inferred discoveries involve relevant materials, we train the deepwalk embedding model after excluding authors from our random walk sequences (Fig. 1c). Cosine similarity in the resulting embedding space can be used as a relevance metric. We use these two relevance metrics, transition probabilities and deepwalk similarities, as twin criteria for selecting materials most likely to emerge as the next discoveries. Additionally, we train deeper graph convolutional neural networks, which confirms the pattern of results obtained from deepwalk (see Methods and Supplementary Information).

Note that our models do not use *more* data in comparison to traditional content-based methods, but instead alter the *type* of data we feed it. Specifically, our approach extracts and adds authorship information, but excludes the vast majority of textual content, excepting only material and property co-occurrences. In other words, our data is richer than traditional datasets in one dimension by adding human and social information, but less informative and dense in terms of content. Overall, our method possesses less data than the baselines against which we compare. In this way, our model's performance improvement, as shown below in Results, does not reflect more data, but more informative data.

## Results on Anticipating Human Discoveries

To demonstrate the power of accounting for human experts, we use transition probability and deepwalk metrics to build two alternative discovery predictors. These algorithms assess the relevance of the focal property to each candidate material based on literature published prior to a given prediction year (e.g., 2001) by embedding the human-aware hypergraph. We contrast our predictions with a random baseline and predictions generated from precisely replicated prior work that uses word embeddings based on the textual content of scientific literature without accounting for the distribution of human scientists[15]. This prior work measured property/material relevance with cosine similarity from a Word2Vec model[12] trained over the contents of scientific articles published prior to the prediction year. Our experiments and evaluation framework are identical to the settings of this study in order to facilitate precise replication. Each evaluated algorithm selects the 50 materials with highest similarity to the focal property based on hypergraph or Word2Vec similarity metrics and reports them as discovery predictions. We evaluate prediction quality based on their overlap with materials discovered and published after the prediction year (see Methods for further details; for alternative evaluation metrics and prediction sizes see Extended Data Fig. 2 and Supplementary Fig. 3).

*Energy-related Materials Prediction*

In our first set of experiments, we considered the valuable electrochemical properties of thermoelectricity, ferroelectricity and photovoltaic capacity against a pool of 100K candidate inorganic compounds. Following the evaluation regime of Tshitoyan et al. on the same dataset (1.5M scientific articles about inorganic materials)[15], we ran prediction experiments with prediction year 2001 for all three properties, predicting future discoveries as a function of research publicly available to contemporary scientists. We computed annual precisions following the prediction year until the end of 2018 (Extended Data Fig. 1c) and visualized them in a cumulative manner (Fig. 2a-c). The results indicate that predictions accounting for the distribution of human scientists outperformed baselines for all properties and materials by an average of 100%.

Sensitivity analyses with $\alpha$ reveal that a deepwalk algorithm with $\alpha=1$, which balances the likelihood of sampling materials and author nodes, had the highest and most consistent precision of prediction. Moreover, even for extremely large values of $\alpha$ (i.e., $\alpha \to \infty$), where our random walk is ignorant of human experts, the deepwalk algorithm still substantially outperforms word2vec model. We conjecture that this occurs as vague title and abstract words, irrelevant to future discoveries, add noise to the proximity of properties and materials. Our hypergraph method ignores these words, but they mislead Word2Vec resulting in weaker predictions. This suggests a more specific conjecture. Material words alone are more relevant, specific, and semantically local to other materials and properties mentioned within a paper. In this way, our hypergraph-based approach infers new discoveries in the vicinity of previous findings. Such a localized process aligns with how scientists make discoveries, leading to stronger predictions[5,6].

*Drug Repurposing Prediction*

We used the same approach to explore the repurposing of ~4K existing FDA-approved drugs to treat 100 important human diseases. We used the MEDLINE database of biomedical research publications and set the prediction year to 2001 (Extended Data Fig. 1c). Ground-truth discoveries were based on drug-disease associations established by expert curators of the Comparative Toxicogenomics Database (CTD)[17], which chronicles the capacity of chemicals to influence human health. Figure 2e reports prediction precisions 18 years after prediction year, revealing how accounting for the distribution of biomedical experts in our unsupervised hypergraph embedding yields predictions with 43% higher precision than identical models accounting for research content alone. We found a strong correlation between our human-aware prediction precision and drug occurrence frequency in literature ($r=0.74$, $p<0.001$), implying that our approach works best for diseases whose relevant drugs are frequently mentioned in prior research.

*COVID-19 Therapy and Vaccine Prediction*

We also considered therapies and vaccines to treat or prevent SARS-CoV-2 infection. Here, prediction year was set to 2020 (Extended Data Fig. 1c), when the global search for relevant drugs and vaccines began in earnest. Following Gysi et al.[18], we considered a therapy relevant to COVID-19 if it amassed evidence to merit a COVID-related clinical trial, as reported by ClinicalTrials.gov. Results shown in Fig. 2d indicate that 36% and 38% of the predictions made by transition probability and deepwalk-based metrics, respectively, were selected by biomedical experts to evaluate using expensive clinical trials within 12 months of the prediction date (i.e., end of Dec, 2020), which further increased to 42% by the end of July, 2021. This is 350 to 400% higher than the precision of discovery candidates generated by scientific content alone (10% after the first 12 months and 12% in July 2021). These precisions were even higher than a recently proposed predictive model based on an ensemble of deep and shallow learning predictors trained on multiply measured protein interactions between COVID-19 and the pool of 3,948 relevant compounds from DrugBank[18], relevant information to which our model was blind.

The success of these COVID-19 predictions suggests how fast-paced research on COVID therapies and vaccines increased the importance of scientists' prior research experiences and networks for the therapies and vaccines they would come to imagine, evaluate and champion in clinical trials. Consider the female progesterone as a candidate material. Despite very few direct literature connections between "Coronavirus" and "Progesterone" before the rise of COVID-19, random walks from our method frequently walked the path between the two literatures through pre-COVID papers published in virology, immunology, studies regarding male/female characteristics of diseases, and the female reproductive system (Fig. 3a, Extended Data Table 1). Shortly after the beginning of 2020 and in 2021, two clinical trials were initiated with similar motivation[19,20]: (1) the lower global death rate of women compared to men from COVID-19, and (2) the anti-inflammatory properties of progesterone that may moderate the immune system's overreaction to COVID-19 in men[19]. Our technique traced a pathway similar to the ones articulated explicitly by researchers sponsoring this trial: 75% of trial-cited papers, published within the five-year period preceding the prediction year we considered in building our hypergraph (2015-2019), were identified by our prediction model, and 60% of scientists authoring those studies were sampled in our random walk sequences. Progesterone and 18 other candidate materials were among the true positive predictions of our human scientist-aware method that could not be captured by the content-only baseline (Fig. 3b). By contrast, only four true positives were exclusively made by content-only prediction (Extended Data Table 2) and these four materials had significantly fewer mentions in comparison to other predicted materials, confirming that human-aware prediction performs better when candidates are mentioned frequently in prior literature.

*Human-Sensitive Prediction*

Our predictive models use the distribution of discovering experts to successfully improve discovery prediction. To demonstrate this, we consider the time required by scientists to make a discovery starting from the prediction year. Materials cognitively close to the community of researchers who study a given property receive greater attention and their relationships to that property are likely to be investigated, discovered and published earlier than those further from the community. In other words, the "wait time" for discovery should be inversely proportional to the size of the expert population aware of both property and candidate material. We measure the size of this population by defining *human expert density* between a property/material pair as the Jaccard index of two sets of human experts: those who mentioned the property and those who mentioned each candidate material in recent publications (Extended Data Fig. 3). This measures the overlap percentage between property and material research communities. For all three electrochemical properties mentioned earlier, COVID-19 therapies and vaccines, and a majority of the 100 diseases we considered above, correlations between discovery date and expert density were negative, significant and substantial (Extended Data Fig. 4). This result confirms our hypothesis that materials receiving attention from a larger crowd of property experts are discovered sooner. Our predictive models efficiently leverage the hypergraph of past publications to incorporate these human expert densities (Extended Data Fig. 5). Similar results can be derived based on embedding proximities: Fig. 4a-c illustrates how our predictions cluster atop density peaks in a joint embedding space of human experts and the materials they investigate. This further establishes that our human-aware approach is likely to select candidates more accessible to experts in the field.

We note that in some cases (e.g., photovoltaics and silicosis), discovery prediction resulted in competitive performance when $\alpha \to \infty$, with the random walker ignoring authors and traversing only material nodes. Nevertheless, the human-ignorant algorithm performs well only when mentions of the targeted property are frequent in the literature (Fig. 4d). An abundance of property-related publications, and their availability to human scientists, make the knowledge space more compact. This compactness enables scientists to infer future discoveries by simply taking in a redundant sample of papers, conference presentations, or review articles without maintaining personal connection to relevant materials, properties, or scientists. Expert awareness is critical for navigation when the knowledge is new or sparse. Even in these situations, however, the human-ignorant case $\alpha \to \infty$ performs much better in predicting discoveries than $\alpha = 0$ and other baselines, arguably because its inferences are local and in the vicinity of previous findings. This supports other evidence suggesting that scientists engage in localized search to make discoveries[6].

In addition to predicting discoveries, human-aware hypergraph proximities are also able to predict *discoverers* most likely to publish discoveries based on their unique configuration of research experiences and collaborations. Here, discoverers are defined as all article authors associated with at least one discovery, disregarding author order. In order to identify potential discoverers of materials with a specific property, we compute the probability of random-walk transition from the targeted property to author nodes through a single intermediate material across our hypergraph (without rerunning the random-walk process). Then we report potential discoverers to be those with highest transition probabilities. Our calculations here are similar to transition probabilities for discovery inferences above, except that destination nodes are authors and intermediate nodes materials (see Supplementary Fig. 2). We evaluate these discoverer predictions against scientist authors who actually published discoveries following the prediction year. Calculating average precisions across 17 prediction years (2001 to 2017) for electrochemical properties, we find that 40% of the top 50 ranked potential authors became actual discoverers of thermoelectric and ferroelectric materials one year after prediction, and 20% of the top 50 predicted authors discovered novel photovoltaics (Fig. 4e). We also employ a method with slightly more subtlety to infer the identity of those predicted to discover a relationship between a targeted property and particular material (Extended Data Fig. 6).

Discoverer prediction serves as a validation of our main algorithm's operation—by implicitly identifying the people most likely to make the discovery. Strong precision values for both our discovery and discoverer predictions imply that discoveries are predominantly performed by individuals and teams familiar with and uniquely able to bridge otherwise disconnected topics and literatures. These results can also be viewed as an initial step towards predicting individuals and teams most likely qualified to achieve specific discoveries. They suggest the potential for a scientific service that recommends potential team members for recruitment on a targeted project.

## Results on Complementing Human Discoveries

We can use our model of human cognitive availability to not only approach and mimic, but also avoid and complement the distribution of human experts. Human concept linkages are guided by previous discoveries and their discoverers (Fig. 5a). In order to build human-aware AI that propose concept linkages unlikely to be imagined by scientists, we invert a measurement of human cognitive accessibility using shortest-path distances (SPD) between pairs of conceptual nodes interlinked by authors in our mixed hypergraph. To rule out candidate hypotheses that lack scientific promise, we couple cognitive unavailability with a signal of scientific plausibility. This signal could be provided by the content of the published research literature and quantified with unsupervised knowledge embedding models[28]. Alternatively, a signal of scientific plausibility could be derived from theory-driven models of material properties. Here we use unsupervised knowledge embeddings for our algorithm, reserving theory-driven property simulations to evaluate the value and human complementarity of our predictions. Specifically, we forecast the scientific merit of any given hypothesis using the cosine similarity between embedding vectors of material and property nodes involved in that hypothesis[28].

Figure 5b provides a general overview of our algorithmic approach to identify discoveries that are both scientifically plausible and human inaccessible or complementary. Initialized with a pool of candidate materials extracted from literature, we compute human accessibility and scientific plausibility signals in an integrated fashion building on our prior analysis for generating human-like predictions. We use our unsupervised word embedding model over prior publications, measuring scientific relevance as cosine distance within the embedding. In parallel, we measure human accessibility by computing shortest-path distances between the property and all materials across the hypergraph. We transform signals of plausibility and human accessibility into a unified scale and linearly combine them with a mixing coefficient $\beta$, which captures human complementarity (see details in Methods and Supplementary Information). With its expert awareness, we designed our algorithm to symmetrically generate either the most or least-human accessible hypotheses—those likely to compete versus complement collective human capacity—based on the sign of the mixing coefficient. Negative $\beta$ values encourage high human accessibility leading to predictions that mimic human experts in discovery. Positive values discourage human accessibility by producing hypotheses least similar to those human experts could plausibly infer, straddling socially disconnected but scientifically linked fields. At extremes, $\beta = -1$ and 1 yield algorithms that generate predictions very familiar or very alien to human experts, regardless of scientific merit. On the other hand, setting $\beta = 0$ (midrange) implies exclusive emphasis on scientific plausibility, blind to the distribution of experts. This mode is equivalent to traditional discovery prediction methods exclusively based on previously published content. Intermediate positive $\beta$s balance exploitation of relevant materials with exploration of areas unlikely considered or connected by human experts. Each $\beta$ value leads to a different model assigning a scalar score per material, which we use to sort candidate hypotheses. Materials with the highest resulting scores are reported as the algorithm's predictions corresponding to that specific $\beta$.

We evaluate our expert-avoiding algorithm with the same framework as before, i.e., building our model using literature prior to a prediction year and evaluating inferred hypotheses based on subsequent actual discoveries. In this section, we expand the drug repurposing cases (properties) to include treatment of 400 human diseases. We use the prediction year of 2001 for all properties except for COVID-19, for which we set the prediction year to 2020. Complementarity of these inferences are evaluated against human scientific knowledge by verifying (1) their distinctness from contemporary investigations and (2) their scientific promise. We anticipate that both features will simultaneously increase in ranges of $\beta$ higher than those that characterize published science. Moreover, scientific merit will naturally reduce at the extremes of our interval [–1, 1], where the algorithm ignores the literature-based plausibility of candidate hypotheses. We expect to observe much higher plausibility in the intermediate ranges, which lead to strong complementarity for positive $\beta$ values.

*Evaluating Discovered Predictions*

Our human-aware model is designed to allow us to dial up and down the degree to which predictions are similar to near-future human discoveries. As we increase $\beta$, the algorithm avoids human accessible inferences that lie within regions of high expert density and focuses on candidate materials and properties that span disciplinary divides and evade human attention. As a result, we expect that generated hypotheses with large $\beta$ will (1) diverge from those pursued by the scientific community, (2) less likely become published, (3) if published, be discovered further into the future, after science has reorganized itself to consider them, and (4) manifest strong scientific performance as scientists conservatively crowd around areas of prior success. In order to verify these hypotheses, we first assess the discoverability of materials by computing the precision between our inferences and published discoveries. Results strongly confirm our expectation that materials inferred at higher $\beta$ values are less discoverable by human scientists (Extended Data Fig. 7).

Moreover, materials distant from a given property in the hypergraph are expected to remain cognitively inaccessible to scientists in the property's proximity for longer (Fig. 5c). It takes more time for researchers in the field to broach knowledge gaps separating unfamiliar materials from valued properties. Among the inferences eventually discovered, we measure the discovery waiting time and expect to observe an increasing trend in wait times as we move from negative (human-competitive) to positive (human-complementary) $\beta$ values in our predictions. Generating 50 hypotheses per $\beta$ value and evaluating the resulting predictions indicates that for the majority of targeted properties, average discovery wait times climb markedly when increasing $\beta$ (Fig. 6) for energy-related chemical properties (Fig. 6a-6c), COVID-19 prevention (Fig. 6d) and treatment for 70% of other human diseases (Fig. 6e). Averaging wait times across all human diseases manifests a clear increasing trend. For some cases, such as COVID-19 (Fig. 6d), none of the complementary predictions made with positive $\beta$ values (larger than 0.4) come to be discovered by humans within the time frame we examine.

*Evaluating Undiscovered Predictions*

To evaluate the scientific merit of our algorithm's predictions, including those that remain undiscovered within the study period, we require data beyond the extant literature. Such hypotheses necessarily grow to comprise the vast majority of cases for large values of $\beta$. If science was an efficient market and experts optimally pursued scientific quality, then in human-avoiding high $\beta$ hypotheses, we would observe a proportional decline in scientific promise and efficacy. On the other hand, if scientists crowd together along the frontier of scientific possibility and their continued efforts yield diminishing marginal returns, we might observe an increase in promise as we move beyond them.

To evaluate the merit of undiscovered scientific inferences, we utilize first principles or data-driven models derived uniquely for each property based on well-established theoretical principles within the field. Similar to our algorithms, such models also assign real-valued scores to candidate materials as a measure of their potential for possessing the targeted properties. These computations may be carried out without regard for whether materials have yet been discovered, making them a suitable scoring function for evaluating undiscovered hypotheses. We produce such scores for approximately 45% of the properties we considered above using models based on first-principles understandings of the phenomenon or models based on databases curated with high-throughput protein screens. To evaluate thermoelectric promise, we used power factor (PF) as an important component of the overall thermoelectric figure of merit, $zT$, calculated using density functional theory for candidate materials as a strong indication of thermoelectricity[29,30]. To evaluate ferroelectricity, estimates of spontaneous polarization obtained through symmetry analysis and relevant theoretical equations serve as a reliable metric[31]. For human diseases including COVID-19, proximity between disease agents (e.g., SARS-CoV-2) and candidate compounds in protein-protein interaction networks suggests the likelihood a material will recognize and engage with the disease agent[32] (for more details on how these theoretical scores are derived see the Supplementary Information). We note that scores based on first-principles equations or simulations represent conservative estimates of scientific merit as they are based on widely-accepted, scientist-crafted and theory-inspired models. Because these scores are potentially available to scientists in the area, they may be considered when guiding investigations such that experiments on these unevaluated hypotheses often lead to promising results. Nevertheless, in what follows we show that modestly positive $\beta$ values manifest an marked improvement even on this conservative measure of quality.

We expect the average theoretical scores of hypotheses to decay significantly at the extremes of the $\beta$ range [–1,1], as at those points the algorithm ignores the merit signal putting it at higher risk of generating scientifically irrelevant (or absurd) proposals. We expect, however, that this decay will occur more slowly than the decrease in hypothesis discovery and publication, which implies the existence of a $\beta$ interval where proposals are not discoverable but highly promising—an ideal operating region for the generation of hypotheses that complement those from the human scientific crowd. In order to verify this, we contrasted changes in average theoretical scores with the discoverability of generated hypotheses for various $\beta$ values. As illustrated in Fig. 7 (first row), discoverability decreases near the transition of $\beta$ from negative to positive values, but its decay is much sharper than average theoretical scores, which do not collapse until nearly $\beta = 0.4$. This holds for electrochemical properties and the majority of diseases. Results for certain individual diseases can be seen in the second row of Fig. 7 (for the full set of results see Extended Data Fig. 8 and Supplementary Table 1). Moreover, note that for the cases investigated, average theoretical scores for inferred hypotheses grow higher than average theoretical scores for actual, published discoveries (the dashed lines) before their eventual decay at high $\beta$ values. For certain properties like thermoelectricity or therapeutic efficacy against the disease Alopecia, theoretical merit of our inferences exhibit striking and dramatic growth from negative (scientist-mimicking) to positive (scientist-avoiding) hypotheses, suggesting strong diminishing returns to following these scientific crowds, whose overharvested fields have become barren for new discovery.

In order to further compare the decay rate of discoverability and theoretical scores, we define and compute an *expectation gap* to measure the distance between expected values for two conditional distributions over $\beta$. These two conditionals are defined as two likelihoods over $\beta$ given that a randomly selected prediction with that $\beta$ is (1) identified as promising based on its corresponding first-principle score, and (2) discoverable, i.e., studied and published by a scientist following prediction year (for details see Methods and Supplementary Information). A positive expectation gap indicates that increasing $\beta$ will preserve the quality of predictions while making them more complementary to human hypotheses. As shown in Fig. 8a, the vast majority of properties considered in this section yield substantial and significantly positive expectation gaps. Building on this, we use a probabilistic model to assess the complementarity of our algorithm's prediction with those of the scientific community for any value of $\beta$. This is done by explicitly computing the joint probability that a randomly selected prediction is plausible in terms of the desired property and beyond current scientists' scope of research (see Supplementary Information). These probabilities specify the optimal $\beta$ to balance exploitation and exploration in augmenting collective human prediction. Results in Fig. 8b indicates the optimal point varies for different properties, but one can distinguish the range 0.2-0.3 as the most consistently promising interval. In this interval, hypotheses are very unlikely to come from the scientific community, but are very likely to yield successful scientific results.

## Discussion

We demonstrate the power of incorporating human-awareness into artificial intelligence systems for accelerating future discovery. Our models succeed by directly predicting human discoveries and the human experts who will make them, yielding up to 400% improvement in prediction precision. These findings offer support for the influence of the human experience and social connection inscribed by our research hypergraph in driving scientific advance. This suggests that the search underlying materials and medical advance is dominated by local exploitation of the familiar over novel exploration of the unknown. Moreover, by tuning our algorithm to avoid the crowd, we generate promising hypotheses unlikely to be imagined, pursued or published without machine recommendation for years into the future. By identifying and correcting for collective patterns of human attention, formed by field boundaries and institutionalized education, these models complement the contemporary scientific community. This demonstrates that connectivities in our expert-aware hypergraph are useful not only for predicting and accelerating human discoveries in the near future, but also for inferring disruptive discoveries that could be imagined by scientists only in distant future.

Our analysis examined a limited space of scientific relationships–those between a material possessing a valuable energy or therapeutic property. Many other scientifically meaningful relationships lie beyond this syntax, such as identity (i.e., *a* is a *b*), composition (i.e., *a* is a part of *b*), or any specific physical or logical relationship (e.g., *a* chemically reacts with *b*; *a* genetically up-regulates *b*). Using a hypergraph formalism, we could extend such relations beyond logical triples that connect a simple concept pair to larger sets of concepts connected by more complex relations. Another limitation involved our singular consideration of co-authorship as the relationship affecting the distribution of expertise. One could consider other relationships, such as scientist collocation within an institution, at a conference they attend, or geographical proximity. Moreover, there are opportunities to technically improve our approach, such as combining content and human-aware information to amplify prediction accuracy, or inferring and exploiting the body of negative knowledge in science where researchers know that certain scientific claims are false[11,21].

Despite these limitations, our investigation underscores the power of incorporating human and social factors to produce artificial intelligence that complements rather than substitutes for human expertise. Successful scientists competitively factor and follow the momentum of advances made by researchers around them in identifying the frontiers of science. By making AI hypothesis generation aware of human expertise, it can accelerate discovery and liberate human scientists to steer science and technology in novel directions. Our system and its recommendations raise ethical concerns; they could be used as a "scoop-machine" to leapfrog human scientists and seize upon answers that they might otherwise ask and answer next. This would accelerate science, but could augment some scientists' capacity at the expense of others. Such a concern would attenuate when scientific recommendation engines became ubiquitous, however, like recommendations for internet and social media search. Moreover, we demonstrate how awareness of human scientific expertise could be used not only to mimic but avoid it, generating insights that punctuate the current flow of discovery[22].

Our investigation also reveals the influence of human scientific institutions that crowd scientists along a shared frontier of likely discoveries. The success of our 'alien' or complementary hypotheses suggests that scientific departments and disciplines limit productive exploration and point to opportunities that could improve human prediction by reformulating science education for discovery. Insofar as research experiences and relationships condition the questions scientists investigate, education tuned to discovery might conceive of each student as a new experiment, recombining knowledge and opportunity in novel ways. Our analysis underscores the power of incorporating human and social factors to produce artificial intelligence that complements rather than substitutes for human expertise. In accounting for not just human expertise, but the complete distribution of scientific experience and exposure, such systems can be designed to race with rather than against the scientific community, expanding the scope of human imagination and discovery.

## Materials and Methods

*Experiments and Data Collection*

Each discovery prediction experiment consists of a target property and a pool of materials, where the materials are scored by a predictor and the 50 materials with highest scores are selected as predictions. Each predictor scores an individual material through computing its similarity with the property. Similarity metrics for our hypergraph-based predictors are the transition probability between material and property nodes with one and two intermediate author nodes (hence two- and three-step transitions, i.e. $s=2$ and $s=3$), and cosine similarity in the deepwalk embedding space. The former can be calculated through Bayes' rule without the need for generating random walks, but the latter require an explicit set of random walk sequences over our hypergraph. The similarity metric from the replicated content-only baseline is the cosine similarity in the embedding space of a Word2Vec model trained on the corpus of publications produced before prediction year. The corpus of publications and ground-truth discoveries are prepared differently for each set of property and potential materials.

Our testbed consisted of two datasets: for the energy-related properties we used a collection of ~1.5M articles published between 1937 and 2018 classified by Tshitoyan et. al (2019) as related to inorganic materials[15], and for the diseases we utilized MEDLINE database that includes more than 28M articles published in various biomedical fields over the span of more than two centuries. Creating our hypergraph required us to have access to disambiguated authors for all articles. We downloaded the database related to inorganic materials using Scopus API provided by Elsevier (https://dev.elsevier.com/), which readily assigns unique codes to distinct authors. In order to author-disambiguate the MEDLINE database, we used disambiguation results provided by the PubMed Knowledge Graph (PKG)[24], which were obtained by combining information from the Author-ity disambiguation of PubMed[25] and the more recent semantic scholar database[26]. This integrative method has a performance comparable to each of its individual components: 98.09% F1-score, 98.62% precision and 97.56% recall. For this dataset, we restricted our experiments to 27.5M papers with available abstracts, metadata (publication year) and disambiguated authors.

For energy-related properties, we extracted the pool of chemical compounds from the collected 1.5M articles using Python Materials Genomics[27] and direct rule-based string processing. Material-property association was defined in terms of co-occurrence of materials with property-related keywords. First-time co-occurrences were considered ground-truth discoveries, following the replicated prior work[15]. For the case of drug repurposing, we began with a pool of 7,800 approved candidate drugs downloaded from the DrugBank database. We then built our drug pool using approximately 4,000 drugs possessing simple names (i.e., dropping complex names containing several numerical components). We chose 100 (or 400, when avoiding experts) diseases from the Comparative Toxicogenomics Database (CTD)[17] with the largest number of relevant drugs from our drug pool. In order to build our hypergraph, we searched for names of drugs and diseases in MEDLINE to detect their occurrence within papers. Ground-truth relevant drugs for the selected diseases (except COVID-19) were extracted from associations curated by CTD. The discovery date for each of the disease-drug associations was set to the earliest publication reported by CTD for curated relevance. We ran separate prediction experiments for each individual disease, where we define the property as drug efficacy in treating or preventing the selected disease. The same pool of drugs and corpus of articles were used for the case of COVID-19, where the ground-truth relevance of drugs to COVID-19 were identified based on their involvement in COVID-related studies reported by ClinicalTrials.org in or after 2020, regardless of the studies' results, following the compared work by Gysi et al.[18]. Date of discovery for each relevance was set to the date the corresponding study was first posted, and if the drug was involved in multiple trials we considered the earliest. There have been 6,280 trials posted as of August 5th, 2021 (ignoring 37 trials dated before 2020), which included 279 drugs from our pool (~7%) included in their study designs.

*Hypergraph Random Walks*

In practice, research and coauthoring that occurred long before the time of prediction are unlikely to be cognitively available, socially accessible, or perceived as of continuing relevance. Therefore, we restrict our prediction experiments to use literature produced in the most recent 5 years prior to the year of prediction. For alternative time windows, the magnitude of precision curves slightly shifted, but their trend and ordering remained the same (see Supplementary Fig. 4). For each property, we took 250,000 non-lazy, truncated random walks with and without $\alpha$-modified sampling distribution sequences. All walks start from the property node and end either after 20 steps or after reaching a dead-end node with no further connections. The $\alpha$-modified sampling algorithm is implemented as a mixture of two uniform distributions over authors and materials such that the mixing coefficient assigned to the latter is $\alpha$ times the coefficient of the former. Hence, $\alpha$ is the ratio of probability for selecting a material to the probability of selecting an author node (see Supplementary Information for more details). We tried three values for this parameter in our experiments: $\alpha = 1$, which implies an equal probability of sampling authors and materials, $\alpha \to \infty$ which only samples materials and $\alpha = 0$ which only samples authors. The author-only mode yielded much weaker performance in comparison to the other two and we do not include it in our results. This implies that mere networking with other human experts, without reading and researching the literature does not typically lead to discoveries in practice. A further perturbation analysis of $\alpha$ showed that increasing it to values larger than one (e.g., 10) is less harmful to precision levels than decreasing it below one (e.g., 0.5). In other words, the balance point leads to highest performance (i.e., $\alpha=1$), but if one break balance between researching (e.g., Googling and reading related research papers) and networking with nearby scientists, overemphasizing research exploration harms prediction less than overemphasizing social networking in predictions of knowledge discovery (see Supplementary Fig. 5 for a more thorough sensitivity analysis of our algorithm with regard to parameter $\alpha$).

*Relevance Metrics*

Once the random walk sequences are drawn, we can compute our two hypergraph-induced similarities. Multi-step transition probabilities are directly computed from transition matrices using Bayesian rules and Markovian assumptions (see Supplementary Information). Calculating probabilities for two- and three-step transitions from properties to materials requires us to sum the probability of all meta-paths with the form PAM and PAAM, where P, A and M stand for property, author and material nodes, respectively[28]. Alternatively, the meta-path that we considered for discover*er* prediction was PMA. For our deepwalk representation, we trained a skipgram Word2Vec model with hyperparameter settings similar to the content-only prior work we replicated[15], including an embedding dimensionality set to 200. One exception is the number of epochs, which we reduced from 30 to 5. The size of vocabulary produced by deepwalk sampling is substantially smaller than the number of distinct words from literature. As a result, deepwalk training required less effort and lower iterations to capture the underlying inter-node relationships. Also note that deepwalk embedding similarity is more global than the transition probability metric, provided that the length of our walks (set to 20) are longer than the number of transition steps (2 or 3). Moreover, it is more flexible since the walker's edge selection probability distribution can be easily modified to explore the network structure more deeply[29]. Nevertheless, because the deepwalk Word2Vec is trained using a window of only length 8, only authors and materials that might find each other through conversation, seminars or conferences would be near one another in the resulting vector space.

We also ran our prediction experiments after replacing the deepwalk representation with a graph convolutional neural network. We used the Graph Sample and Aggregate (GraphSAGE) model[30] with 400 and 200 as the dimensionality of hidden and output layers, respectively, with Rectified Linear Units (ReLU) as non-linear activations in the network. Convolutional models require feature vectors for all nodes but our hypergraph is inherently feature-less. Therefore, we utilized the word embeddings obtained by our Word2Vec baseline as feature vectors for materials and property nodes. A graph auto-encoder was then built using the GraphSAGE architecture as the encoder and an inner-product decoder and its parameters were tuned by minimizing the unsupervised link-prediction loss function[31]. We took the output of the encoder as the embedded vectors and selected the top 50 discovery candidates by choosing entities with the highest cosine similarities to the desired property. In order to evaluate the importance of the distribution of experts for our prediction power, we trained this model on our full hypergraph and also after withdrawing author nodes (see Supplementary Information). Running the convolutional model on energy-related materials and properties yielded 62%, 58% and 74% precisions on the full graph, and 48%, 50% and 58% on the author-less graph for thermoelectricity, ferroelectricity and photovoltaics, respectively. These results show a pattern similar to those obtained through deepwalk model although with somewhat smaller margin due to the use of Word2Vec-based feature vectors, which limited the domain of exploration by the resulting embedding model to within proximity of the baseline.

*Complementary Hypotheses Generation*

Our predictor consists of two scoring functions. The first measures the human inaccessibility (i.e., alienness) of candidate materials via Shortest-Path distance (SPD) between the nodes corresponding to the targeted property and candidates. The second measures scientific plausibility through the semantic cosine similarities of their corresponding keywords. For this purpose, we use our Word2Vec baseline pretrained over the literature (collected on inorganic materials for energy-related properties and MEDLINE for the diseases) produced prior to the prediction year. We combine the alienness and plausibility scores with a mixing coefficient, denoted by $\beta$, adjusting their contributions to obtain a final score for the candidate. The plausibility component yields continuous scores distributed close to Gaussian, whereas the alienness component offers unbounded ordinal SPD values. Simple normalization methods are insufficient to combine scores with such distinct characteristics. As a result, we first standardize the two scores to a unified scale by applying the Van der Waerden transformation[41], followed by a Z-score normalization. The final step includes taking the weighted average of the resulting Z-scores with weights depending on $\beta$ (see Supplementary Information for more details).

We want our predictor to infer undiscoverable yet promising hypotheses. Setting $\beta$ to a more positive value makes predictions less familiar and more alien, i.e., less discoverable. Moreover, increasing $\beta$ to the positive extreme (i.e., +1) excludes scientific merit from the algorithm's objective in materials selection. Hence, growing $\beta$ causes both discoverability and plausibility of predictions to decay. What matters to us is that plausibility decreases more slowly than discoverability, suggesting that the predictor achieves a close-to-ideal state where predictions are simultaneously alien and promising. In order to verify this with a single number, we define the *expectation gap* criterion, computed as the difference between expected values of the following two distributions over $\beta$: $\mathbb{P}(\beta|\text{plausible})$ and $\mathbb{P}(\beta|\text{discoverable})$. The terms "plausible" and "discoverable" on the conditional sides could be substituted by the precise statements "a randomly selected inferred hypothesis is theoretically plausible" and "a randomly selected inferred hypothesis is discoverable"—it will be published by scientists, respectively. While we know both of these distributions reduce as $\beta$ approaches +1, the expectation gap measures any positive shift in the mass of $\mathbb{P}(\beta|\text{plausible})$ against $\mathbb{P}(\beta|\text{discoverable})$. The likelihood of discovery $\mathbb{P}(\beta|\text{discoverable})$ can be

estimated through an empirical distribution of predictions discovered and published. Scientific plausibility can be estimated by leveraging properties' theoretical scores obtained from prior knowledge and first-principles equations and data from relevant fields. We estimate $\mathbb{P}(\beta=\beta_0|$ plausible) in two steps: (1) converting theoretical scores to probabilities, and (2) computing weighted maximum likelihood estimates of $\mathbb{P}(\beta=\beta_0|$plausible) given a set of predictions generated by our algorithm operated with $\beta_0$ (see Supplementary Information for details). We restrict experiments in this section to only those properties for which we could obtain a reliable source of theoretical scores (see Supplementary Information for details of the scores): thermoelectricity, ferroelectricity, COVID-19 and 175 other human diseases (178 out of 404 total properties).

Finally, note that expectation gaps and average discovery dates (described above) say nothing about the $\beta$ interval most likely to lead to better complementarity. We introduce an additional probabilistic criterion for this purpose, which explicitly and jointly models these two features and computes their likelihood for various $\beta$ values, $\mathbb{P}(\text{undiscoverable, plausible} \mid \beta)$. One can use this distribution to screen the best operating point for complementary artificial intelligence (see Supplementary Information).

### *Data Availability*

DOIs of papers used for the electrochemical properties together with the PubMed identifiers of the MEDLINE entries used in our experiments can be found in our GitHub repository: https://github.com/jsourati/accelerate-discoveries.

Abstracts of papers for electrochemical properties could not be shared due to copyright issues. But MEDLINE abstracts are accessible through their identifiers from the PubMed website.

### *Code Availability*

All codes for our algorithms can be found in the following GitHub repository: https://github.com/jsourati/accelerate-discoveries.

### *Competing Interests*

The authors declare no competing interests.

### *Author Contributions*

JS: Conceptualization, Methodology, Software, Validation, Investigation, Writing - Original Draft, Visualization; JE: Conceptualization, Methodology, Writing - Original Draft, Visualization, Funding acquisition.

**Acknowledgments**

The authors wish to thank our funders for their generous support: National Science Foundation #1829366; Air Force Office of Scientific Research #FA9550-19-1-0354, #FA9550-15-1-0162; DARPA #HR00111820006. The funders had no role in study design, data collection and analysis, decision to publish or preparation of the manuscript. We thank Laszlo Barabasi and Deisy Morselli Gysi for helpful data related to their network-based forecast of COVID-19 drugs and vaccines with protein-protein interactions [18], and Anubhav Jain, Vahe Tshitoyan and Alex Dunn for sharing data and code to help replicate their work on unsupervised word embeddings and latent knowledge about material science [15]. We also thank participants of the Santa Fe Institute workshop "Foundations of Intelligence in Natural and Artificial Systems", the University of Wisconsin at Madison's HAMLET workshop, and colleagues at the Knowledge Lab for helpful comments.

**Fig. 1. Motivation and design of our approach to simulate human accessible scientific inferences** (**a**) Three scenarios where a hidden underlying relationship between material M and property P waits to be discovered. Uncolored circles represent non-overlapping populations of human experts, with colored nodes indicating materials (blue) or properties (colored). Background colors represent overlapping disciplinary communities, within which scientists and topics are embedded. Solid lines between uncolored and colored nodes imply that experts studied or have experience with the material or property. Dashed lines represent property-material links that exist but have not yet been discovered by human scientists, and gray arrows represent new hypotheses proposed by our algorithm. The P-M relation in the upper left scenario is likely to be discovered and published in the near future and is proposed by our algorithm; the P-M relation in the upper right is likely to escape scientists' attention, and also the notice of our algorithm that simulates human accessible hypotheses. Nevertheless, our algorithm also captures transitive inference as scientists do through research and conversation over time; let P–$M_1$–$M_2$–$M_3$ be a chain of materials connected to property P, and every consecutive pair $M_i$–$M_{i+1}$ are strongly connected either because they are already shown to be connected in published articles or because there is a group of researchers familiar with both, having studied both across their opus of research. Our algorithm walks over consecutive pairs and infers the existence of P–$M_3$ relationship and its likelihood of discovery in future. (**b**) Four examples of random walk paths starting from "Coronavirus" (property) and ending at "Progesterone" (a chemical under clinical trial investigation for COVID-19 therapeutic efficacy). Each arrow connecting two nodes indicates a sampling step, where the paper shown on top of the receiving node comprises the selected hyperedge for that step, which by construction contains both nodes sampled in the prior and current steps. (**c**) Illustration of our hypergraph deepwalk algorithm: 1) We construct a hypothetical hypergraph based on literature represented by three papers. Uncolored shapes represent authors and colored shapes indicate properties (red) or materials (blue) mentioned in article titles or abstracts. 2) We perform classic or α-modified random walk sampling , which 3) result in a set of sequences consisting of authors, materials and the focal property. 4) We remove authors from sequences, retaining only the materials on which discovery inference will be applied. 5) We train a word embedding model (e.g., Word2Vec) on these sampled human accessible sequences of material/property tokens, which results in 6) a vector representation of property and materials we use to compute similarities for prediction.

**Fig. 2. Evaluating human-accessible discovery predictions against various baselines.** Precision rates for human accessible discovery predictions regarding materials associated with different properties and prediction years: (**a-c**) chemical compounds and electrochemical properties including thermoelectricity, ferroelectricity, and photovoltaic capacity, respectively, with prediction year 2001; (**d**) therapeutics and vaccines for COVID-19 in prediction year 2020; (**e**) general disease-drug associations for prediction year 2001. Precisions reported for general disease-drug associations are individual rates computed 19 years after prediction year, but computed annually for electrochemical properties and monthly for COVID-19 efficacy (See Extended Data Fig. 1c). Gray bars in Figs. a-d indicate the number of actual new discoveries each month or year of the prediction period. The curve labeled "theoretical" in the case of COVID-10 represents predictions generated based on protein-protein interaction network by Gysi et al.[18]. Predictions accounting for the distribution of human experts are far superior to those that ignore it.

**Fig. 3. A prediction example of progesterone as COVID-19 therapy.** (**a**) An example random walk from the property node "Coronavirus" to the material node "Progesterone", where selected hyperedges (papers) are shown in detail. Every article in this path is a hyperedge (denoted $e_i$ in the *i*-th step) connecting the prior to the subsequent node. The last article was cited by the University of Southern California clinical trial that investigated the effectiveness of progesterone for COVID-19 treatment. Relevant MeSH Terms from articles are shown to demonstrate their scope, indicating hints regarding the reasoning of human scientists championing the treatment. The path indicates a clear transition from Coronavirus-related topics to male-female differences in pathological conditions and lastly to progesterone-based therapy. Similar bridges between topics were highlighted by the trial's investigator as the main motivation for her study in a published news interview[23]. (**b**) List of true positive discovery predictions made by our human accessible deepwalk algorithm, which were misclassified by the content-only predictor. Edge colors represent the ratio of $rank_{word2vec} / rank_{deepwalk}$, where the numerator denotes the rank of the material in terms of our deepwalk scoring function that simulates the inferences made by human experts, and the denominator indicates the rank based on Word2Vec's scoring function that consider research contents alone. Because we display only true positives, expert-deepwalk rank$_{deepwalk}$ ≤ 50 and rank$_{word2vec}$ > 50 for all shown materials. A higher rank ratio reveals a larger disparity in the accuracy of algorithmic assessments. The largest ratio is associated with Ethanol inhalation (rank$_{deepwalk}$ = 15, rank$_{word2vec}$ = 2,762), widely used in treating pulmonary edema, and the smallest to Sofosbuvir (rank$_{deepwalk}$ = 38, rank$_{word2vec}$ = 102), an antiviral used to treat hepatitis C.

**Fig. 4. The contribution of human expert awareness for predicting discoveries and discoverers.** (**a-c**) 2D projections of human accessible material predictions made by deepwalk (blue circles) and the content-exclusive Word2Vec model (red circles) for thermoelectricity (left), ferroelectricity (center) and photovoltaic capacity (right). Circles with center dots indicate true positive predictions discovered and published in subsequent years, while empty circles represent false positives. Predictions are plotted atop the density of experts (topological map and contours estimated by Kernel Density Estimation) in a 2D tSNE-projected embedding space. Before applying tSNE dimensionality reduction, the original embedding was obtained by training a Word2Vec model over random walks generated across the hypergraph of published science (similar to our deepwalk procedure shown in Fig. 1 but without removing authors). Red circles are more uniformly distributed, but blue circles concentrate near peaks of expert density. (**d**) Precision shifts in predictions attributable to the inclusion of authors, defined as the percentage of precision change when switching from $\alpha \to \infty$ to $\alpha = 1$, plotted against the fraction of property-related papers within the literature. Higher density in the literature obviates the need for human author information. (**e**) Precision rates for predicting *discoverers* of materials with electrochemical properties. Predictive models are built based on two-step transitions between property and expert nodes with an intermediate material in the transition path. Bars show average precision of human expert predictions for each year following prediction. Note that an expert can publish a discovery in multiple years. Total precision rates are also shown after each property, ignoring the repetition of discovering experts.

**Fig. 5. Motivation and design of our approach to generate complementary scientific predictions by avoiding human scientists.** (**a**) Distribution and overlap of experts investigating (and publishing on) topics represented by yellow geometric shapes. Dashed lines represent paths of more or less human cognitive availability between topics ("triangle", "diamond" and "square"). (**b**) Overview of our complementary discovery prediction algorithm. Beginning with a scientific corpus and a targeted property, candidate materials are extracted from the corpus and used along with property mentions and authors to form the hypergraph. The algorithm follows two branches to compute plausibility from word embedding semantic similarities and human inaccessibility or "alienness" from hypergraph shortest-path distances. These two signals are combined after proper normalization and standardization through the mixing coefficient $\beta$ to generate a prediction more or less complementary to the flow of human discovery (higher $\beta$s, more human inaccessible and so more complementary; lower $\beta$s, more human accessible and so more in competition). Candidate materials are sorted based on resulting scores and those with highest rank are reported as proposed discoveries. (**c**) Discovery wait times for relations between "triangle"–"diamond" and "triangle"–"square". The time one needs to wait for a relationship to be discovered is proportional to the path length of human accessibility between the two relevant topics. The denser presence of human experts around the pair "triangle"–"diamond" implies greater cognitive availability leading to earlier discovery and publication versus "triangle"–"square" where the connection requires a longer path.

**Fig. 6. The wait time for published discoveries increases with human inaccessibility (higher $\beta$ values).** (**a-d**) Average annual/monthly discovery wait times are shown as thick gray arcs, where thickness represents the percentage of materials discovered in the corresponding year/month. Each orbit is associated with a particular $\beta$ value with larger (more red) orbits representing larger $\beta$ values and greater human inaccessibility as computed by our algorithm's human expert avoidance. The values we consider here vary between -0.8 (the smallest, bluest orbit) and 0.8 (the largest, reddest orbit). The plot in the upper right quarter of the orbits reveals the total average of discovery wait times including all years/months for each considered $\beta$ value. (**e**) Average wait times for discoveries across all human diseases (except COVID-19) from our experiments.

**Fig. 7. Precision in predicting human discovery falls before a comparable drop in theoretical expectations. (a-b)** Green bars show the precision of complementary predictions with human published discoveries, while curves indicate theoretical expectations of first-principle simulations, which include **(a)** average Power Factor (PF) for thermoelectricity and **(b)** spontaneous polarization for ferroelectricity. **(c)** Precisions of complementary predictions for human discovery and average theoretical scores (i.e., protein-protein interaction similarity scores) for therapeutic predictions. Horizontal dashed lines in all cases show average theoretical scores computed for actual discoveries following the prediction year. **(d)** Published discovery prediction versus average protein-protein similarity scores for nine human diseases. The *y*-axis indicates precision at predicting discovery, while the color gradient represents average theoretical scores for predictions. In all cases, prediction of human discoveries fall much faster than theoretical expectations, which themselves are accessible to human experts and so represent a conservative estimate of scientific plausibility.

**Fig. 8. Complementary AI predictions outperform human discoveries. (a)** We formalize and estimate the expectation gap for properties with first-principle or data-driven theoretical plausibility scores. We plot the conditional distributions $\mathbb{P}(\beta|\text{plausible})$ and $\mathbb{P}(\beta|\text{discoverable})$ separately for materials with valuable energy-related properties of thermoelectricity and ferroelectricity, and for therapeutics fighting COVID-19 and hundreds of other human diseases (shown collectively in a normalized histogram). The first row demonstrates that the quality of our complementary hypotheses improves or maintains beyond those materials accessible to human scientists, as discovered and later published by them. The second row shows the individual gaps between discoverability by human scientists and plausible performance based on theoretical and data-driven models for a subset of human diseases. **(b)** The joint probability of simultaneous undiscoverability and plausibility for different values of the alien parameter *β*, where low values (*β*=-1) indicate mimicry of human discovery, zero values (*β*=0) its ignorance of human discovery, attending only to the research literature, and positive values (*β*=1) its avoidance of human discoverability.

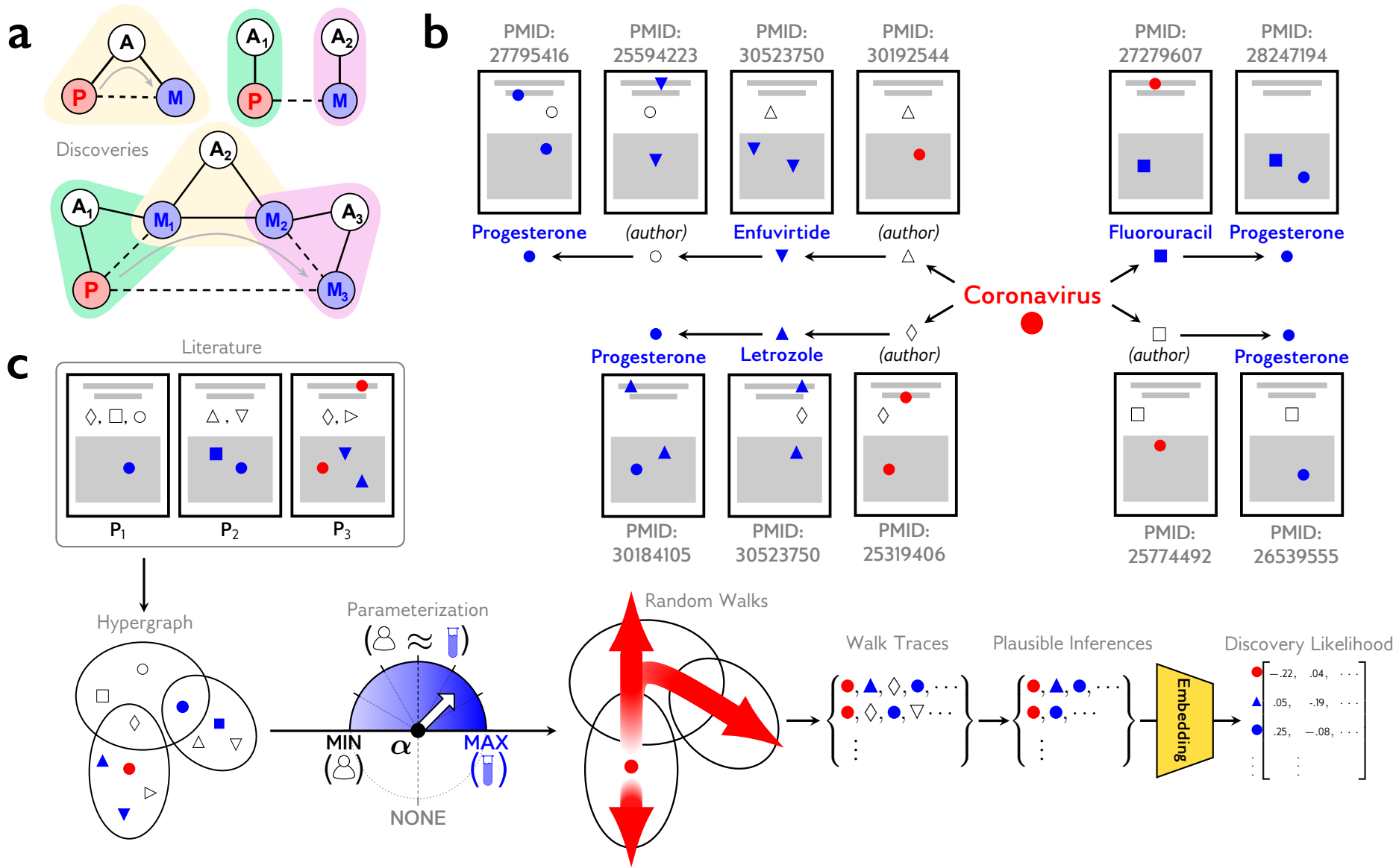

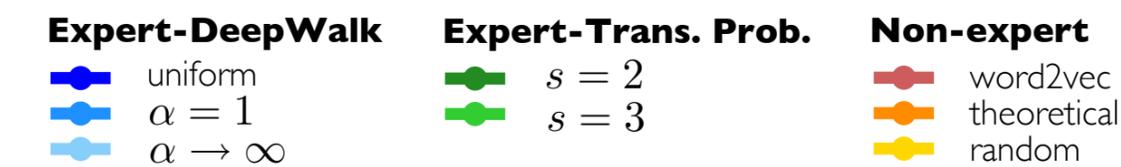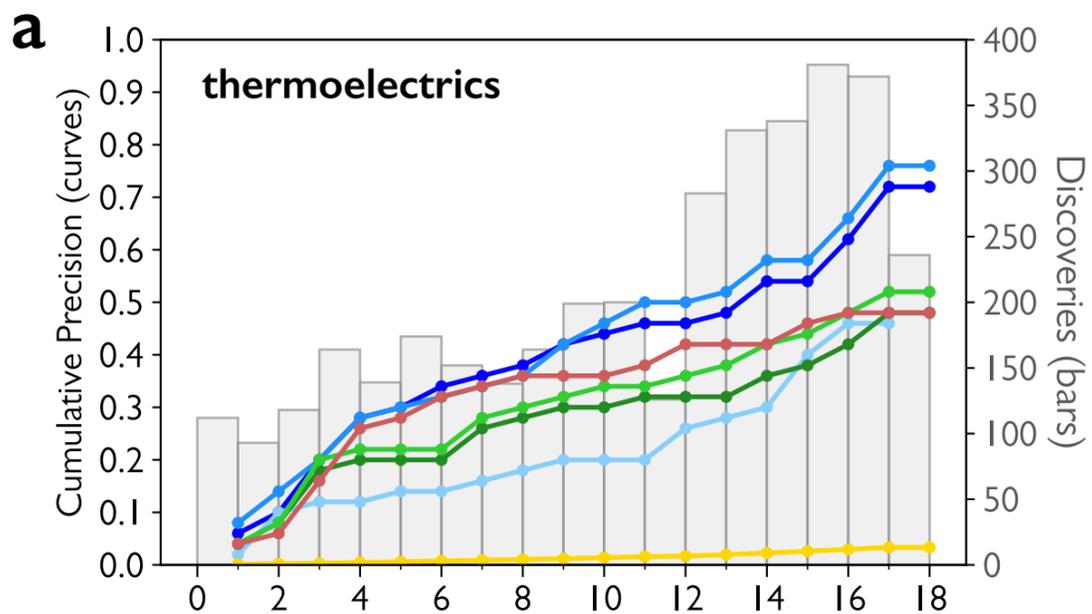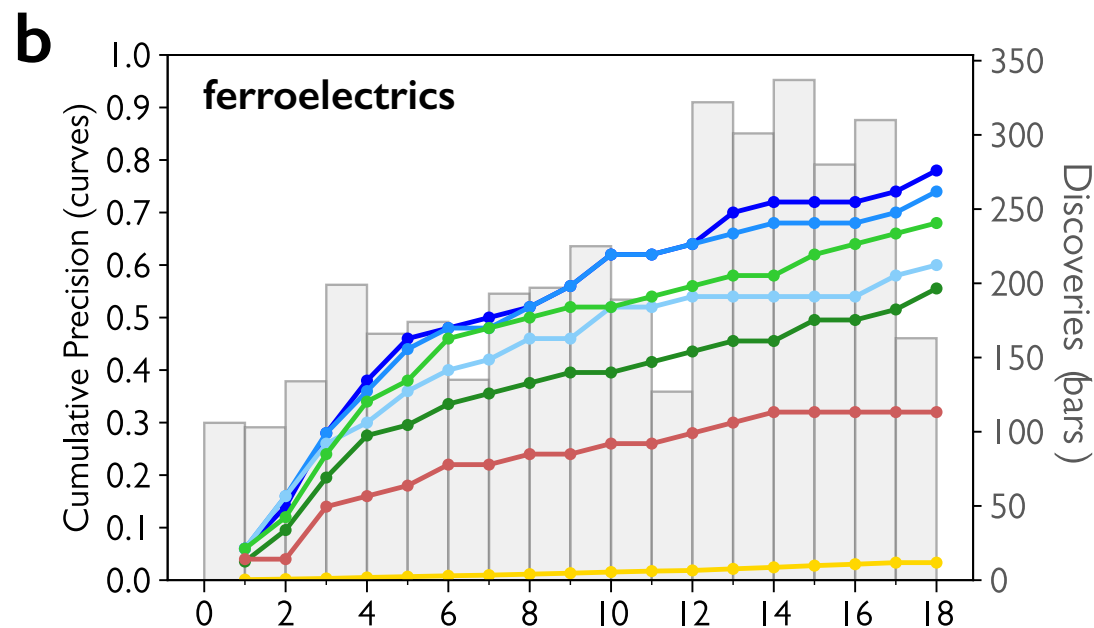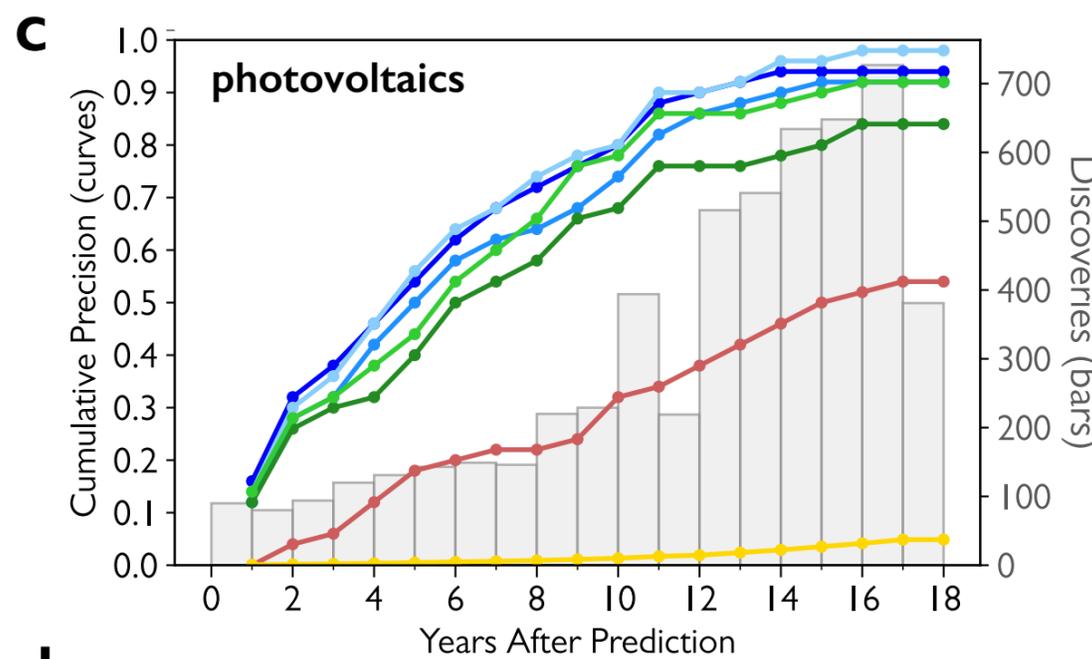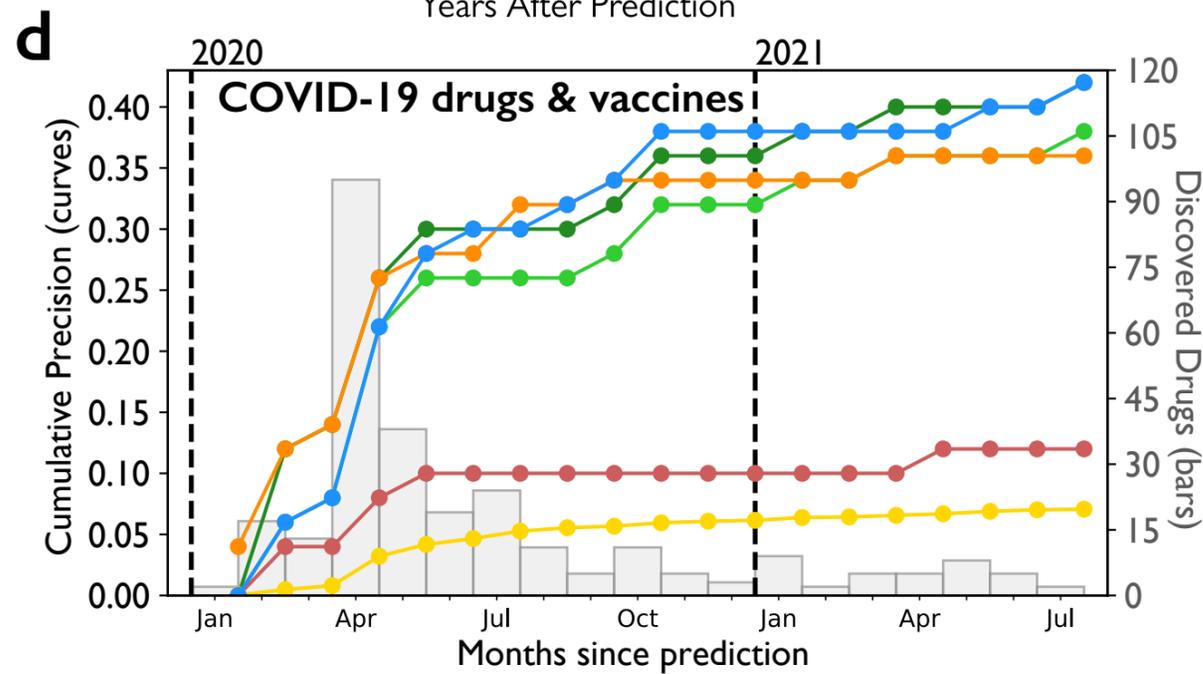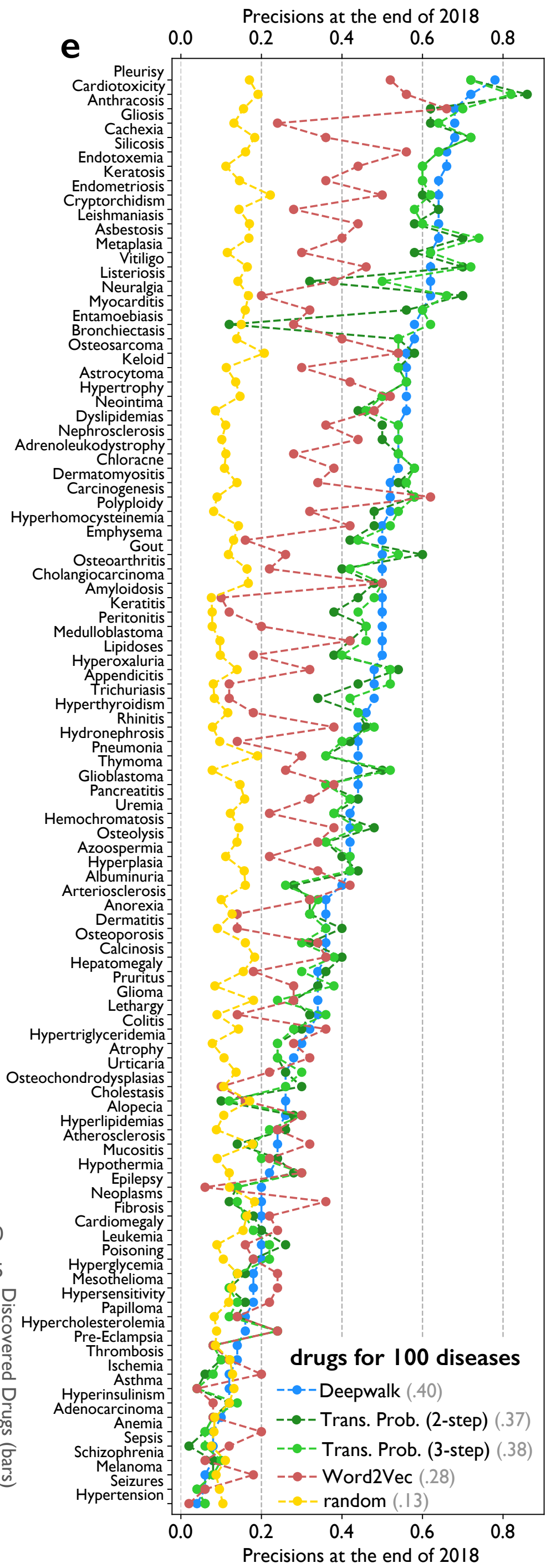

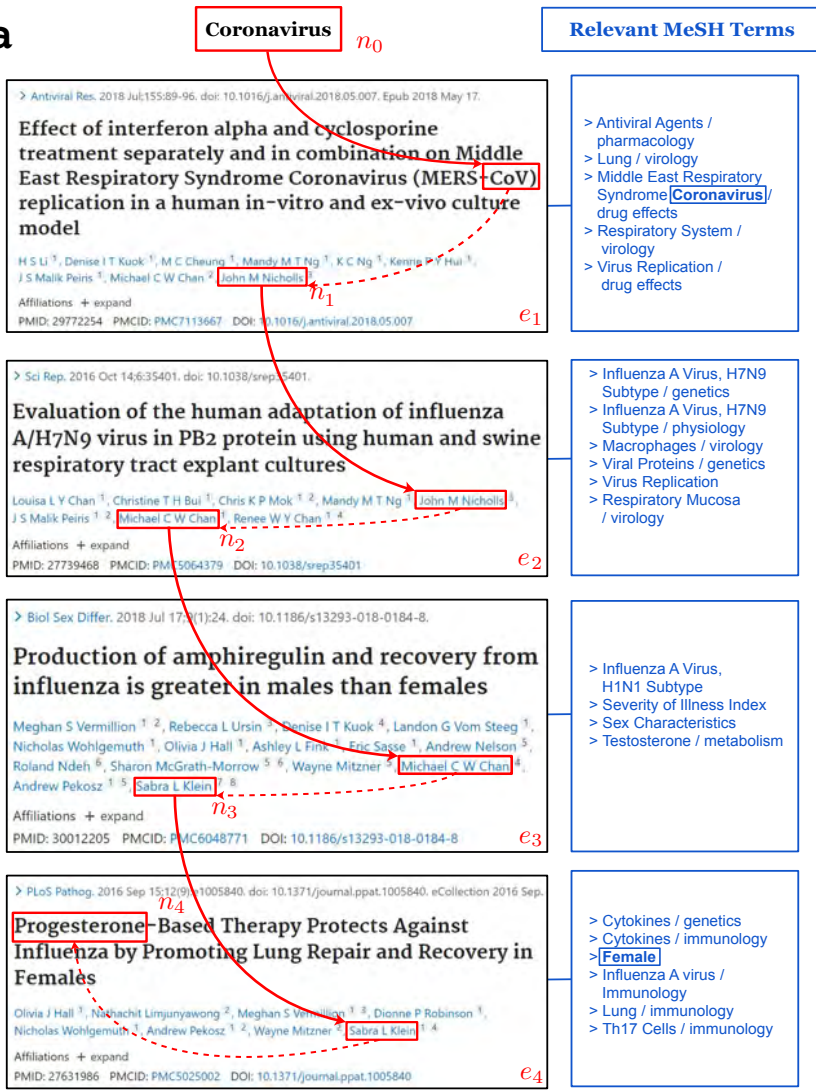
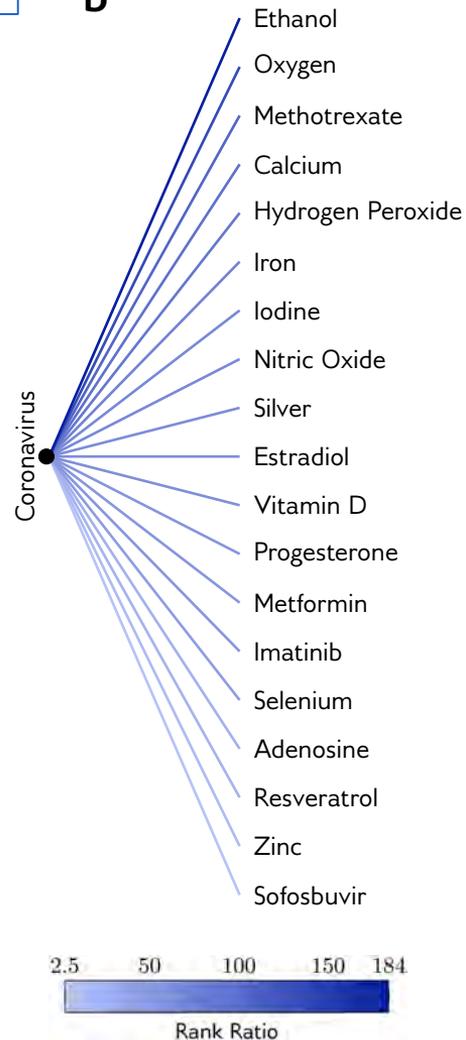

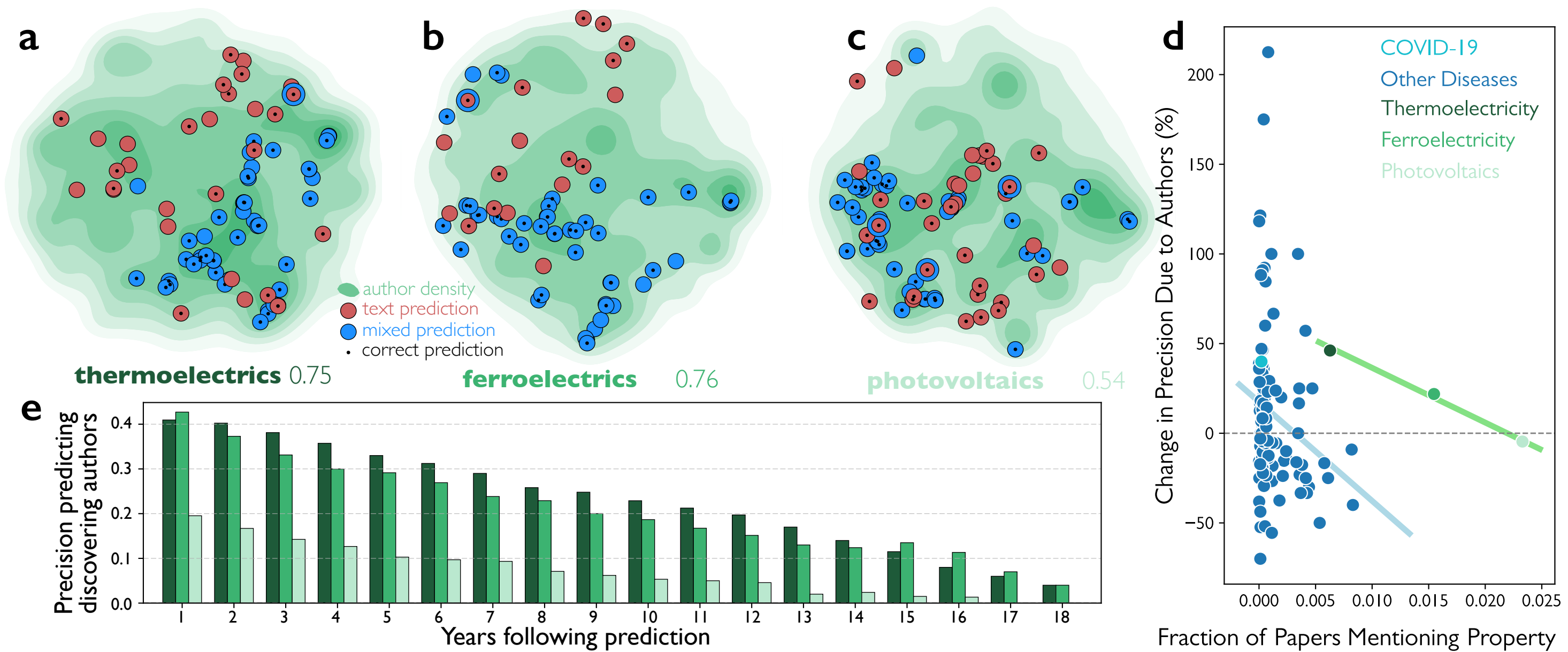

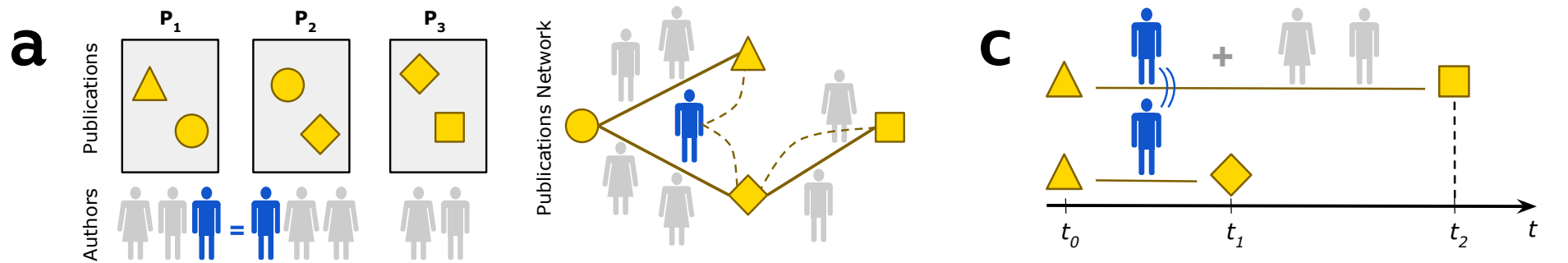
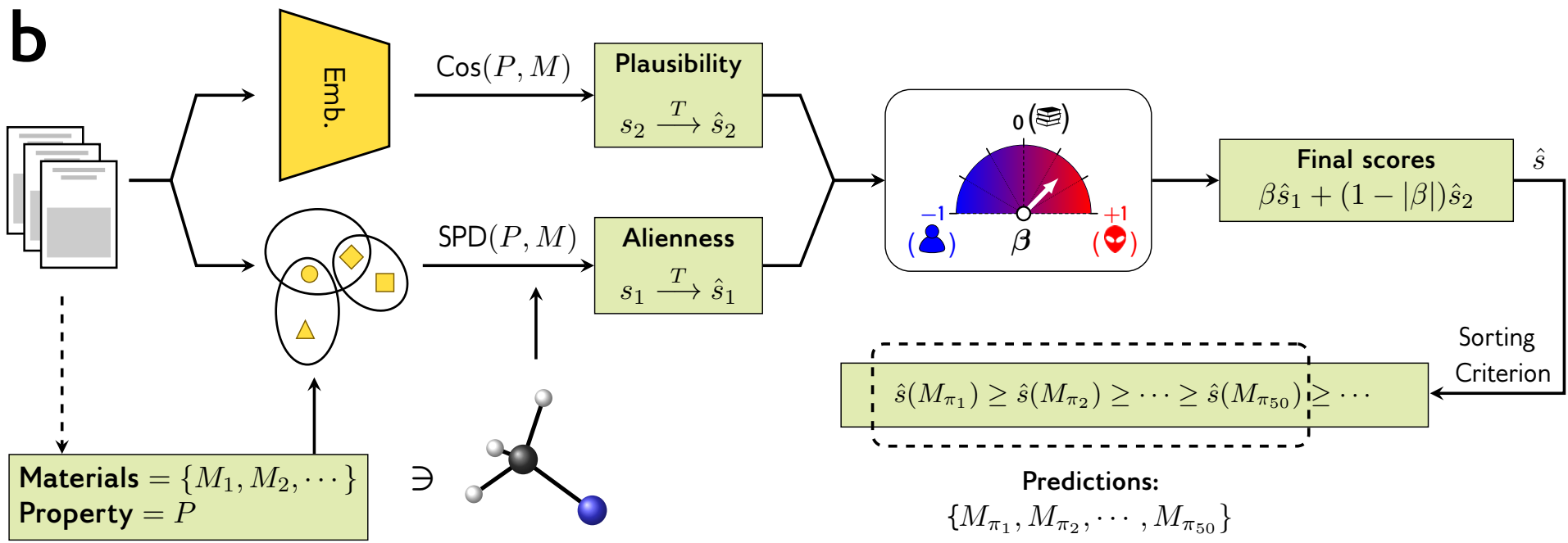

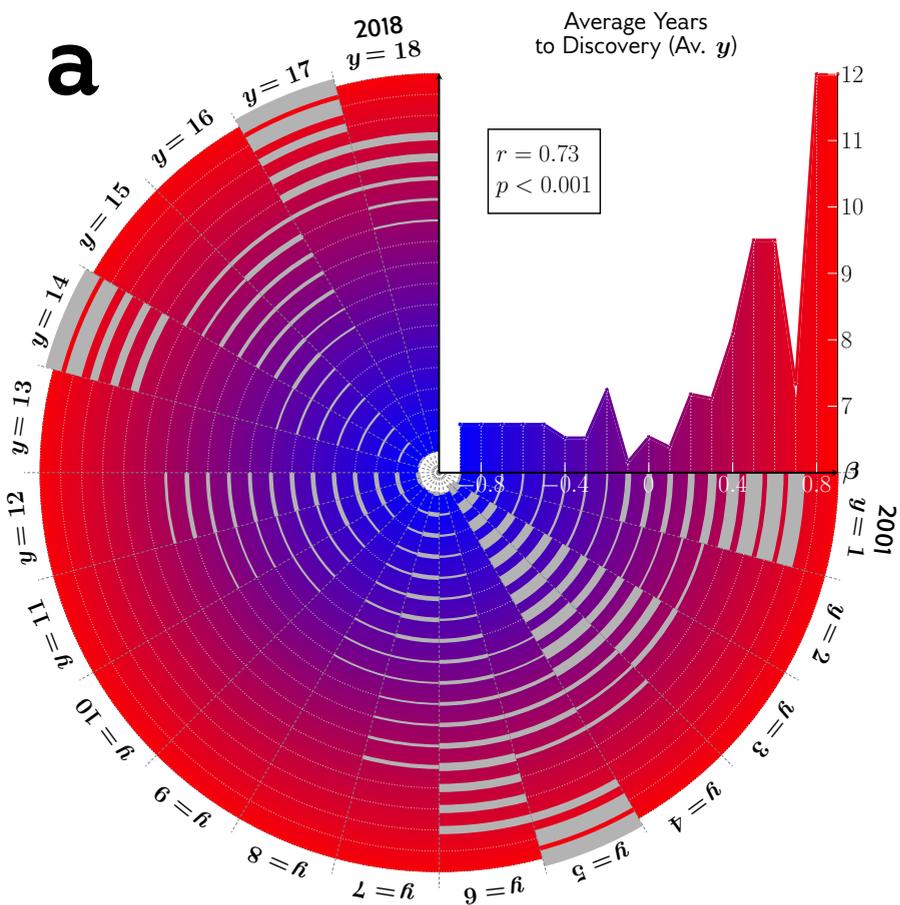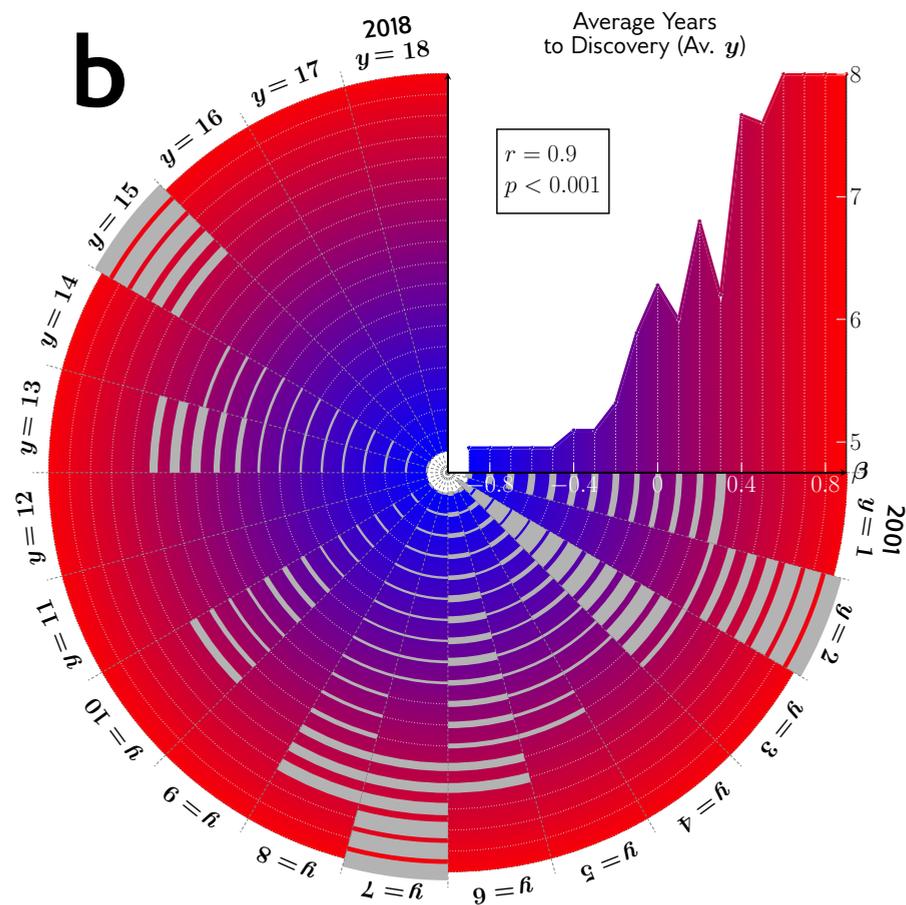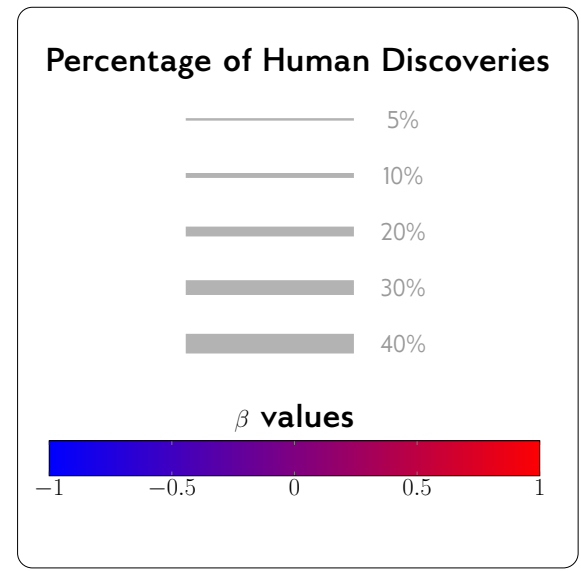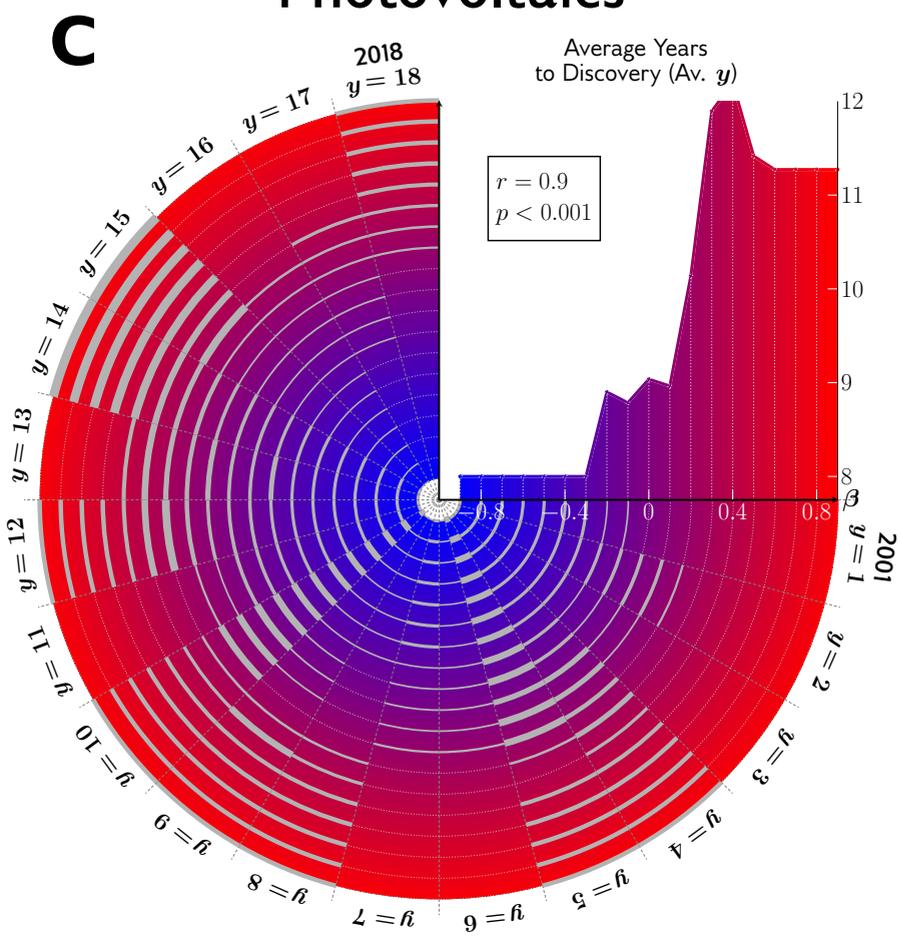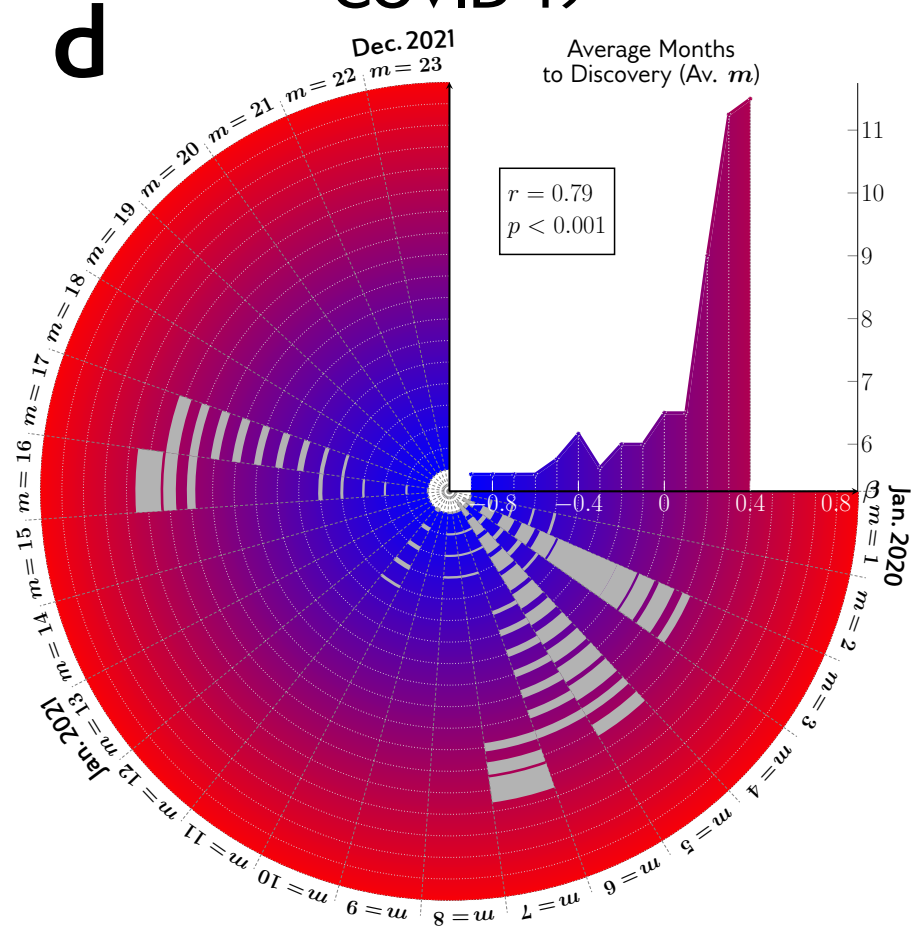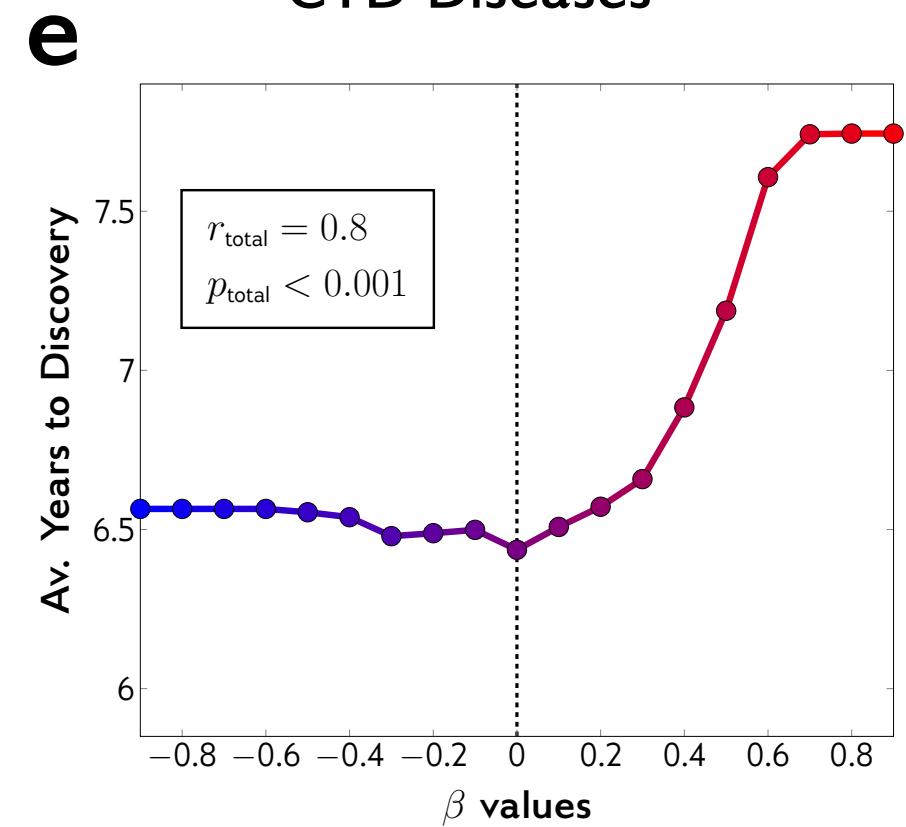

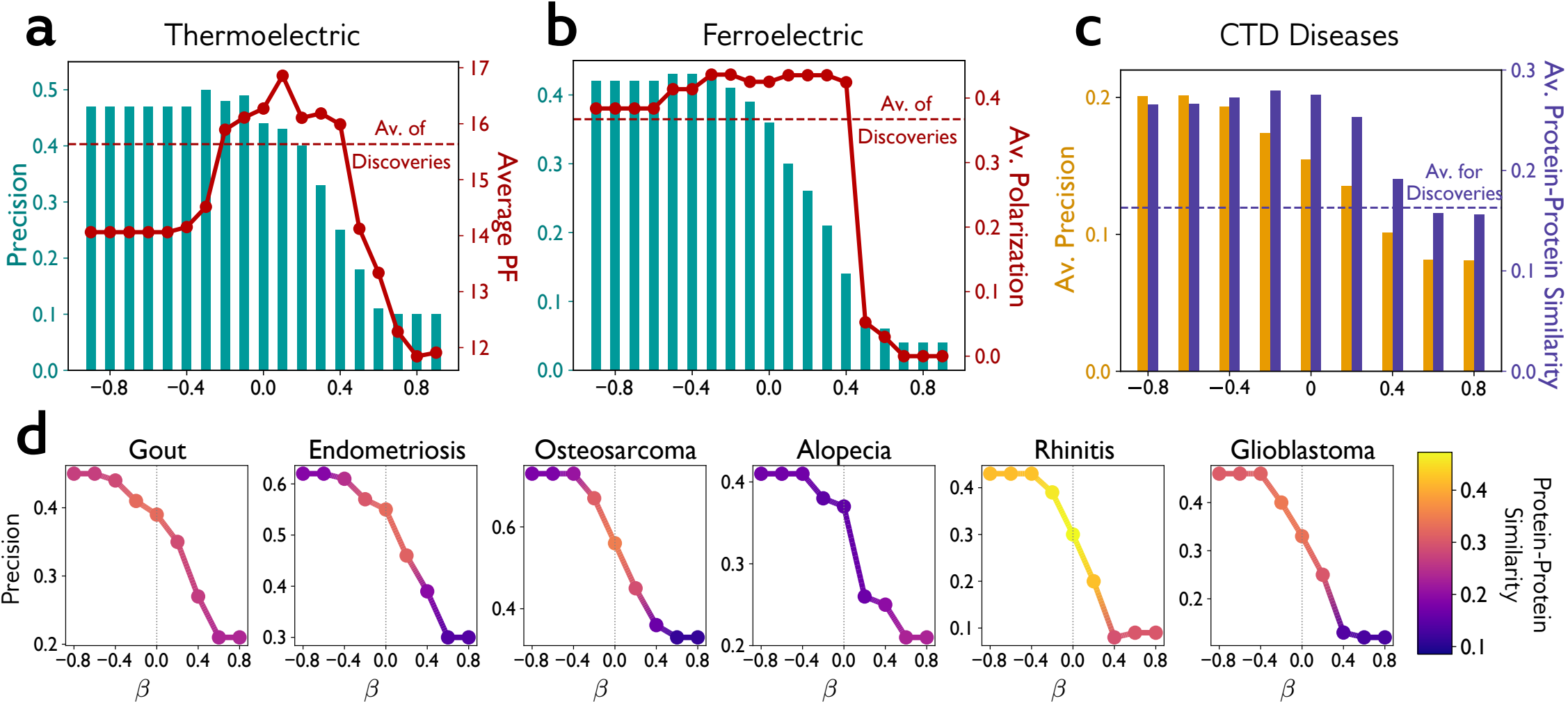

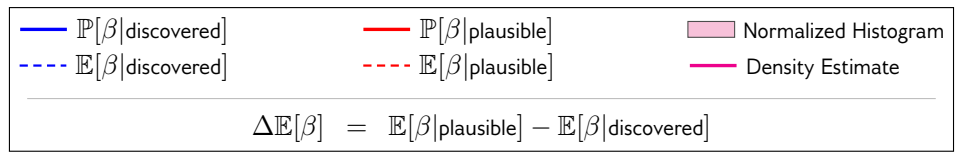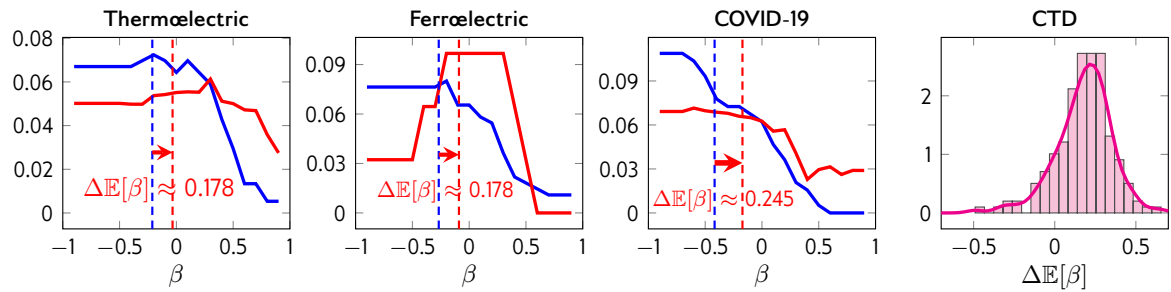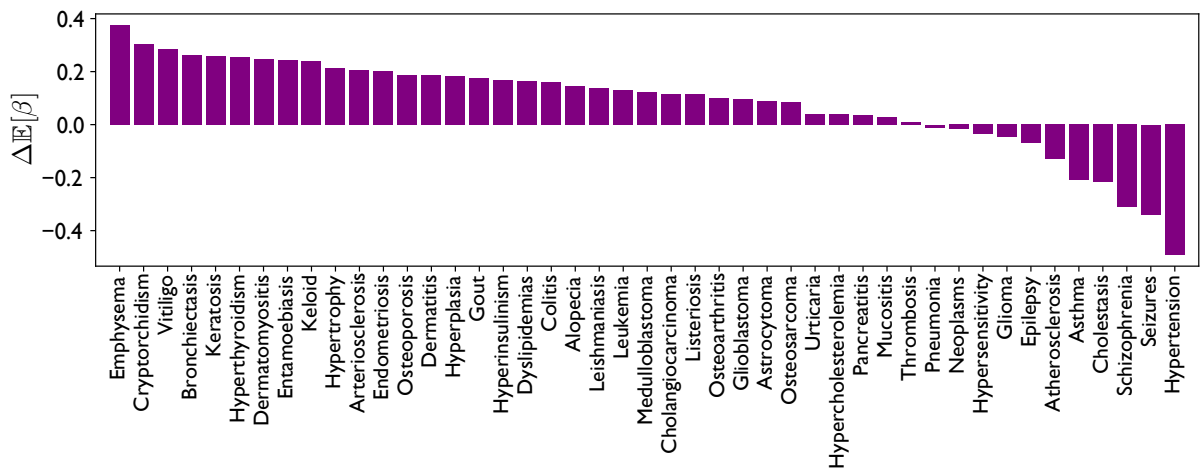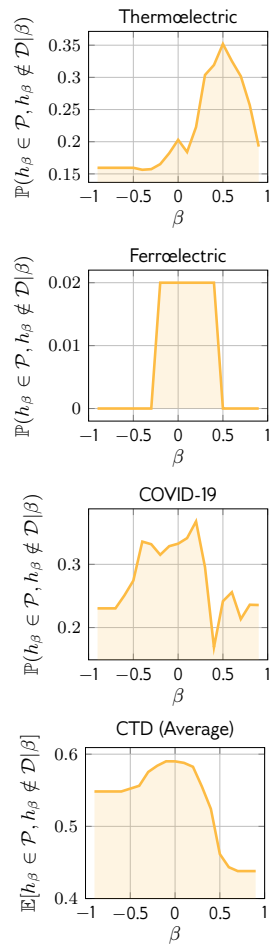

**Extended Data Fig. 1. (a-b)** Sanity checks on our hypergraph-induced transition probability similarity metric: **(a) Between an author and a conceptual node:** Histogram of the similarities between nodes of two sets of authors and the node associated with the term "coronavirus". The two sets of authors are defined as authors of 5,000 randomly selected papers from journals Nature Medicine (dark purple) and Applied Optics (light purple) between 1990 and 2019. We computed similarities between the hypernodes as the logarithm of the average transition probabilities with one and two random walk steps. The histograms are plotted considering only non-zero transition probabilities: 92% of the authors of *Nature Medicine* (28,396 in total) and 51% of the selected *Applied Optics* authors (18,530 in total) had non-zero similarity values. Also, the average non-zero similarities associated with *Nature Medicine* authors (red dashed line) is almost 5 times larger than that of *Applied Optics* authors (blue dashed line), implying that based on the hypergraph-induce similarity metric the authors publishing in *Nature Medicine* write papers more relevant to coronavirus in comparison to those publishing in *Applied Optics*. **(b) Between two conceptual nodes:** similarities between several conceptual keywords shown on the x-axis and the node corresponding to "coronavirus". Similarities between the hypernodes are computed as the average transition probabilities with one and two intermediate nodes. The terms and symptoms known to be more relevant to coronavirus have larger average transition probabilities. **(c) Schematic of our experimental settings:** Starting and ending dates of the experiments are shown. For energy-related functions and 100 human diseases, we used the beginning of 2001 as the prediction year and the end of 2018 as a single evaluation date ($V_1$). For COVID-19, the prediction year is the beginning of 2020, and we cumulatively reported monthly precision values until July of 2021 ($V_1$ to $V_{19}$).

**Extended Data Fig. 2.** Precision-Recall (PR) curves and area under the curves for various predictions and databases: energy-related material science properties, i.e., thermoelectrics **(a)**, ferroelectrics **(b)** and photovoltaics **(c)**, therapies and vaccines for COVID-19 **(d)**, and generic drug repurposing **(e)**. Except for COVID-19, we only displayed the PR-AUC values for the selected prediction years skipping the PR curves themselves. Note that whereas for Receiver Operating Curves (ROC) random predictions always result in AUC of 0.5, PR-AUC of the random baseline depends on the ratio of positive samples in the data set.

**Extended Data Fig. 3.** Calculation of expert density between property (node *P*) and each material (node *M*). Density is defined as the Jaccard index between the set of authors who have published on the property (denoted by $A_P$) and those who have mentioned the material in their publications (denoted by $A_M$). Jaccard formulation includes taking the ratio of the size of overlapping authors (denoted by $A_\cap$) to the size of union of the two sets of authors (i.e., $A_P \cup A_M$).

**Extended Data Fig. 4.** Spearman correlation coefficients between expert density (Jaccard index) of properties and materials and their date of discovery (if discovered). Negative correlations imply that materials with higher expert densities are likely to be discovered earlier than others. These results were obtained with the prediction year set to 2001 for energy-related properties and drug repurposing applications, and to the beginning of 2020 for COVID-19. Turquoise and red bars represent negative and positive correlations, respectively. For seven diseases in the CTD (shown in the bottom of the figure), all the discoveries were established in a single year and therefore no correlation coefficients could be reported. This is because we did not have accurate access to the month or day of discoveries in our database. Results indicate that energy-related properties and also COVID-19 post strong negative correlations. In the case of CTD database, 67 out of 100 diseases (i.e., properties) showed statistically significant correlations, among which only one disease had a positive coefficient. The mean correlation coefficients across these 67 diseases was -0.18.

**Extended Data Fig. 5.** Distribution of expert densities between discovery predictions and properties: **(a)** drug repurposing application (considering only the 67 diseases with statistically significant Spearman

correlation coefficients, see Extended Data Fig. 3); **(b-d)** energy-related materials science properties, i.e., thermoelectricity, ferroelectricity and photovoltaic capacity, respectively; and **(e)** therapies and vaccines for COVID-19. Curves measure normalized histograms over the logarithm of expert densities plotted by fitting a Beta distribution over expert densities for predictions. Solid and dashed vertical lines represent mean values for corresponding densities. It is clear that the distribution of expert densities for hypergarph-induced metrics (transition probability and deepwalk-based similarity) are concentrated around larger Jaccard index values than word embedding models tracing content alone. In content models, all estimated densities peak at zero ($0<a<1<b$, with $a,b$ shape parameters of Beta distributions). CTD diseases are sorted by average expert similarity between them and the complete pool of drugs.

**Extended Data Fig. 6.** Precision-Recall Area Under the Curve (PR-AUC) for predicting experts who will discover materials possessing specific properties (discoverers): **(a)** thermoelectrics, **(b)** ferroelectrics, and **(c)** photovoltaics. Materials were selected to be True Positive discovery predictions of our deepwalk-based predictor ($\alpha=1$). Our evaluation compares scores assigned to candidates and actual discovering experts who ultimately discovered and published the property associated with True Positives. We developed a deepwalk-based scoring function for this purpose. Expert candidates we considered here are those sampled at least once in deepwalk trajectories, produced over our five-year period hypergraph. For a fixed (discovered) material, scores were computed based on the proximity of experts to both property and material. An expert is a good candidate discoverer if she is close (in cosine similarity) to both property and material nodes in the embedding space. Discovered associations whose discoverers were not present in sampled deepwalk trajectories were ignored. In order to summarize the two similarities and generate a single set of expert predictions, we ranked experts based on their proximity to the property and the material and combined the two rankings using average aggregation. This ranking was used as the final expert score in our PR-AUC computations. We compared the log-PR-AUC of this algorithm with a random selection of experts and also with a curve simulating an imaginary method whose log-PR-AUC is five times higher than the random baseline. Results reveal that predictions were significantly superior to random expert selection for all electrochemical properties.

**Extended Data Fig. 7.** Illustration of decaying discoverability for predictions as $\beta$ increases. Discoverability of predictions is measured through computing the precision metric, i.e., their overlapping percentage with respect to actual discoveries made after prediction year. Decreasing precision curves and their highly negative Pearson correlation coefficients are shown for (a) thermoelectricity, (b) ferroelectricity, (c) photovoltaics and (d) COVID-19. We also visualize these statistics for the remaining human diseases with a scatterplot of their Pearson correlation coefficients (e).

**Extended Data Fig. 8.** Discoverability and scientific merit for predictions made with varying $\beta$ values in the research case repurposing drugs for treating human diseases. (a) Precision values for predictions generated with eight levels of $\beta$ and computed for all 400 human diseases we considered (except COVID-19). Diseases are sorted in terms of the number of relevant drugs. (b) Average theoretical scores measured through protein-protein similarity between diseases and candidate drugs for predictions generated with the same $\beta$ values. We compute such protein-based theoretical scores for 176 diseases out of 400 total cases (44%). In both subfigures, horizontal lines show average values across all diseases.

**Extended Data Table 1.** List of 47 clinical MeSH terms that appeared with a frequency higher than random selection in the hyperedges of random walk sequences from property node "Coronavirus" to candidate material node "Progesterone".

**Extended Data Table 2.** True positive candidate materials that were exclusively predicted by our expert-aware deepwalk algorithm (unshaded rows) and content-only algorithm (shaded rows). $rank_{DW}$ and $rank_{W2V}$ denote the ranks assigned to materials by our deepwalk-based approach and the Word2Wec baseline, respectively. The scores generated by each prediction algorithm is defined as cosine similarity of the materials with respect to the targeted property ("Coronavirus") based on their corresponding trained embedding model, i.e., word2vec models trained on random walk sequences (for $rank_{DW}$) and text of the abstracts (for $rank_{W2V}$). A material will be reported as a discovery prediction, if the algorithm ranks it higher than 50.

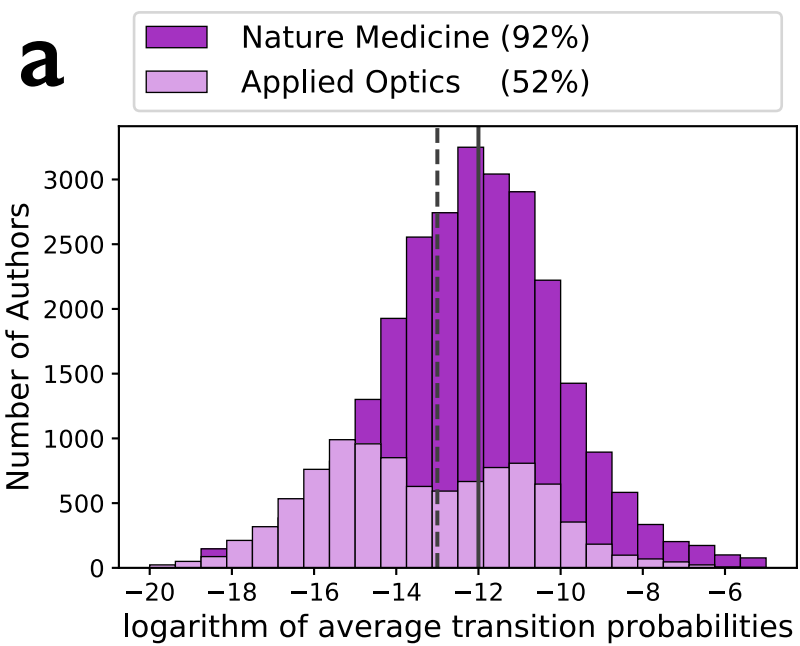
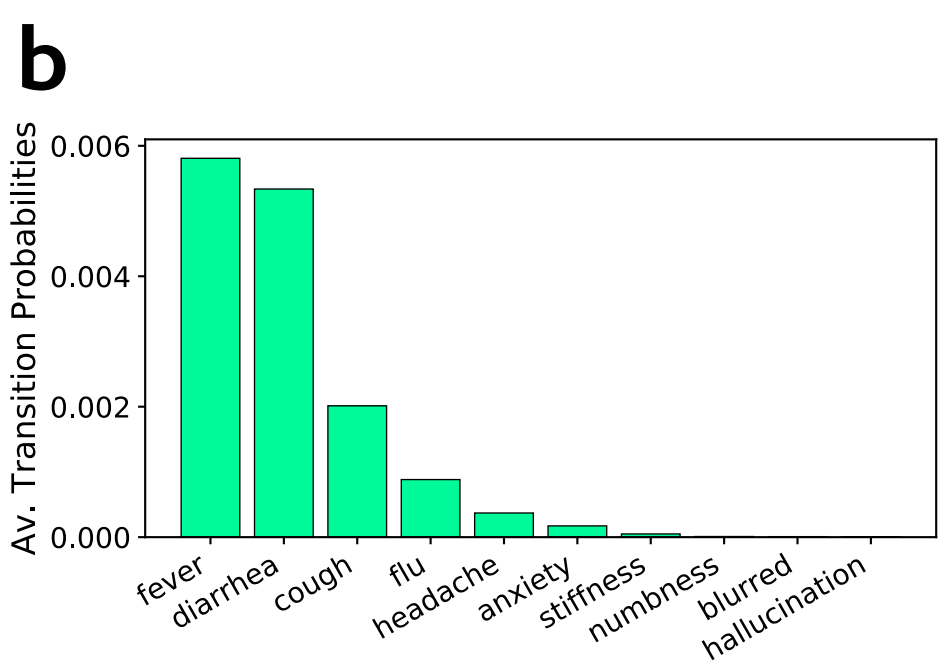
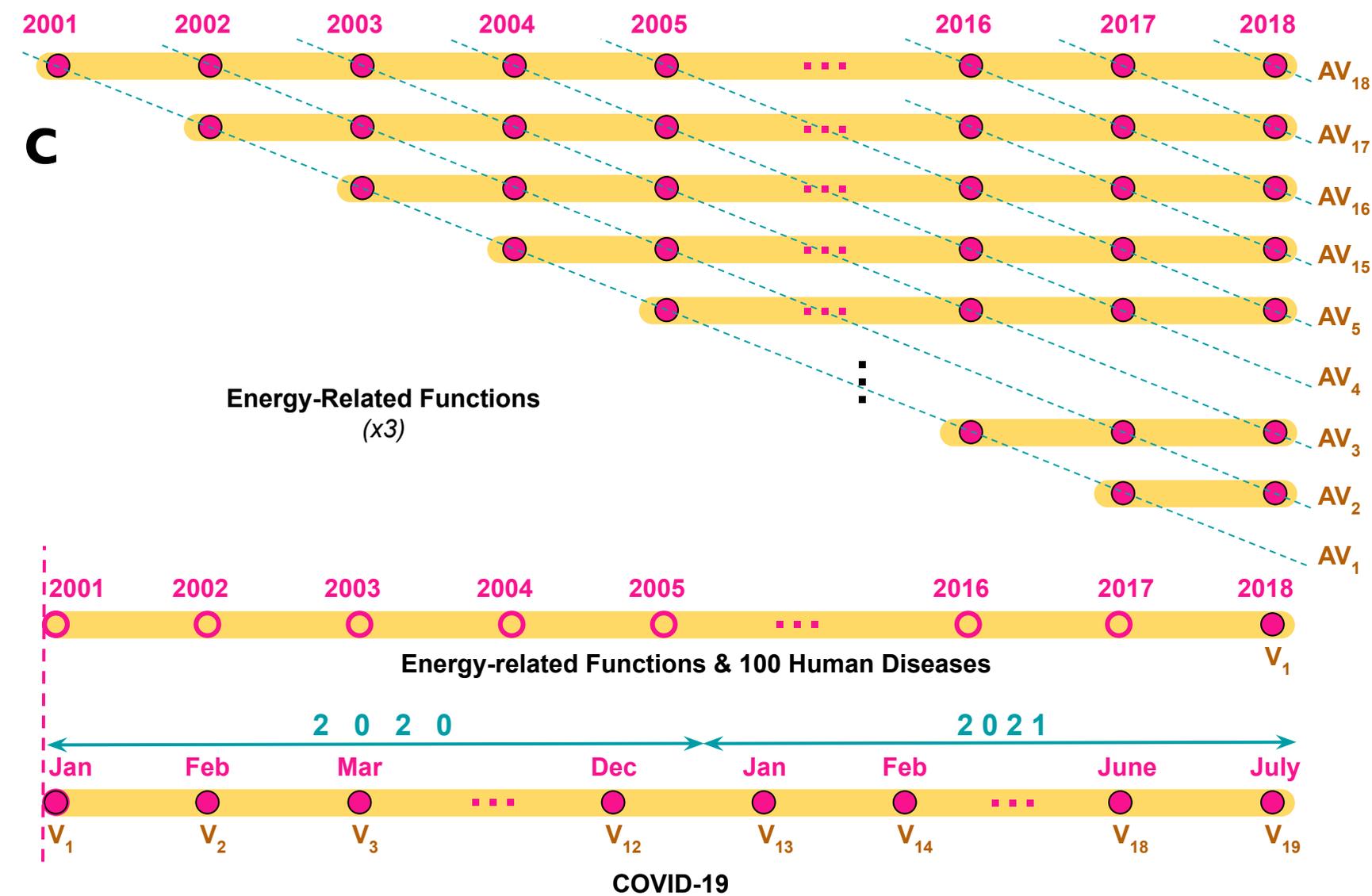

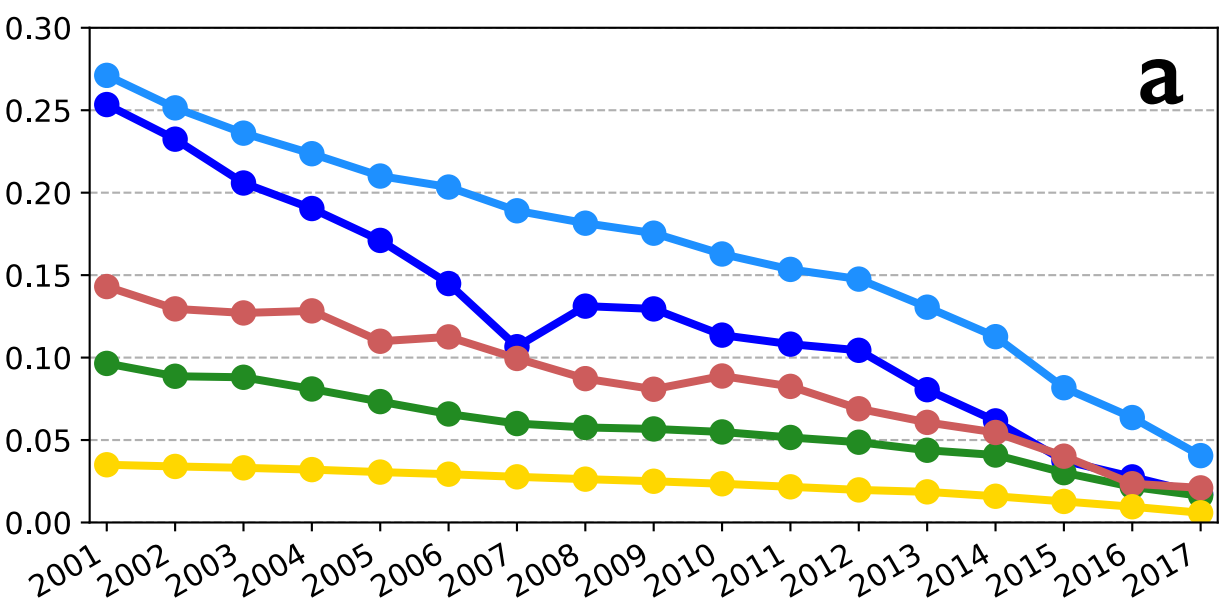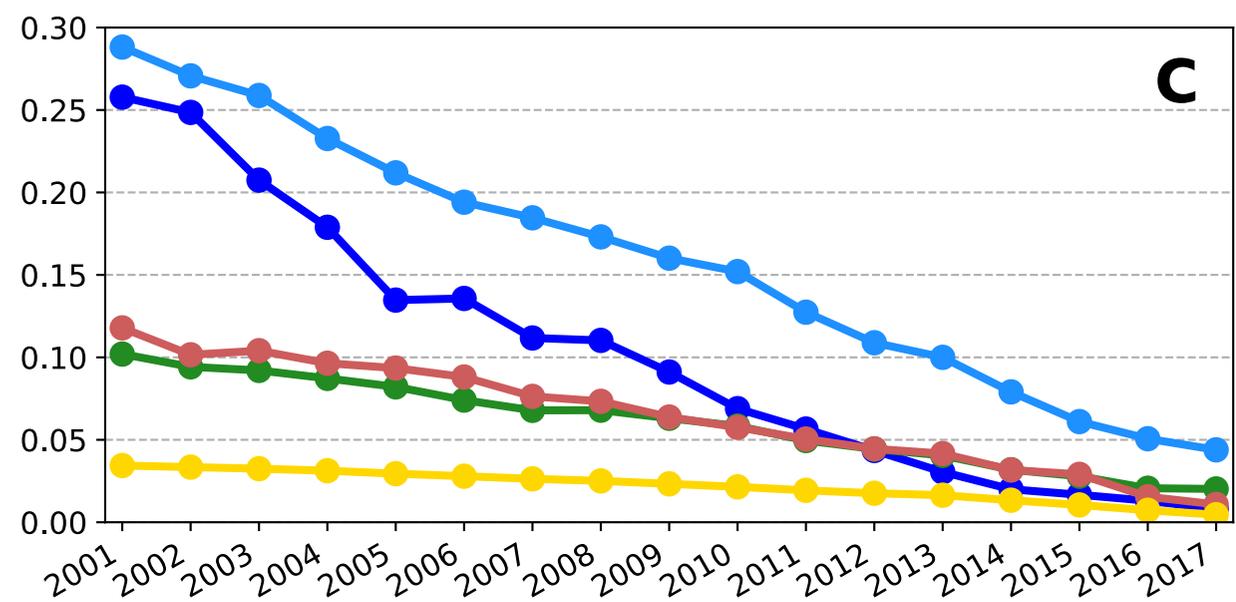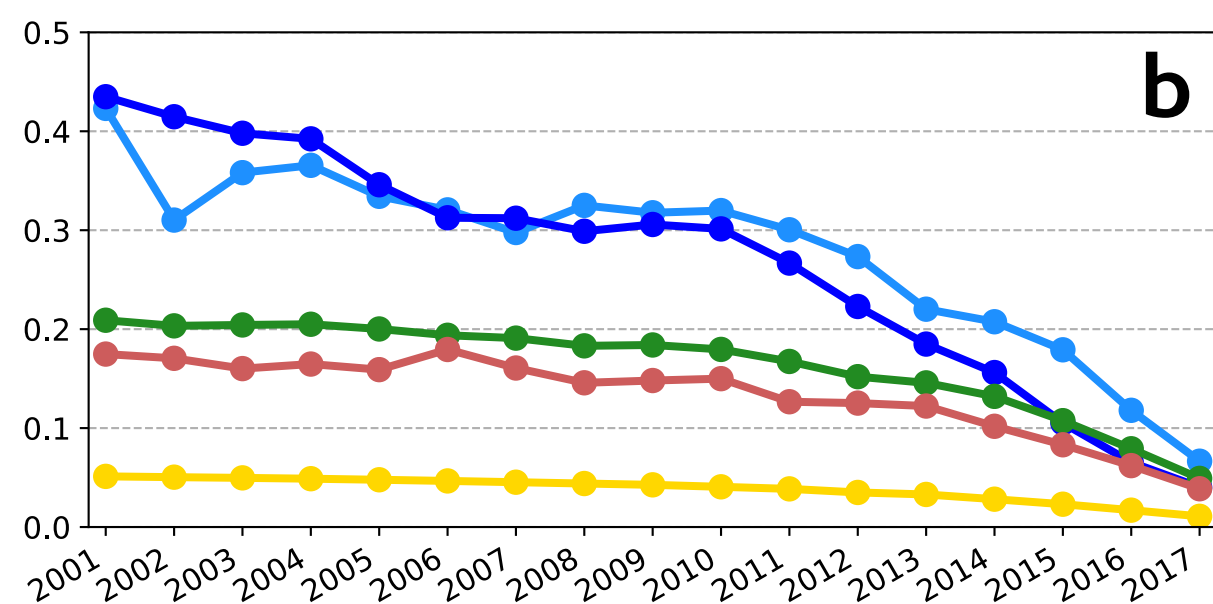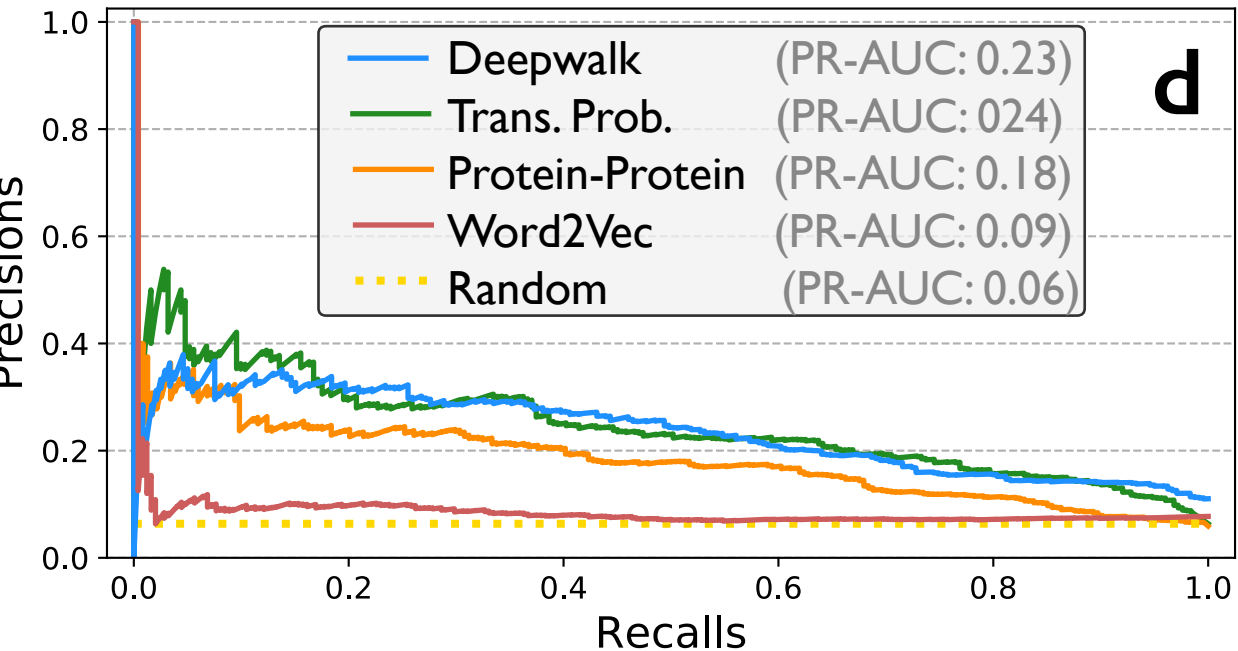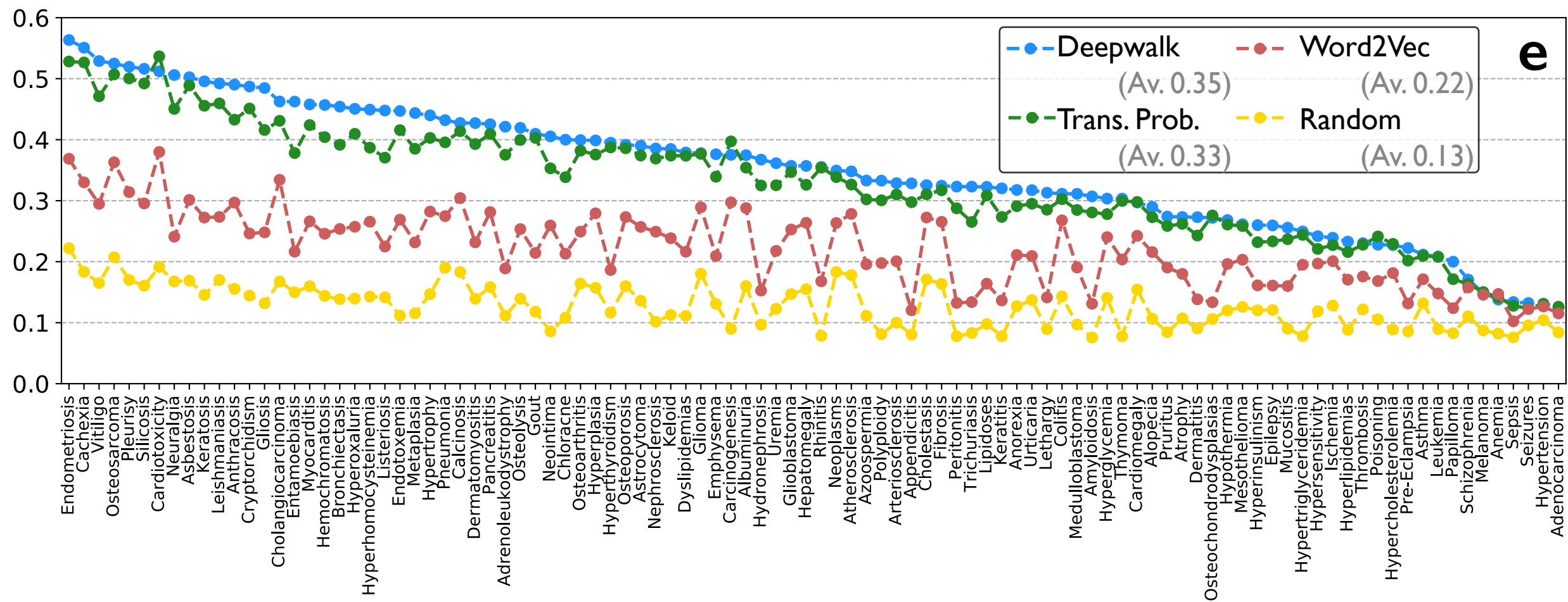

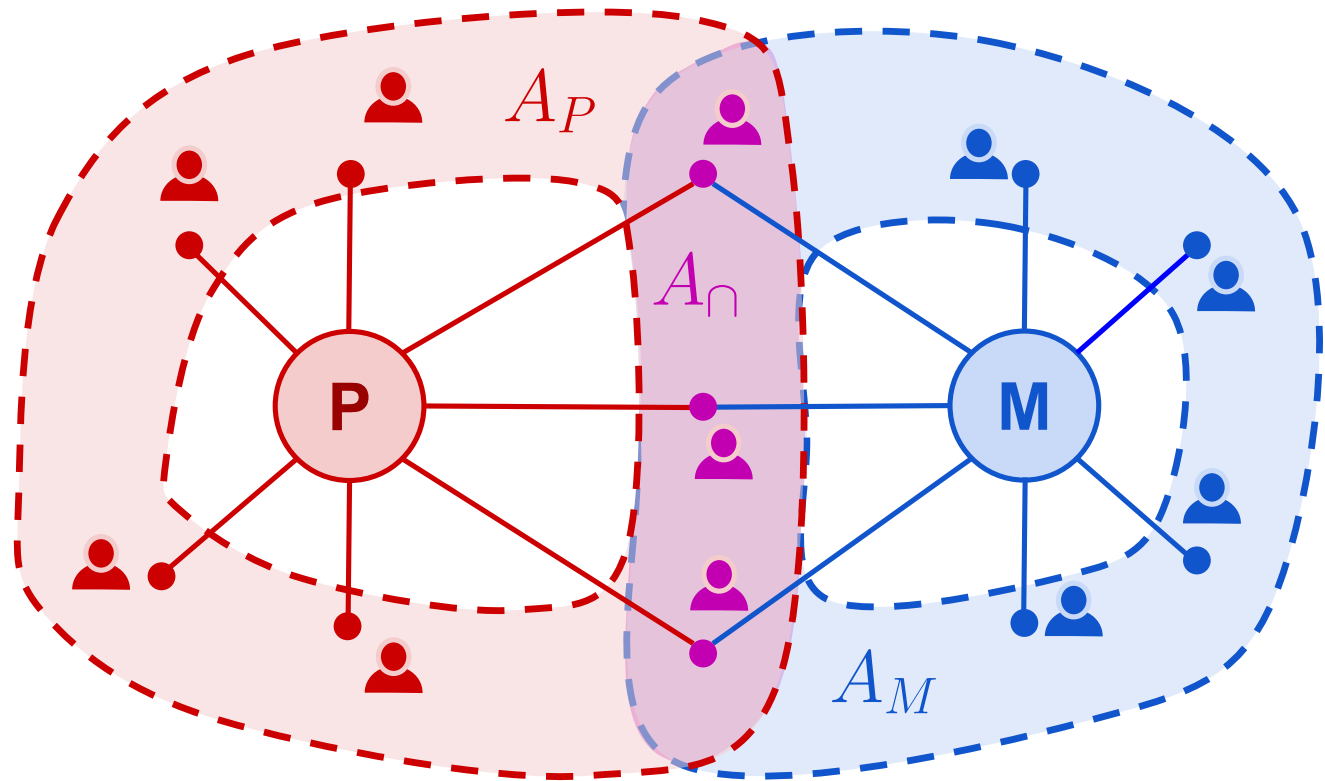

$$\text{Expert Density}(P, M) = \frac{|A_P \cap A_M|}{|A_P \cup A_M|} = \frac{|A_\cap|}{|A_P \cup A_M|}$$

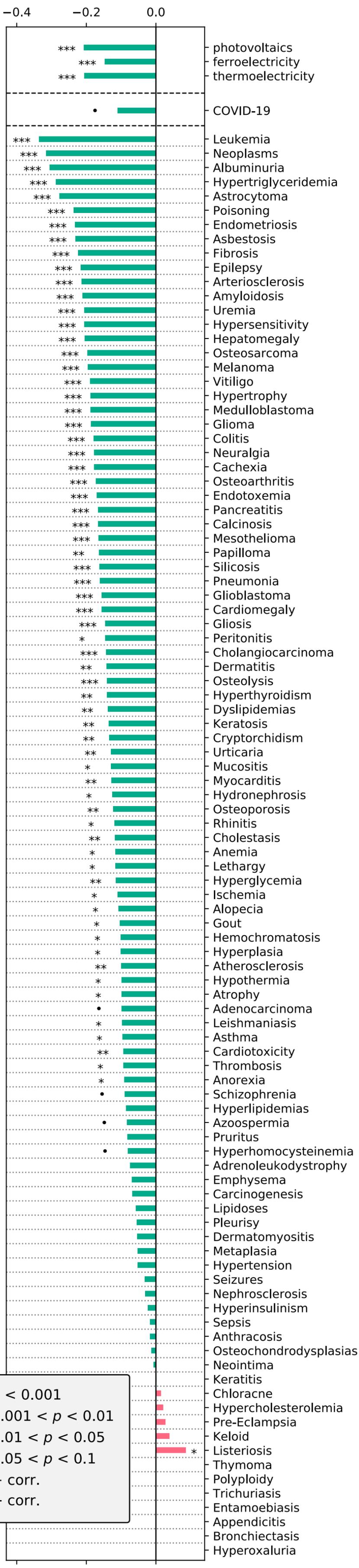

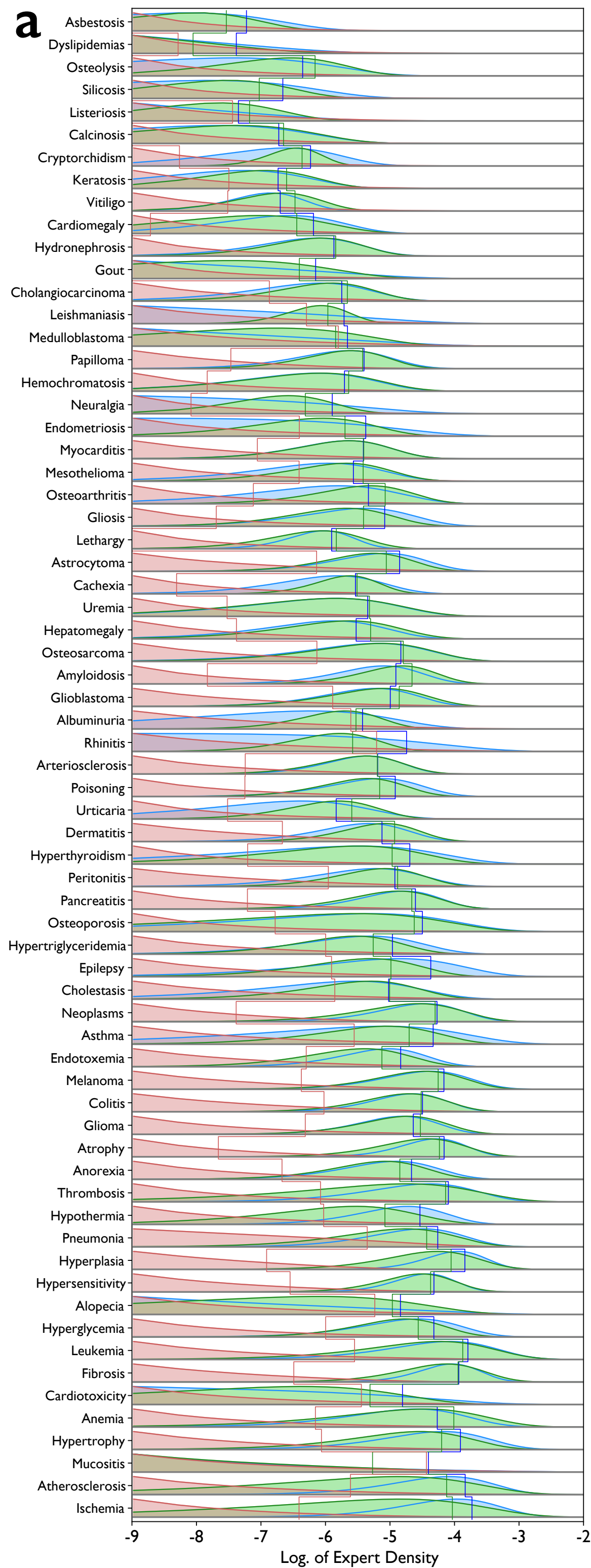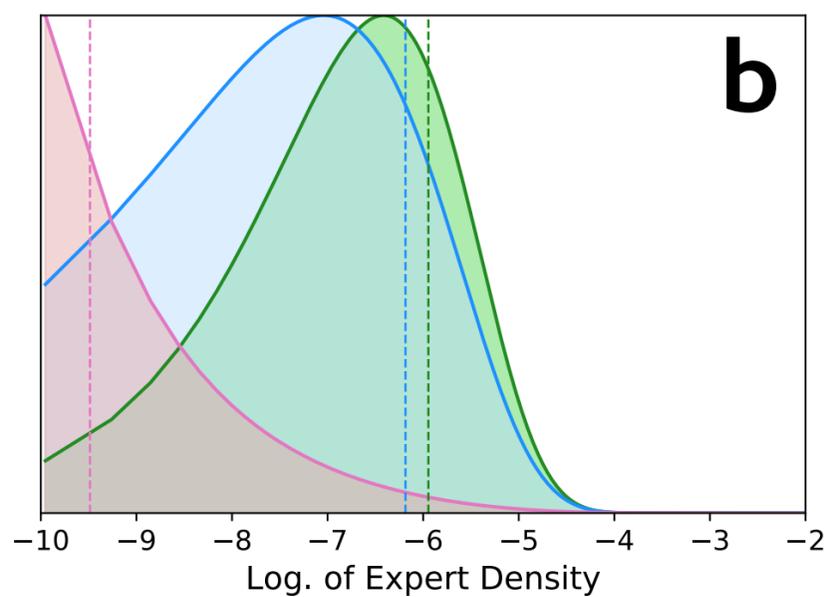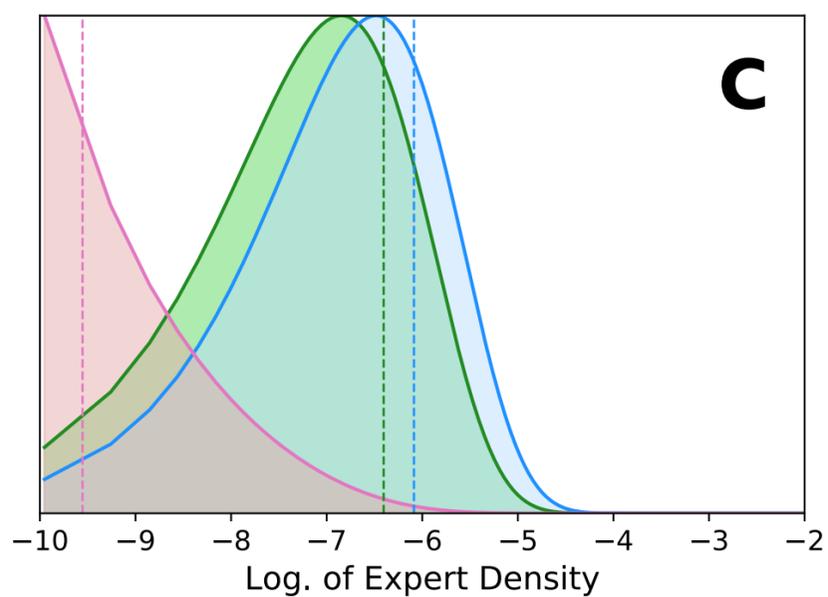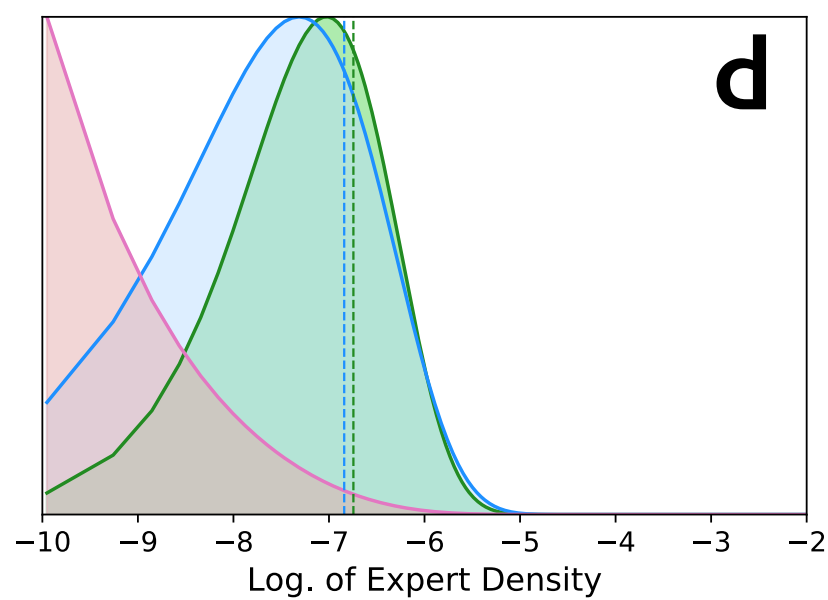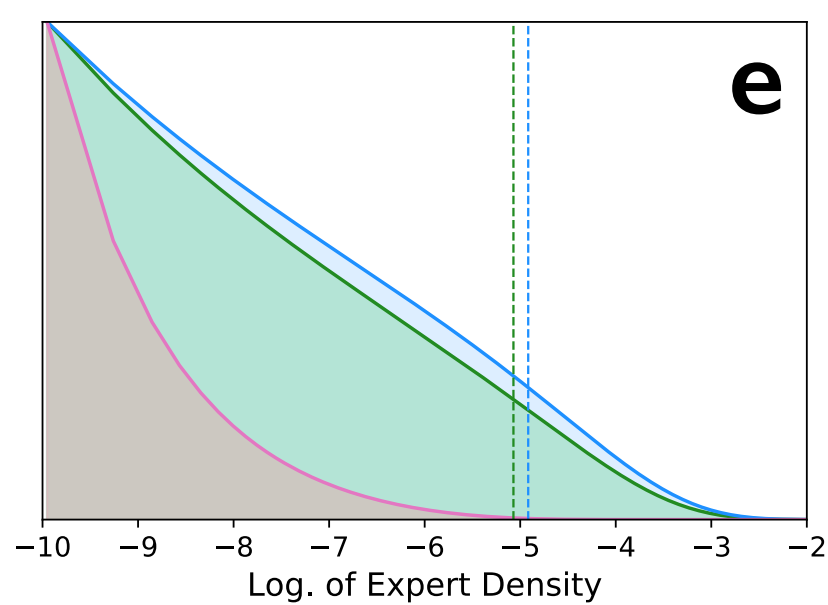

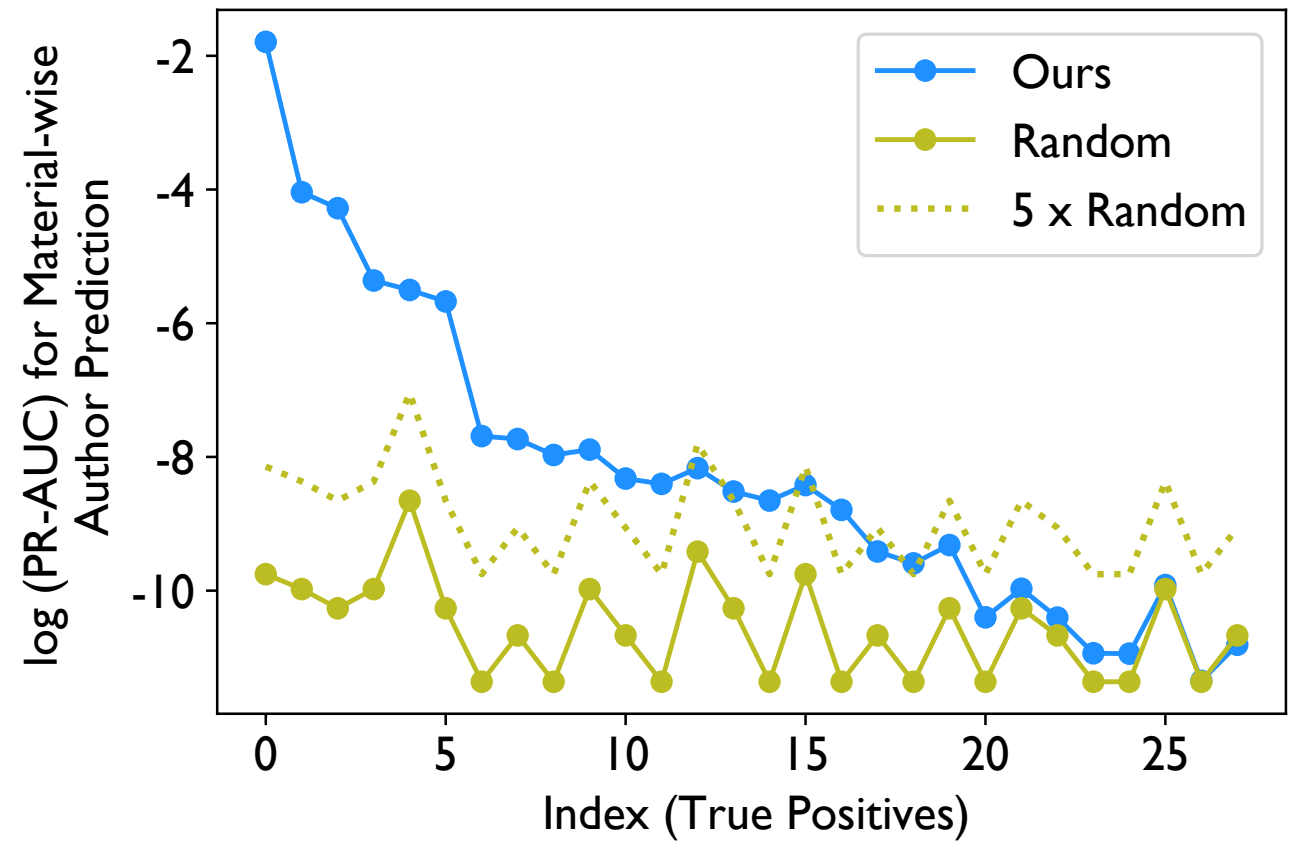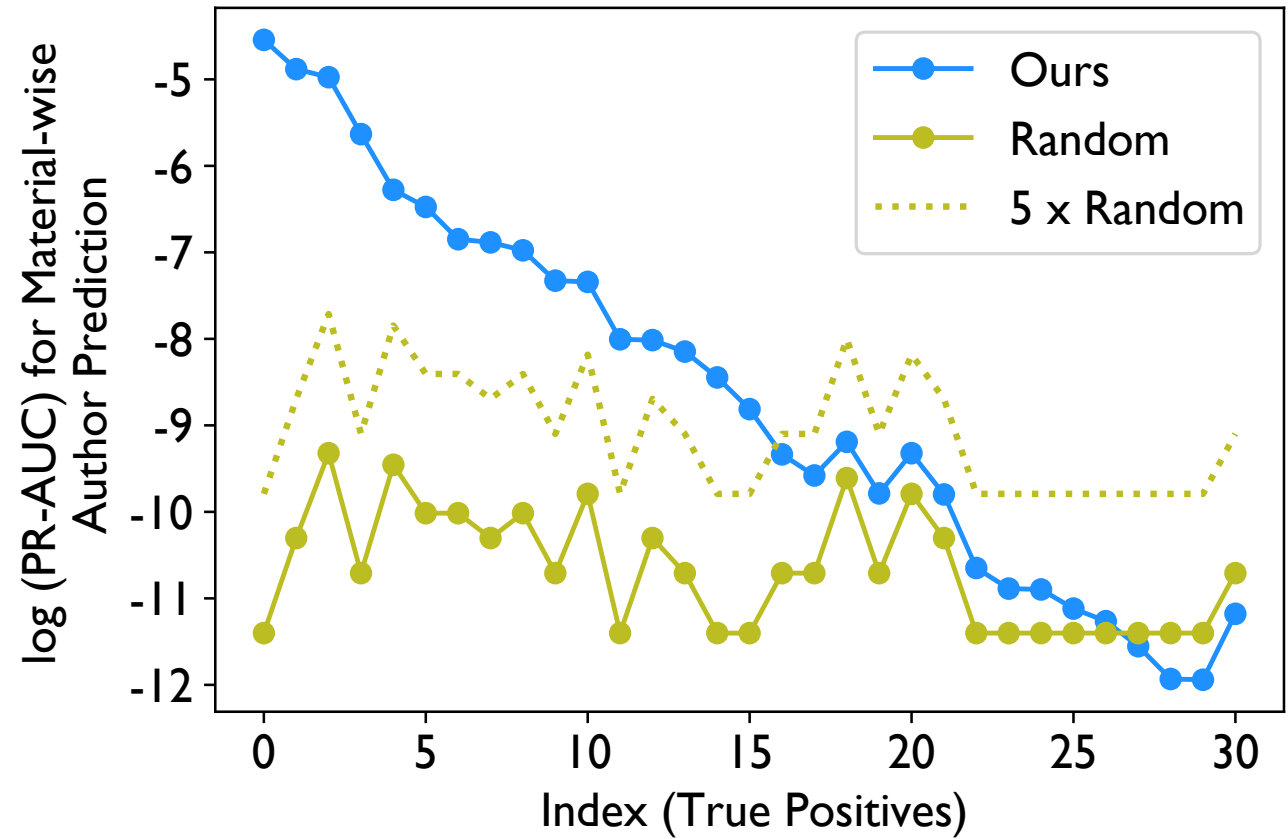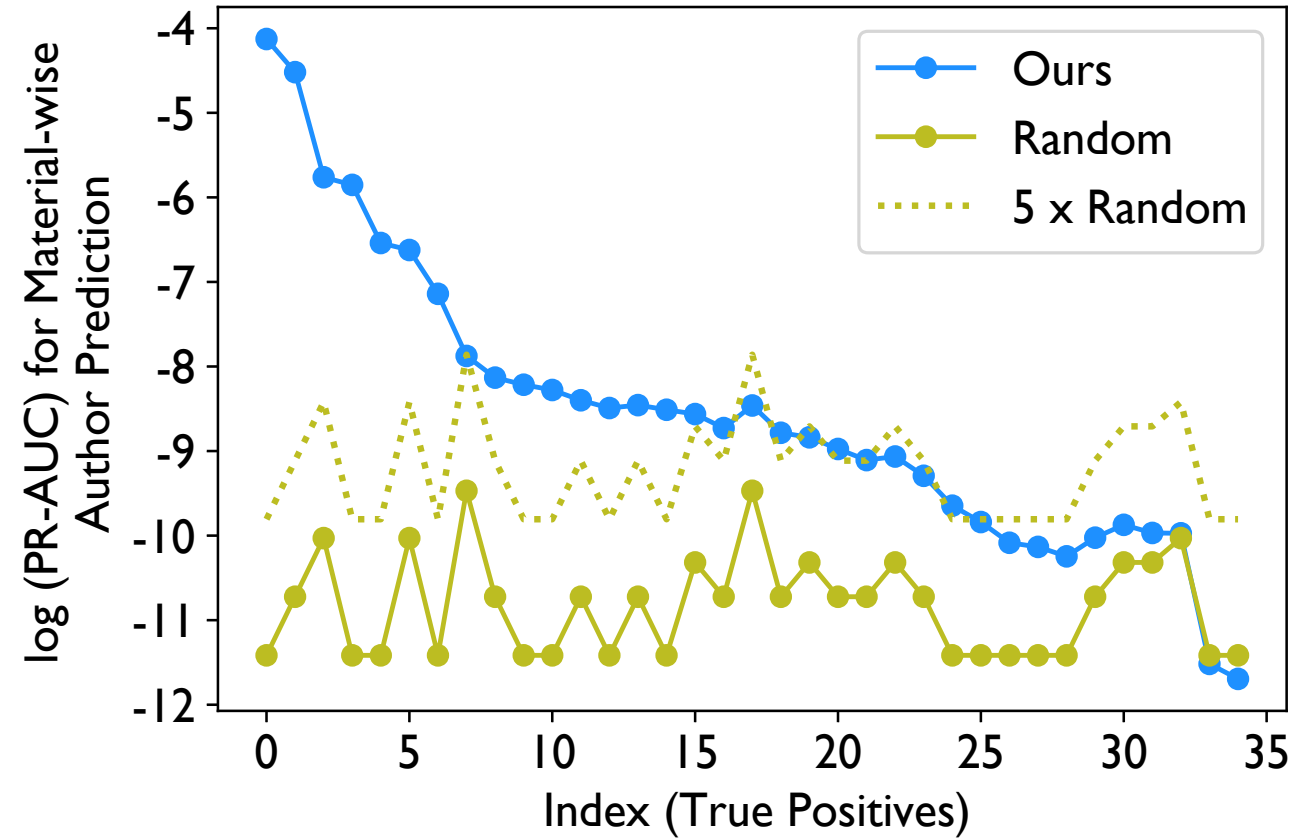

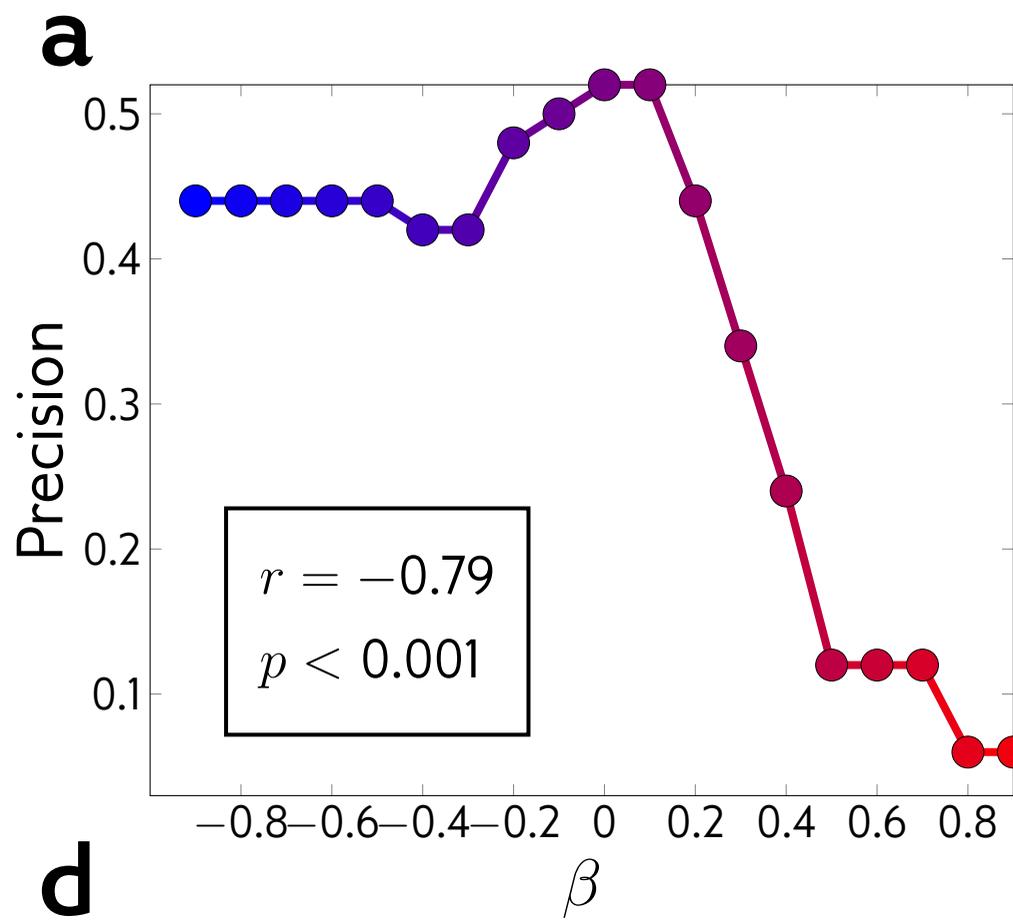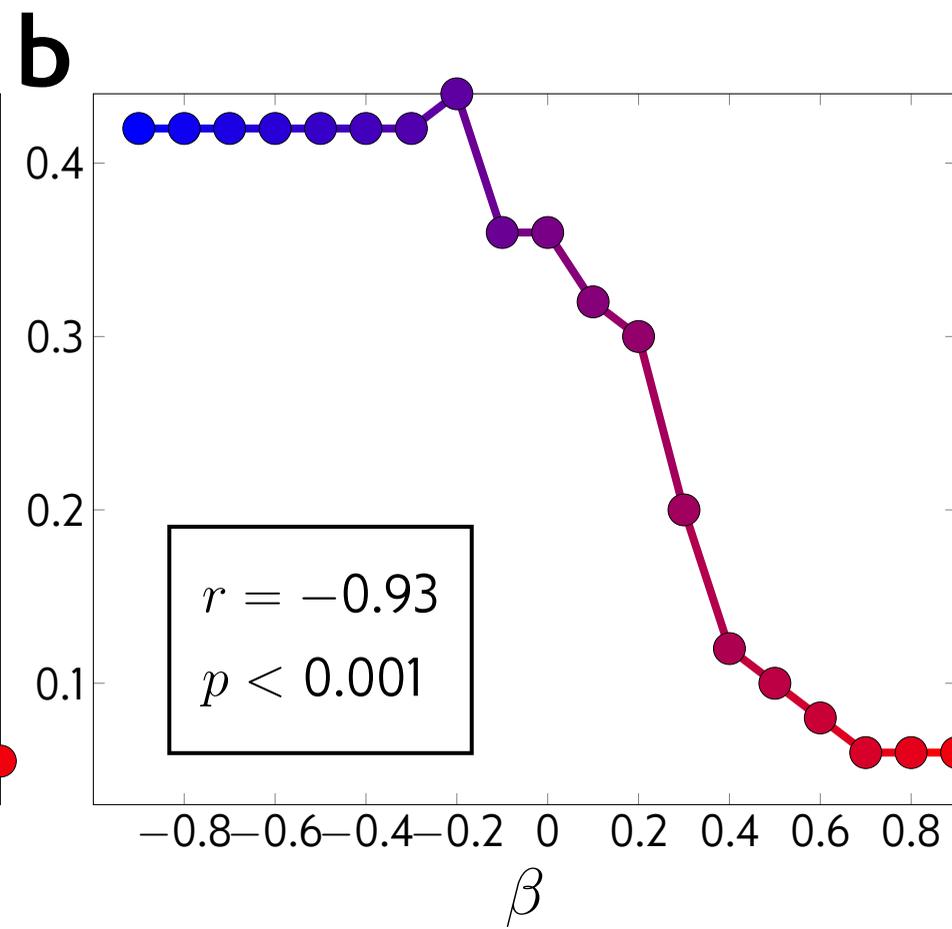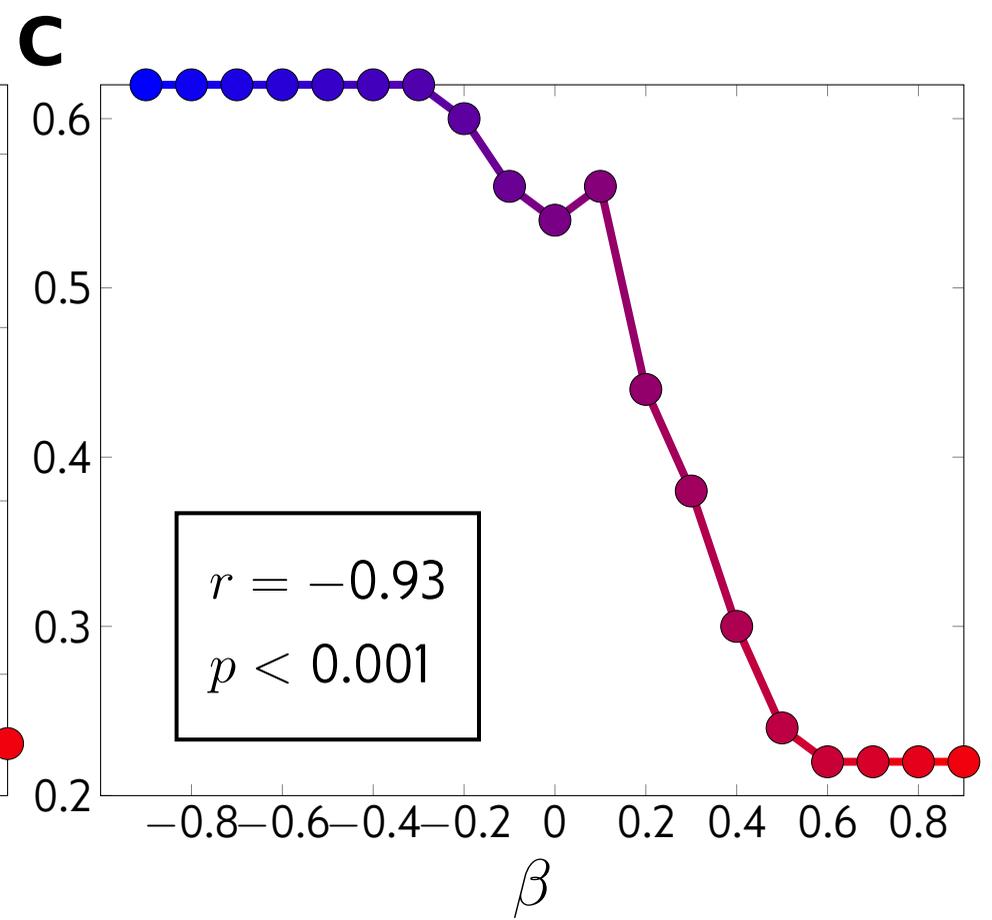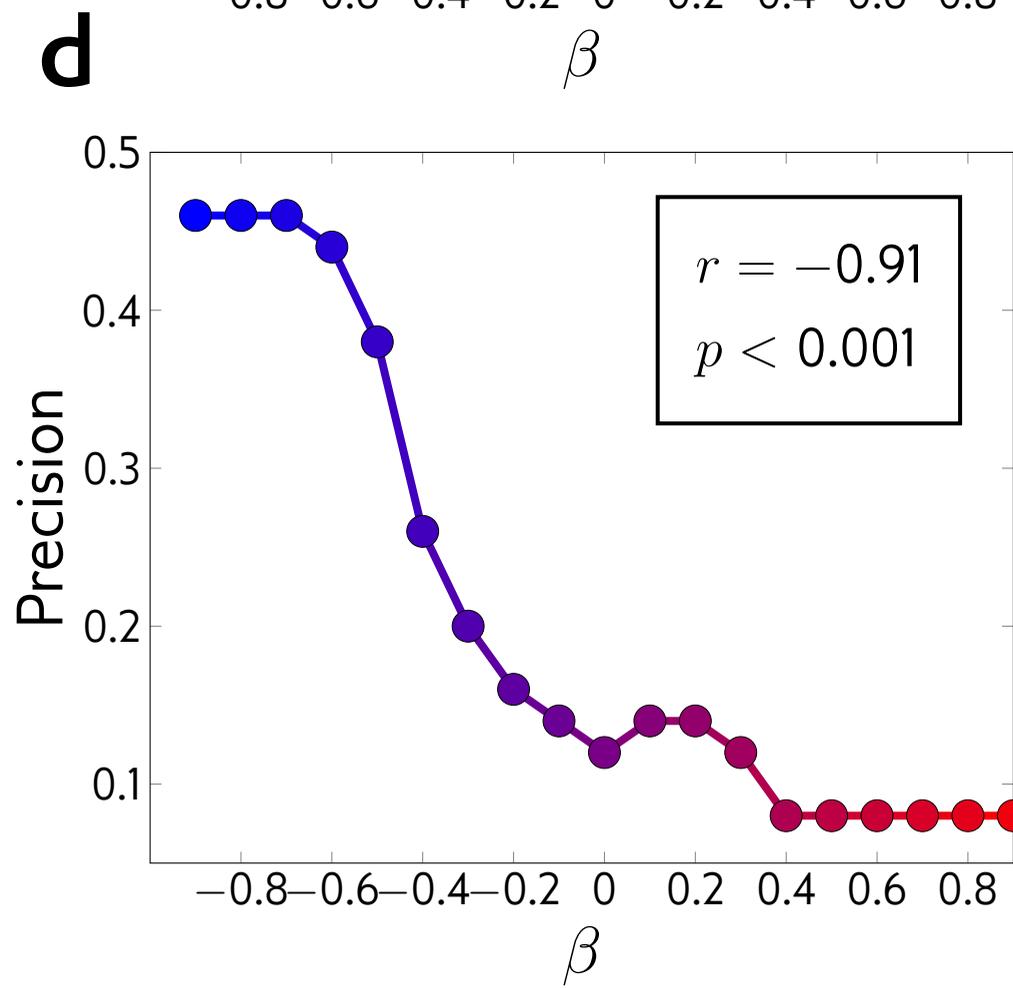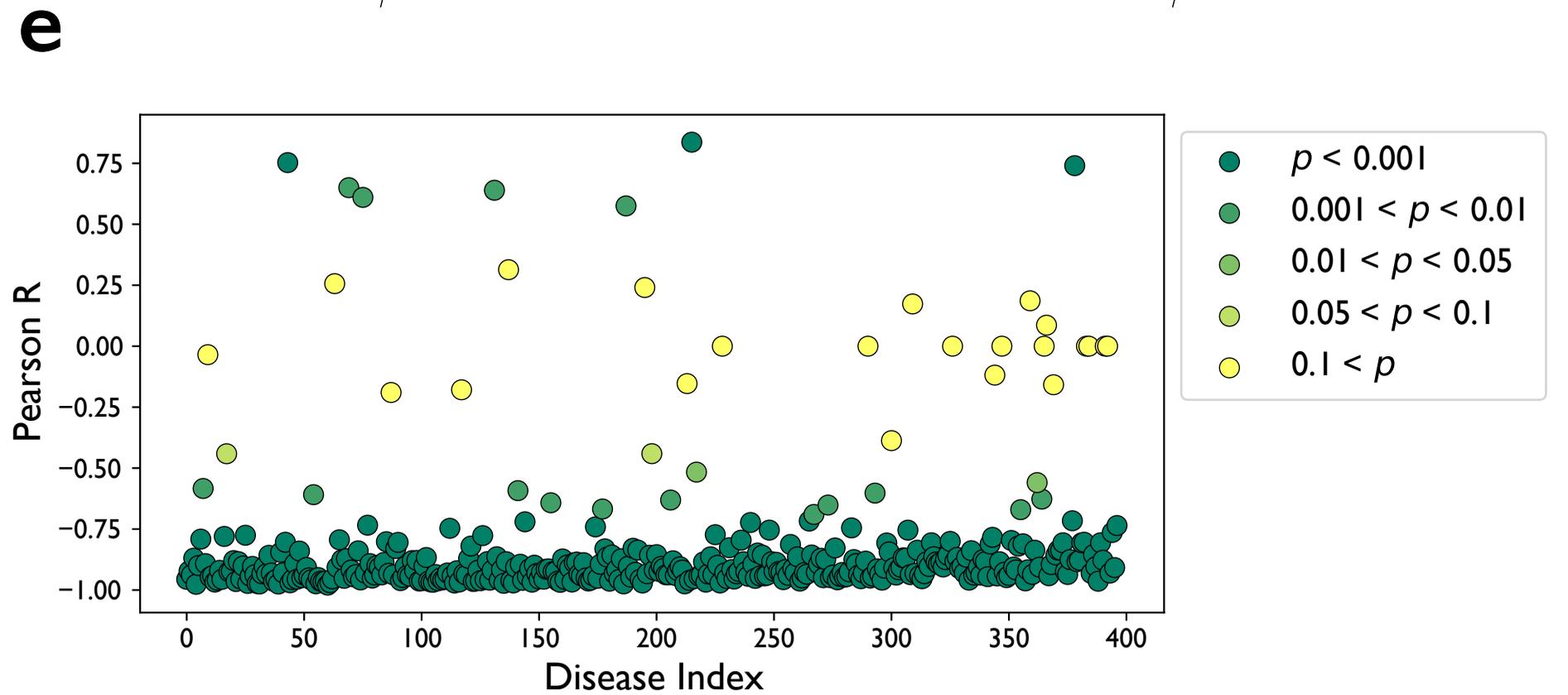

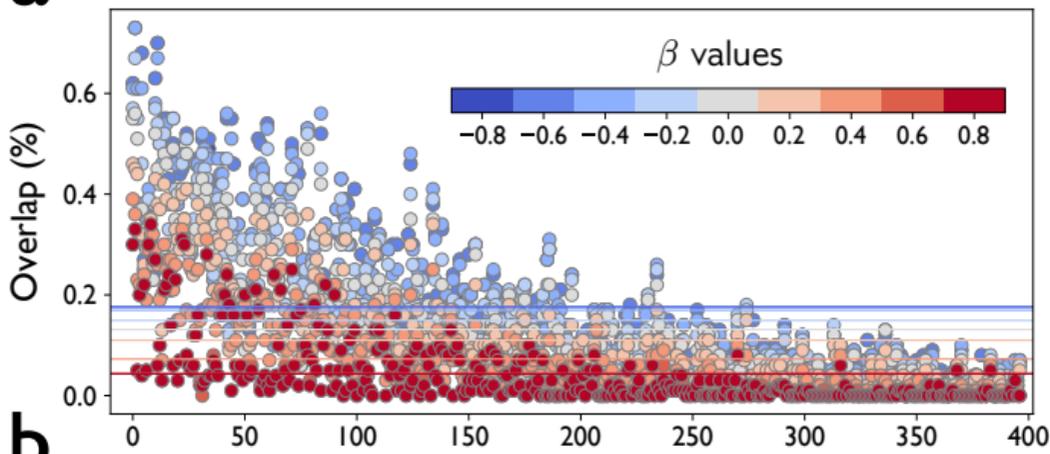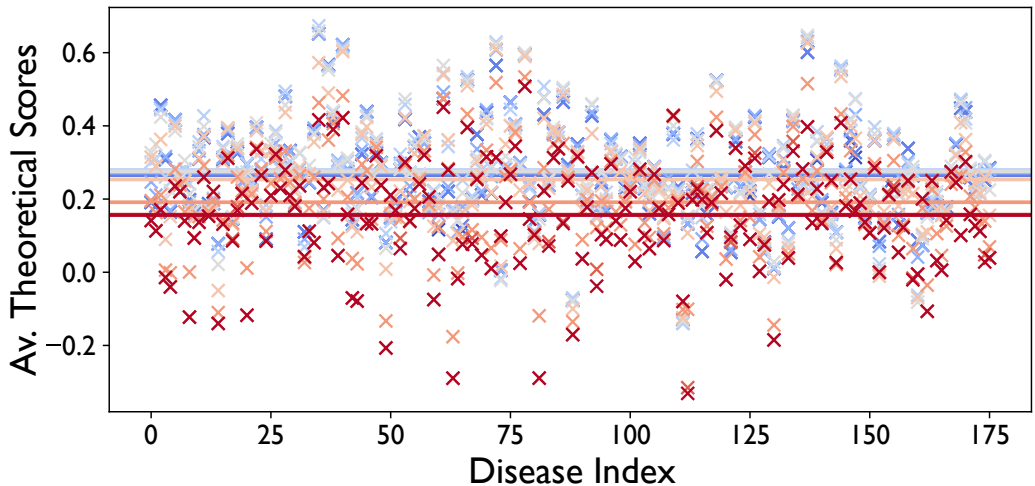

| No. | MeSH Descriptor | MeSH Qualifier | No. | MeSH Descriptor | MeSH Qualifier |
|---|---|---|---|---|---|
| 1 | Middle East Respiratory Syndrome Coronavirus | Immunology | 21 | Endometrium | |
| | | Genetics | 22 | Gonadotropin-Releasing Hormone | |
| | | Isolation & Purification | 23 | Vero Cells | |
| 2 | SARS Virus | | 24 | Antibodies, Neutralizing | Immunology |
| 3 | Coronavirus | Diagnosis | 25 | Follicle Stimulating Hormone | |
| | | Virology | 26 | Influenza A virus | |
| | | Immunology | 27 | Viral Nonstructural Proteins | |
| | | Epidemiology | 28 | Estrogen Receptor alpha | |
| | | Prevention | 29 | Genome, Viral | |
| | | Veterinary | 30 | Pregnancy Rate | |
| 4 | Progesterone | Administration & Dosage | 31 | Influenza, Human | Virology |
| | | Blood | 32 | Receptor, ErbB-2 | Metabolism |
| | | Pharmacology | 33 | Zoonoses | |
| | | Metabolism | 34 | RNA, Viral | Genetics |
| | | Genetics | 35 | Triple Negative Breast Neoplasms | |
| 5 | Insemination, Artificial | Veterinary | 36 | Uterus | |
| 6 | Estrous Cycle | | 37 | Carcinoma, Ductal, Breast | |
| 7 | Respiratory Tract Infections | Virology | 38 | Viral Proteins | Genetics |
| 8 | Swine Diseases | Virology | 39 | Testosterone | Blood |
| 9 | Progestins | | 40 | Ovary | |
| 10 | Estradiol | Blood | 41 | Influenza, Human | Epidemiology |
| | | Pharmacology | 42 | Virus Replication | |
| 11 | Ovulation | | 43 | Antiviral Agents | Pharmacology |
| 12 | Feces | Virology | 44 | Seroepidemiologic Studies | |
| 13 | Ovarian Follicle | | 45 | Breast Neoplasms | Mortality |
| 14 | Respiratory Tract Infections | Epidemiology | | | Metabolism |
| 15 | Viral Vaccines | Immunology | | | Genetics |
| 16 | Receptors, Estrogen | Metabolism | | | Pathology |
| 17 | Virus Internalization | | | | Drug Therapy |
| 18 | Luteinizing Hormone | | | | Diagnosis |
| 19 | Orthomyxoviridae Infections | | 46 | Disease Outbreaks | |
| 20 | Antibodies, Viral | Immunology | 47 | Lactation | |

| No. | Candidate | rank$_{DW}$ | rank$_{W2V}$ | # Mentions | Clinical Trial Identifiers |
|---|---|---|---|---|---|
| 1 | Ethanol | 15 | 2,762 | 18,155 | NCT04554433, NCT04554433 |
| 2 | Oxygen | 13 | 1,763 | 90,304 | NCT04842448, NCT04500626, NCT04398290 NCT04327505, NCT04425031, NCT04251871 |
| 3 | Methotrexate | 28 | 2,938 | 5,935 | NCT04352465, NCT04610567 |
| 4 | Calcium | 12 | 1,114 | 45,221 | NCT04379310, NCT04379310 |
| 5 | Hydrogen Peroxide | 17 | 1,476 | 9,252 | NCT04603794, NCT04584684, NCT04723446 NCT04721457, NCT04659928 |
| 6 | Iron | 27 | 1,984 | 31,609 | NCT04643691, NCT04424134, NCT04826822, NCT04345887 |
| 7 | Iodine | 47 | 3,136 | 7,173 | NCT04603794, NCT04941131, NCT04410159, NCT04347954 NCT04344236, NCT04371965, NCT04449965, NCT04478019 NCT04364802, NCT04549376, NCT04446104, NCT04473261 NCT04473261, NCT04510402, NCT04721457, NCT04393792 |
| 8 | Nitric Oxide | 6 | 400 | 21,813 | NCT04388683, NCT04383002, NCT04601077, NCT04460183 NCT04338828, NCT04398290, NCT04358588, NCT04476992 NCT04305457, NCT04312243, NCT04312243, NCT04312243 NCT04337918, NCT04842331, NCT04290858, NCT04306393 |
| 9 | Silver | 18 | 1,126 | 13,328 | NCT04978025 |
| 10 | Estradiol | 48 | 2,932 | 8,455 | NCT04853069, NCT04865029, NCT04359329 |
| 11 | Vitamin D | 21 | 1,267 | 15,063 | NCT04664010, NCT04386850, NCT04709744, NCT04780061 NCT04482673, NCT04411446, NCT04525820 NCT04489628, NCT04621058 |
| 12 | Progesterone | 44 | 2,334 | 8,895 | NCT04365127, NCT04865029 |
| 13 | Metformin | 46 | 2,374 | 6,374 | NCT04604678, NCT04510194, NCT04625985, NCT04626089 |
| 14 | Imatinib | 42 | 2,167 | 2,736 | NCT04394416, NCT04346147, NCT04422678 NCT04953052, NCT04794088 |
| 15 | Selenium | 49 | 2,179 | 4,491 | NCT04869579 |
| 16 | Adenosine | 19 | 421 | 12,099 | NCT04588441 |
| 17 | Resveratrol | 35 | 707 | 3,918 | NCT04400890, NCT04799743 |
| 18 | Zinc | 25 | 289 | 19,993 | NCT04370782, NCT04959786, NCT04446104, NCT04468139 NCT04377646, NCT04447534, NCT04472585 |
| 19 | Sofosbuvir | 38 | 102 | 1,585 | NCT04532931, NCT04535869, NCT04443725, NCT04530422 NCT04561063, NCT04497649, NCT04498936 NCT04460443, NCT04773756 |
| 1 | Amantadine | 240 | 9 | 398 | NCT04952519, NCT04854759, NCT04894617 |
| 2 | Tenofovir Alafenamide | 1,011 | 25 | 210 | NCT04405271 |
| 3 | Nitazoxanide | 395 | 13 | 138 | NCT04523090, NCT04532931, NCT04486313, NCT04348409 NCT04463264, NCT04959786, NCT04343248, NCT04359680 NCT04459286, NCT04746183, NCT04918927, NCT04435314 NCT04441398, NCT04423861, NCT04561063, NCT04552483 NCT04392427, NCT04498936, NCT04729491, NCT04561219 NCT04382846, NCT04788407, NCT04605588 NCT04920838, NCT04406246, NCT04341493 |
| 4 | Danoprevir | 1,540 | 40 | 20 | NCT04345276 |

# Table of Contents

α-modified Random Walk

The number of authors and material nodes in our hypergraphs are not balanced in any of the data sets: (94% vs. 6%) in the energy-related materials data set and (>99.95\% vs. <0.05%) in the drug repurposing data set. Hence, classic random walk with uniform node sampling in each step will result in sequences where author nodes severely outnumber materials. This especially can be seen in drug-disease cases. In order to mitigate this issue, we devised a non-uniform node sampling distribution that can be tuned through a positive parameter denoted by $\alpha$. Depending on the value of $\alpha$, the algorithm samples materials more or less frequently than the authors. This parameter is officially defined as the ratio of the probability of sampling a material (if any) in any given paper to the probability of sampling a non-material node (either author or property nodes). We implemented this algorithm by a mixture of set-wise uniform sampling distributions. Let us denote the set of all nodes existing in the paper (hyperedge) of the $i$-th random walk step as $N_i$. This set can be partitioned into material and non-material nodes denoted by $M_i$ and $A_i$, respectively. While the standard random walk samples the next node $n_{i+1}$ from a uniform distribution over the unpartitioned set of nodes, i.e., $n_{i+1} \sim U(N_i)$, the $\alpha$-modified random walk selects the next node by sampling from the following distribution (assuming both sets $M_i$ and $A_i$ are non-empty):

$$n_{i+1} \sim \frac{1}{\alpha+1} U(A_i) + \frac{\alpha}{\alpha+1} U(M_i)$$

We illustrated the sampling procedure in Supplementary Figure 1. In the first step, one paper $e_1$ is uniformly selected from the set of publications containing the property keyword in their title or abstract. The set of nodes and its two partitions are shown as $N_0$, $M_0$ and $A_0$. With probability $\pi = \frac{1}{\alpha+1}$ the walker picks out the next sample $n_1$ from $A_0$ and with probability $\frac{\alpha}{\alpha+1}$ the next sample will be from $M_0$. As $\alpha$ gets larger the probability of sampling materials, and therefore their frequency in the resulting random walk sequences, increases. In the limit as $\alpha \to \infty$, the walker only samples materials unless the sampled paper does not contain any material nodes, in which case the sampling process is terminated. On the other hand, setting $\alpha=0$ results in author-only sequences.

Multistep Transition Probabilities

A simple way of calculating node similarities in (hyper)graphs is based on transition probabilities. The transition may happen in one or multiple steps. One-step transitions would give non-zero probabilities only to pairs of nodes that have appeared in the same paper. In order to incorporate higher-order connectivities, we consider two- and three-step transitions (Supplementary Figure 2) when computing similarities between properties and other nodes. When predicting discoveries (i.e., materials possessing a targeted property), transitions are considered between the property node and the materials through author nodes (Supplementary Figure 2a-b) and their probabilities are computed as the property-material similarity scores. Moreover, when predicting discoverers (i.e., authors that make discoveries regarding the property), we compute the probability of transitions between the property and the candidate authors through a single material node (Supplementary Figure 2c). These probabilities can be easily formulated via matrix-vector multiplications, as we demonstrate below. However, note that the computation costs can quickly explode and slow down the calculations for transitions with more than three steps even though our hypergraph's vertex matrix is highly sparse.

*Formulation*—Here, we briefly demonstrate the formulation for computing the probability of property-material transitions. Similar formula can be derived for property-author transitions. Let us denote the property node by *P* and the material node by *M*. We considered two- and three-step transitions with intermediate nodes conditioned to belong to the set of authoring experts (denoted by *A*). In each case, the starting node $n_0$ is set to the property node and we compute the probability that a random walker reaches *M* in two or three steps, i.e., $n_2 = M$ or $n_3 = M$, respectively. Therefore, the probability of a two-step transition through an intermediate author node is computed as:

$$Pr(n_2 = M, n_1 \in A \mid n_0 = P) = \sum_{a \in A} Pr(n_2 = M, n_1 = a \mid n_0 = P)$$

$$= \sum_{a \in A} Pr(n_1 = a \mid n_0 = P) \cdot Pr(n_2 = M \mid n_1 = a)$$

where the second line draws on the independence assumptions implied by the Markovian process of random walks. Similar formulation could be derived for three-step transition. The individual transition probabilities in the second line are readily available based on our definition of a hypergraph random walk. For example, for a classic random walk with uniform sampling distribution, we get

$$Pr(n_1 = a \mid n_0 = P) = \frac{1}{d(P)} \sum_{e:\{P,a\} \in e} \frac{1}{d(e)}$$

where $d(P)$ is the degree of node *P*, i.e., the number of hyperedges (papers) it belongs to, and $d(e)$ is the size of hyperedge *e*, i.e., the number of distinct nodes inside it. The first multiplicand in the right-hand side of above equation accounts for selecting a hyperedge that includes *P* and the second computes the probability of selecting *a* from one of the common hyperedeges (if any).

The above computations can be compactly represented and efficiently implemented through matrix multiplication. Let **P** represent the transition probability matrix over all nodes such that $\mathbf{P}_{ij} = Pr(n_1 = j \mid n_0 = i)$. Then, two- and three-step transitions between nodes *P* and *M* could be computed via **P**(*P*,[*A*])·**P**([*A*],*M*) and **P**(*P*,[*A*])·**P**([*A*],[*A*])·**P**([*A*],*M*), respectively, where **P**(*P*,[*A*]) defines selection of the row corresponding to node *P* and columns corresponding to authors in set *A*.

Computational Specifications

The random walk sequences were generated in a high-performance cluster, where each of the compute nodes has a processor Intel E5-2680v4 2.4GHz. This task is highly parallelizable. For each experiment, we generated 250,000 sequences using 40 compute nodes. For electrochemical properties (e.g., thermoelectric), the running time for such a random walk is on the order of 2 hours. On the other hand, these experiments take longer for diseases due to the substantially larger literature database than we used for energy-related properties. For example, generating 250,000 random walk sequences for COVID-19 (i.e., property="coronavirus") took approximately 16 hours on 40 nodes.

The word embedding models were trained on a local workstation with AMD Ryzen Threadripper 2970WX processor with 48 cores and 128GB of RAM. Training a word2vec model (through "gensim" package version 3.8.0 in Python) on 20 CPU cores took about 1 to 2 minutes for both electrochemical properties and diseases.

Study of Hyperparameters

Our predictive pipeline runs with a set of hyperparameters. Among them are the non-uniform sampling distribution parameter (denoted by $\alpha$), the number of years prior to the prediction year considered when building our hypergraph (which we call "memory" here) and the number of high-ranked materials when reporting the predicted materials (denoted by $k$). Here, we study the sensitivity of our algorithm with respect to values of these parameters.

*Non-uniform Sampling Distribution Parameter ($\alpha$)*—We selected two more intermediate values for the $\alpha$ parameter (i.e., 0.5 and 10) and re-ran all the deepwalk experiments with the new values. The resulting precision values in Supplementary Figure 3 show that increasing the value of $\alpha$ beyond the balanced case (i.e., $\alpha>1$—hence, sampling more materials than authors) had less detrimental effects on the inference's precision than making it less than one. In other words, sampling materials more frequently than authors would harm less than sampling authors with higher frequency. Therefore, if one is to break the balance between researching and networking (deviating from $\alpha=1$), overemphasizing research harms less than overemphasizing networking in terms of discovering scientific knowledge. However, even with these perturbations, performance of deepwalk predictions are superior to the word2vec baseline in almost all cases (except for Photovoltaics, when $\alpha=0.5$).

*Number of Predictions ($k$)*—In our main results, we chose $k=50$ materials closest to the property node (in terms of one of the similarity metrics) as the algorithm's predictions. Here, we show the results when we report alternative number of materials (i.e., 25 and 100). Supplementary Figure 4 shows that using smaller number of predictions ($k=25$) generally resulted in higher precision values. However, the order of algorithms in terms of the quality of predictions showed very minor changes. More importantly, the precision margin between our hypergraph-based methods (i.e., deepwalk and transition probability criteria) and the word2vec baselines showed negligible differences among these cases.

*Memory*—We have restricted our framework to the most recent literature prior to the prediction year. The question is how recent we want to stay. Here, we refer to the number of years that we consider before the prediction year as "memory". On one hand, decreasing the memory (hence, ignoring/forgetting all publications produced before that) might unnecessarily deprive us from precious information in the previous years. On the other hand, increasing the memory might bring in too many articles whose authors are not working in the area anymore, hence their experience might have already become irrelevant to the random walker. In order to probe the sensitivity of our algorithm to the memory hyperparameter, we re-ran our discovery prediction experiments on inorganic materials again with memories 2 and 10. Figure S5 shows the resulting cumulative precision curves of these experiments in comparison with the value originally used in our main experiments (i..e., memory=5). In both scenarios, the prediction performance of our deepwalk algorithm slightly decayed from our original experiments, however, in all cases they still remained significantly superior to word2vec.

Experiments with Deep Neural Networks

We also evaluated the effect of incorporating the distribution of expertise in our predictive models after replacing our deepwalk method with graph convolutional neural networks.

Graph Neural Networks (GNNs) have become a popular tool for learning low-dimensional graph representations or solving high-level tasks such as classification of graph nodes [1]. They owe this popularity to their unique and efficient way of exploiting graph connectivities to propagate information between a central node and its neighborhood, their ability to incorporate feature vectors for nodes and/or edges, and their superior generalization to unseen (sub)graphs. Similar to deepwalk, these models build a low-dimensional embedding space where graph-based similarities are preserved. However, unlike deepwalk, they incorporate node feature vectors and directly utilize graph connectivities for message passing between nearby nodes when constructing the embedding space.

The embedding vector of a central node is constructed by sequentially processing messages passed from its local neighbors. There are numerous ways of aggregating the signals reaching out from neighbors. In our experiments, we use the Graph Sample and Aggregate (GraphSAGE) platform, which applies the aggregations function on a subset of neighbors to avoid computational overhead [2]. Let $h_i^l$ denotes the message from the $i$-th node in the $l$-th step of this sequential procedure. Then, the representation of the $i$-th node at the next level will be computed as

$$h_i^{l+1} = \sigma(f_{AGG}(\{h_j^l\}_{j \in \tilde{N}(i)})W_l)$$

where $f_{AGG}$ is an aggregation function (e.g., mean, pooling, etc) applied on the concatenation of the local neighborhood's messages $\{h_j^l\}_{j \in \tilde{N}(i)}$, and $\tilde{N}(i)$ is a subset of $k_l$ uniformly sampled nodes from the immediate neighbors of the $i$-th node $N(i)$. The resulting aggregated messages will undergo a single-layer neural network parameterized by $W_l$ (the bias term is ignored for simplicity) and the non-linear activation σ. The input messages in the first step, i.e., $h_i^0 \forall i$, are set to the provided node feature vectors. The final representation of the $i$-th node will be reached after $L$ steps. We used the same set of hyperparameters as the original paper [2]; we considered two steps ($L=2$) with samples sizes $k_1=25$ and $k_2=10$. We also used the mean aggregation function and applied the non-linearity through Rectified Linear Unit (ReLU) activation.

We approached the discovery prediction problem in an unsupervised manner through a graph autoencoder [3], where the encoder component was modeled using the GraphSAGE architecture and the decoder component consisted of a parameter-less inner-product of the encoder's output. This autoencoder was trained by minimizing a link-prediction loss function, which was approximated with negative sampling. The approximate loss has two parts accounting for the similarity of positive samples (pairs of nearby nodes) and the dissimilarity of negative samples (pairs of unconnected nodes).

Our mechanism of sampling positive and negative pairs closely resembled that which was used in deepwalk: the former is formed by pairing central/contextual nodes within windows sliding over short random walks, and the latter by means of sampling from the unigram distribution raised to power 3/4 over the full set of nodes [4]. Once positive samples were drawn using a sliding window size of 8, we began minimizing the loss function in a mini-batch setting by iterating over pairs. We used a batch size of 1000, negative sampling size of 15 (per positive pair), learning rate of $5 \times 10^{-6}$ and the Adam optimizer [5] with default parameters.

We trained our graph autoencoder in two different settings: (i) using our full hypergraph, and (ii) after dropping author nodes. In both settings, we only considered the material and property nodes. In the full setting, we took account of author nodes when drawing positive samples and computing the adjacency matrix. In this setting, the positive samples were drawn from deepwalk sequences associated with $\alpha = 1$, whereas the experiment without authors used sequences generated after excluding the author nodes ($\alpha \to \infty$). Moreover, connectivities between nodes were different for the two settings. In the author-less network, we connected two property or material nodes only if they appeared in the same paper. In the full setting, we kept these connections and added more edges between nodes with at least one common author neighbor even in the absence of papers in which they co-occur.

Combining Scores for Complementary Predictions

We combine two sources of information for measuring human availability and scientific plausibility, each of which separately scores the candidate materials. The two scores will then be combined through $\beta$ to result in a single scalar score such that the magnitude of $\beta$ determines the weight of whether human availability is encouraged (negative) or discouraged (positive) among predictions, as opposed to scientific plausibility (0). As $\beta$ grows from one extreme (–1) to the other (+1) the algorithm lessens its objective of human accessibility, instead avoiding it after passing $\beta = 0$. Hence, at least three special operating points are distinguishable based on the value of $\beta$:

- ($\beta = -1$): scientific plausability is given zero attention and full weight lies on finding hypotheses near human reasoners, i.e., maximizing the likelihood that the algorithm's predictions will be discovered by human scientists in the near future.
- ($\beta = 0$): scientific plausibility is given full emphasis and human availability of predictions is ignored.
- ($\beta = +1$): scientific plausibility is ignored and the full weight rests on human avoidance, i.e., maximizing the likelihood that predictions will escape human scientists' collective attention.

In addition to these special conditions, we desire an algorithm where contribution from the two sources (plausibility and human accessibility) become equal when $|\beta| = 1/2$, and the output score varies continuously as $\beta$ shifts.

For any given candidate material *x*, let us denote its shortest-path distance (SPD) and its semantic similarity with respect to the property node by $s_1(x)$ and $s_2(x)$, respectively. These scores have distinct units and vary in different scales, therefore a naive *β*-weighted averaging is inappropriate as it does not lead to equal contribution when $|\beta|=1/2$. Moreover, the SPD values are unbounded as they can become arbitrarily large for entities disconnected from the property node in our hypergraph. As a result, Z-scores could not be directly applied either. Instead, we applied a Van der Waerden transformation to first standardize the scores. Suppose *S* is a set of scores and $s(x) \in S$, then its Van der Waerden transformation of *x*, denoted by $\tilde{s}(x)$, is defined as

$$\tilde{s}(x) = \varphi\left(\frac{r(x)}{|S|+1}\right)$$

where *φ* is the quantile function of the normal distribution, *r(x)* is the rank of *s(x)* within the set *S* and |*S*| denotes the cardinality of *S*. We then take the weighted average of Z-scores for the transformed signals $\tilde{s}_1(x)$ and $\tilde{s}_2(x)$ for each material *x* as the ultimate hybrid score to be used in our final ranking. We further normalize the resulting transformed scores by computing their Z-scores yielding $\hat{s}_1(x)$ and $\hat{s}_2(x)$ before combining them in the following weighted averaging:

$$s_{final}(x) = \beta\,\hat{s}_1(x) + (1 - |\beta|)\,\hat{s}_2(x)$$

Note that when *β*<0, the algorithm tends to select materials with lower $\hat{s}_1(x)$, which in turn implies smaller SPD and materials with more contextual familiarity in terms of property. Alternatively, when *β*>0, the algorithm scores higher those materials with a greater SPD and more unfamiliar or "alien" predictions will result.

Evaluating the Alien Prediction Algorithm

An ideal alien prediction algorithm generates discovery candidates that are complementary to those published by human scientists, i.e., unimaginable yet plausible candidates. Hence, running our alien AI at a certain prediction year $y_{pred}$, the majority of predictions are expected not to be discovered by human scientists in the following years. However, for the few cases that are actually discovered in reality, we show that the average year of their discoveries tends to grow, thus becoming less imaginable to those at the present moment as alienness (*β*) increases. In order to assess the full set of predictions that are dominated by undiscovered predictions for large *β* values, we evaluate their scientific plausibility as the likelihood that they are potential disruptive discoveries. For this purpose, we conservatively rely on the simulative and theoretical prior knowledge that exist in the field (see next section). Specifically, we show that while the percentage of predictions that become discovered after the prediction year sharply decays with *β*, the quality of the predictions remains high for longer. This observation, which we will measure by a criterion named the *expectations gap*, implies the existence of intervals of *β* where our alien AI approaches an ideal alien predictor by producing plausible undiscoverable predictions. Additionally, we explicitly estimated the joint probability of being plausible and undiscoverable at the same time to identify those intervals of *β* where the our alien algorithm approaches the ideal operation point.

*Expectation Gap*—Our goal is to define a single scalar score indicating whether our alien predictor is capable of generating unimaginabled yet promising predictions. We define two distributions over $\beta$ conditioned on (1) discoverability and (2) plausiability of predictions. The separation between these two distributions such that the latter is more concentrated on larger $\beta$ values than the former will be our indicator, which will be quantified by the gap between their means (hence expectation gap).

For a fixed property and pool of candidate materials, let us denote the set of all materials that will be discovered after $y_{pred}$ by $D$ and the set of all plausible materials by $P$. Also, let $H_\beta$ and $h_\beta$ be the full set of predictions and a randomly selected prediction generated by our alien algorithm operating with $\beta$, respectively. Precision of the algorithm in terms of identifying near-future discoveries is defined as $Pr(h_\beta \in D|\beta)$, which can be simply computed by dividing the count of discoveries by the number of predictions: $|H_\beta \cap D|/|H_\beta|$. Now using a uniform prior distribution over $\beta$, i.e., $Pr(\beta)$=const., and applying Bayes rule these precisions can be converted to $Pr(\beta|h_\beta \in D)$ by normalization across all $\beta$ values such that they sum to unity.

Computing the second distribution is not as straightforward. The subtlety arises from the fact that we do not fully know $P$, instead we have one real-valued score per material characterizing the likelihood of its $P$-membership. These scores, denoted by $\tau$, are obtained from field-related theoretical knowledge and first-principles laws (see next section). In the first step, we transform the theoretical scores to probabilities, such that for every material $x$, $\tau=\tau(x)$ goes to $Pr(x \in P)$. Let $\tau_{min}$ and $\tau_{max}$ be the global minimum and maximum of all the theoretical scores at hand. We engineered a monotonically increasing transformation $T$ in the form of $\text{logit}[\tan(\pi(\hat{\tau} - \frac{1}{2})+b)]$, where $\hat{\tau} = (\tau - \tau_{min})/(\tau_{max} - \tau_{min})$ such that

- $T(\tau_{min}) = 0$

- $T(\tau_{max}) = 1$

- $T(\tau_{mid}) = \frac{1}{2}$, where $\tau_{mid} = \frac{1}{|D|} \sum_{x \in D} \tau(x)$, which is the average of theoretical scores attributed to discovered materials. This condition uniquely specifies parameter $b$.

The resulting probabilities will be thresholded by 1/2 to probabilistically indicate which materials belong to $P$. Setting the midpoint as above is a direct consequence of the assumption that the majority of the materials discovered by the scientists to have the targeted properties are plausible findings, hence we take their average theoretical scores as a baseline and every material with a higher $\tau$ will also be considered as plausible. Such probabilistic classification of a material $x$ to $P$ is done with a confidence level proportional to the distance between the probability $T(\tau(x))$ and the threshold 1/2. The confidence level of our decision regarding $P$-membership of sample $x$ is

$$c(x) = \begin{cases} T(\tau(x)) & \text{, if } T(\tau(x)) \geq 1/2 \\ 1-T(\tau(x)) & \text{, if } T(\tau(x)) < 1/2 \end{cases}$$

Now, for any prediction set $H_\beta$, we use weighted maximum likelihood estimation to compute the probability of being plausible given $\beta$:

$$Pr(h_\beta \in P|\beta) = \sum_{x \in H_\beta : T(\tau(x)) \geq 1/2} c(x) \Big/ \sum_{x \in H_\beta} c(x)$$

Finally, similar to the previous case, the likelihood of $\beta$ given plausibility, $Pr(\beta|h_\beta \in P)$, can be obtained by simply normalizing these probabilities across all $\beta$ values such that they sum to one.

The expectation gap is defined as the difference between the mean values of the two likelihoods described above:

$$\Delta E[\beta] := E[\beta|h_\beta \in P] - E[\beta|h_\beta \in D]$$

Having a positive gap suggests that theoretical plausibility is higher for more alien predictions than for those where the predictions are made and published. Therefore, because the discovery precision goes down with $\beta$, a positive expectation gap also means that there exists a non-empty gap when the alien algorithm approaches its ideal mission of human complementarity. Zero or negative gaps only occurred for a few human diseases.

*Joint Probabilities*—Our expectation gap provided us with a single evaluation score for the performance of the alien prediction. However, it does not say anything about the desired range of $\beta$ where the algorithm operates closest to its mission to serve complementary, high-value predictions. Here, to provide a clearer overview of the performance of our alien prediction for different $\beta$ values, we directly model and calculate the probability that it outputs unfamiliar yet scientifically promising (plausible) predictions. As is also described above, unfamiliarity of a random prediction $h_\beta$ means its unimaginability in context of the considered property and therefore its undiscoverability (i.e., $h_\beta \notin D$). Thus, we calculate the probability that $h_\beta$ is unfamiliar and plausible given a certain $\beta$ by the joint distribution $Pr(h_\beta \notin D, h_\beta \in P|\beta)$. Applying Bayes rule, this joint probability decomposes into two simpler distributions:

$$Pr(h_\beta \notin D, h_\beta \in P|\beta) = Pr(h_\beta \notin D|\beta) \, Pr(h_\beta \in P|\beta, h_\beta \notin D)$$

where the first term in the right-hand-side is the complementary of discovery precision (i.e., 1-$Pr(h_\beta \in D|\beta)$) and the second term can be computed similarly to the probability of plausibility, $Pr(h_\beta \in P|\beta)$, described above, except that all computations are to be repeated on the predictions that are *not* discovered after the prediction year ($h_\beta \notin D$), hence replacing $H_\beta$ with $H_\beta$-$D$ when doing the weighted maximum likelihood above.

Theoretical Scoring of Candidate Hypotheses

In order to assess the quality of predictions resulting from our algorithm, we used theoretically driven scores derived from first-principles equations or simulation models from the relevant disciplines. The underlying procedure for defining and curating such data naturally differs for distinct properties. In this section, we describe these theoretical scores for properties about which datasets could be found.

*Thermoelectricity*—Thermoelectricity is the property of producing electrical voltage when temperature varies on two sides of a material. An important measure of thermoelectricity that depends on the Seebeck coefficient and electrical conductivity is called Power Factor (PF). PF is also the major part of another common dimensionless criterion named its *figure of merit (zT)*. First-principles methods based on Density Functional Theory (DFT) have been widely used to estimate energy-related chemical properties including PF[6]. Recently, DFT-based PF estimates have been used to evaluate content-based discovery predictions[7], where it is shown that materials studied in connection with thermoelectricity have larger PF values on average than unstudied materials. Here, we use the same set of PF estimates, which have been prepared by taking the maximum of the average PFs computed for various temperatures, doping levels and semiconductor types. Among 12.6K materials that were mentioned at least once before the prediction year of 2001 (i.e., within the five-year period [1996,2000]), only about 3.5K materials were assigned a PF estimate.

*Ferroelectricity*—Ferroelectric materials are characterized by their spontaneous electric polarization, which is reversible in the presence of an external electric field. Smidt et al. developed an automated workflow that uses symmetry analysis and first-principles calculations to curate a list of ferroelectric materials[8]. Their final list included 255 candidates, where entries are rated based on the magnitude of their spontaneous polarization. There are 167 distinct chemical formulae in this list, and for some several ferroelectricity scores are reported corresponding to various crystal structures. For compounds with multiple values, we considered the maximum computed ferroelectricity over all available structures as the final score. Among the materials mentioned in the most recent five-year period preceding the prediction year of 2001, only 167 distinct compounds that have been reported as ferroelectric by Smidt et al. are assigned non-zero scores and the rest are considered to have zero ferroelectricity. This leads to a sparse scoring system but, as is shown in our results, remains nevertheless sufficient to demonstrate the performance of our method.

*Human Diseases*—The similarity between two sets of proteins targeted by a particular disease and a certain drug forms a basis for measuring their underlying association and therapeutic potential. Protein-protein interaction networks include protein, drug and disease nodes with pairwise interactions encoded within the edges. We utilize drug-disease proximity in such networks as the core of our theoretical scoring framework to assess drug candidates in terms of their relevance to the treatment of a disease. Recently, Gysi et al. showed that drugs whose target proteins are within or in vicinity of the COVID-19 disease module are potentially strong candidates for repurposing to treat or prevent the infection[9]. They used various proximity measures to compute network similarities and identify the most relevant drug candidates. Among these measures was cosine similarity of embedding vectors resulting from a pretrained Graph Neural Network (GNN) over the protein-protein network, which we used in our evaluations of drug-disease proposals.

The GNN-based embedding vectors of the interaction network produced and shared by Gysi et al.[9] included 1.6K drug nodes and 2.5K disease nodes (including COVID-19), among which 1.5K drugs belonged to our pool of candidates and 176 diseases were common with the set of 400 human diseases we considered in our expert-avoiding experiments. We ran our complementary algorithm at prediction year 2001 for all diseases but COVID-19, where we restricted our pool of candidates to 1,179 drugs that existed in the protein-protein network and also appeared in the literature between 1996 and 2000. For COVID-19, the prediction year is set to the beginning of 2020 with 1,436 candidate drugs that existed in both the interaction network and in the literature between 2015 and 2019.

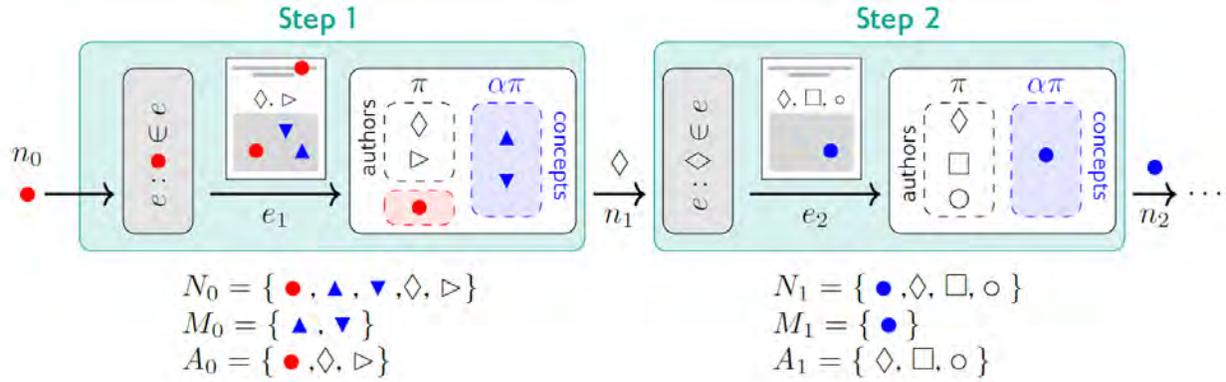

$N_0 = \{\bullet, \blacktriangle, \blacktriangledown, \Diamond, \triangleright\}$
$M_0 = \{\blacktriangle, \blacktriangledown\}$
$A_0 = \{\bullet, \Diamond, \triangleright\}$

$N_1 = \{\bullet, \Diamond, \square, \circ\}$
$M_1 = \{\bullet\}$
$A_1 = \{\Diamond, \square, \circ\}$

**Supplementary Figure 1.**
Two steps in an α-modified random walk. Blank shapes represent author nodes and colored shapes represent materials (blue) and the property (red). Papers (hyperedges) are sampled uniformly, whereas nodes are selected such that the probability of sampling a material node is α times the probability of sampling a non-material node. Here, π denotes the probability of sampling non-material nodes, which is uniquely determined by α itself. In each random walk step, the selected hyperedge is shown over the arrow ($e_1$ or $e_2$) and the hypernodes that it contains are listed below the figure ($N_0$ or $N_1$), which are in turn partitioned into material ($M_0$ or $M_1$) and non-material ($A_0$ or $A_1$) subsets. The output of the considered step will be a random draw from these hypernodes ($n_1 \in N_0$ or $n_2 \in N_1$).

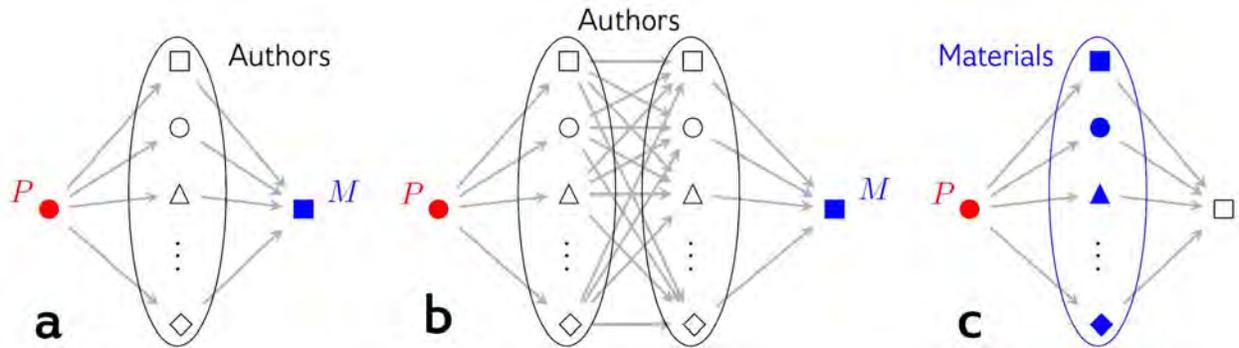

**Supplementary Figure 2.**
Using transition probabilities as the similarity criterion in our inferences. **(a)** Two-step transitions ($s=2$) between the property node $P$ and a particular material $M$ through a single intermediate author node. **(b)** Three-step transitions between ($s=3$) between the property node $P$ and a particular material $M$ through two intermediate author nodes. **(c)** Two-step transitions between property $P$ and a particular author (uncolored square at the right) through a single material node. The probability of such property-author transitions is taken as the similarity between author and property nodes, which in turn, is used for selecting $k=50$ authors that are most similar to the property and reporting them as the potential discoverers of materials possessing property $P$.

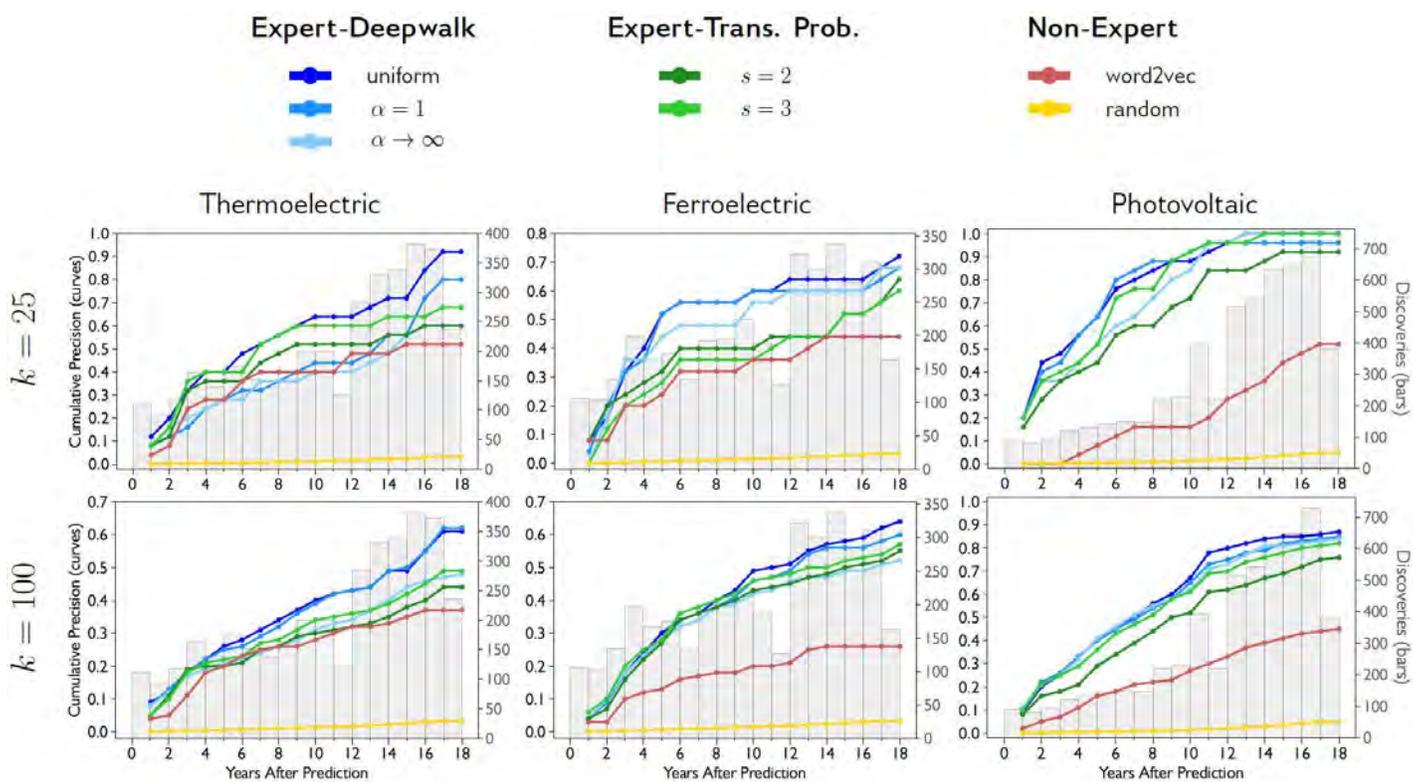

**Supplementary Figure 3.**
Precision of predicting energy-related properties for inorganic materials when different numbers of materials are reported as the predictions ($k$=25 and 100).

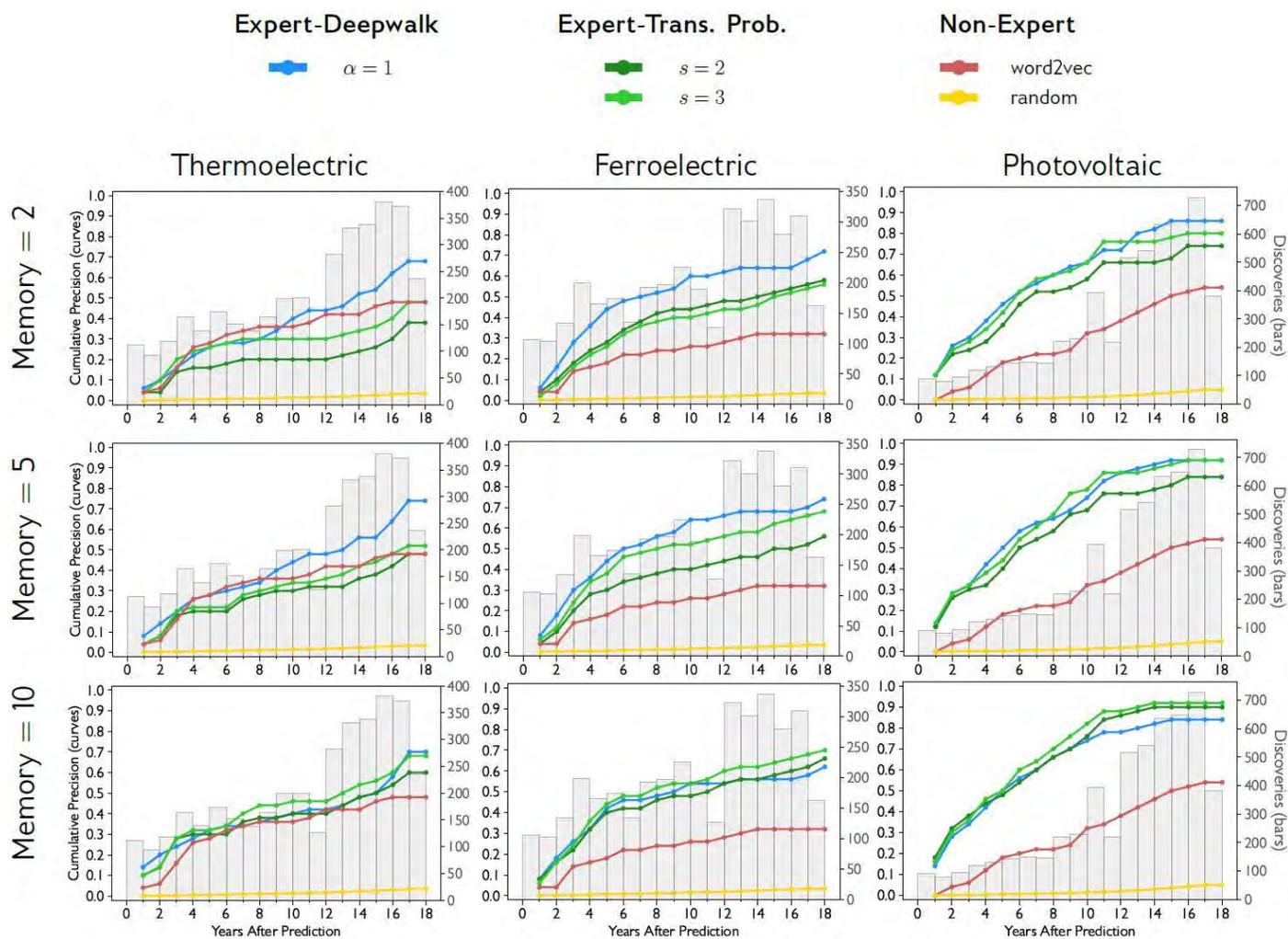

**Supplementary Figure 4.**
Precision of predicting energy-related properties for inorganic materials for different memory values (i.e., the number of years considered prior to the prediction year when constructing our hypergraph).

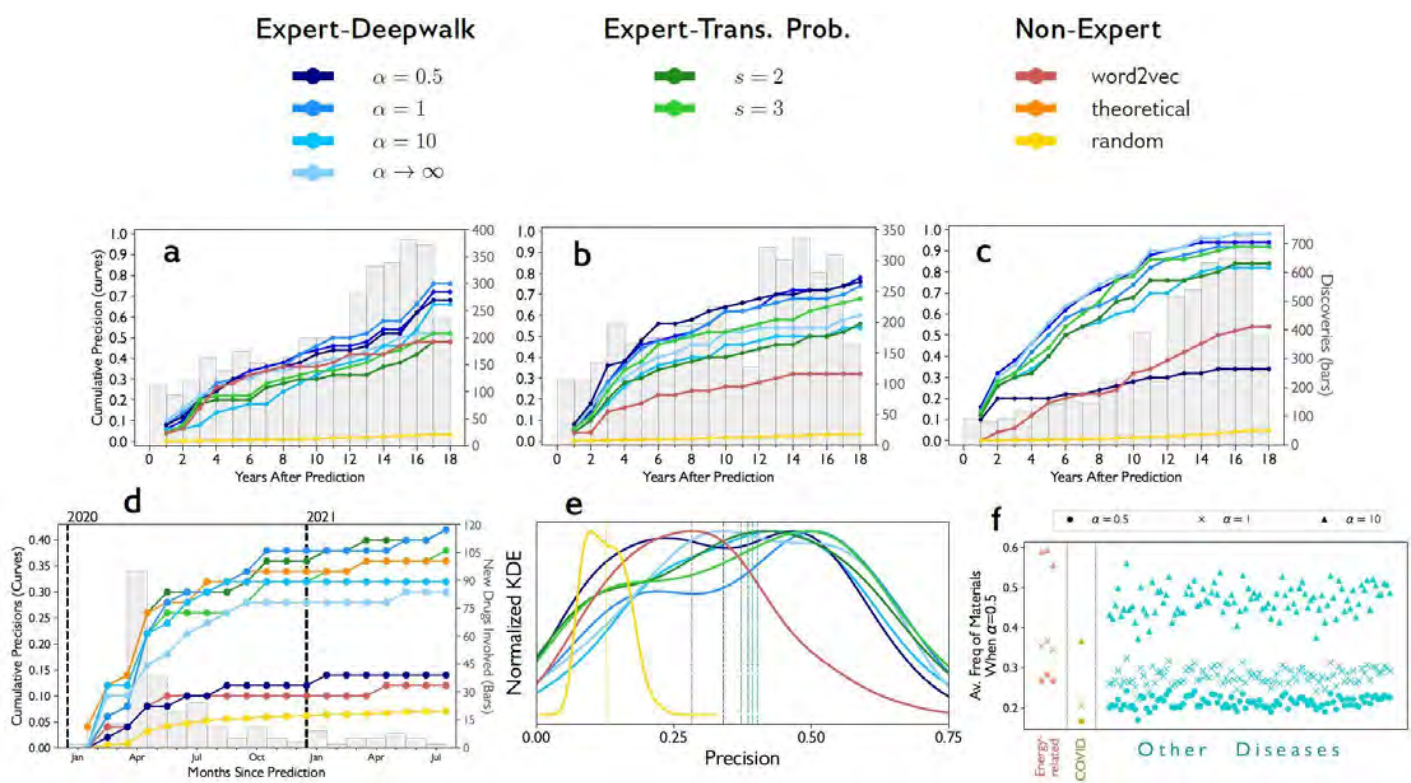

**Supplementary Figure 5.**
Precision of discovery predicting with more intermediate values for α parameter. Subfigures **(a-d)** show cumulative precision curves for properties associated with being **(a)** thermoelectric, **(b)** ferroelectric, **(c)** photovoltaic and **(d)** effective in therapy of or vaccination against COVID-19. Subfigure **(e)** shows the KDE-estimated distribution of precision values for inferring materials that are useful for therapy of 100 diseases in CTD, done by different algorithms. Finally, subfigure **(f)** indicates average frequency of materials in random-walk sequences generated with three different intermediate values for α. In all cases, whereas performance of our deepwalk algorithm with α=10 did not drastically change from α=1, one can see a significant drop in precision when α=0.5 is used.

**Supplementary Table 1.** Evaluating predicted discoveries by our complementary discovery prediction algorithms operated with $\beta$ values 0, 0.2 and 0.4.

| Disease | $\beta = 0$ | | $\beta = 0.2$ | | $\beta = 0.4$ | |
|---|---|---|---|---|---|---|
| | Overlap (%) | Pr-Pr Sim. | Overlap (%) | Pr-Pr Sim. | Overlap (%) | Pr-Pr Sim. |
| Acromegaly | 0 | 0.323 | 0 | 0.32 | 0 | 0.259 |
| Adenocarcinoma | 11 | 0.066 | 10 | 0.009 | 13 | -0.133 |
| Alcoholism | 23 | 0.565 | 19 | 0.539 | 15 | 0.475 |
| Alopecia | 37 | 0.152 | 26 | 0.165 | 25 | 0.2 |
| Amenorrhea | 3 | 0.264 | 3 | 0.268 | 1 | 0.273 |
| Amyloidosis | 22 | 0.269 | 18 | 0.257 | 16 | 0.253 |
| Anaphylaxis | 4 | 0.367 | 3 | 0.369 | 1 | 0.335 |
| Anemia | 22 | 0.155 | 21 | 0.15 | 22 | 0.168 |
| Angioedema | 11 | 0.183 | 10 | 0.205 | 5 | 0.225 |
| Appendicitis | 19 | 0.281 | 13 | 0.222 | 8 | 0.145 |
| Arteriosclerosis | 41 | 0.441 | 35 | 0.423 | 28 | 0.406 |
| Arthritis | 13 | 0.309 | 9 | 0.276 | 6 | 0.129 |
| Arthrogryposis | 2 | 0.363 | 2 | 0.364 | 1 | 0.305 |
| Ascites | 6 | 0.142 | 5 | 0.176 | 1 | 0.245 |
| Asthma | 10 | 0.351 | 9 | 0.329 | 16 | 0.243 |
| Astrocytoma | 39 | 0.169 | 32 | 0.165 | 25 | 0.111 |
| Ataxia | 23 | 0.423 | 17 | 0.405 | 12 | 0.333 |
| Atherosclerosis | 33 | 0.387 | 35 | 0.4 | 32 | 0.281 |
| Blepharospasm | 2 | 0.305 | 1 | 0.259 | 1 | 0.14 |
| Blindness | 0 | 0.219 | 1 | 0.212 | 1 | 0.208 |
| Brachydactyly | 4 | 0.238 | 4 | 0.164 | 1 | 0.036 |
| Bronchiectasis | 36 | 0.408 | 34 | 0.387 | 20 | 0.341 |
| Brucellosis | 2 | 0.288 | 2 | 0.215 | 1 | 0.065 |
| Candidiasis | 6 | 0.225 | 5 | 0.155 | 0 | 0.069 |
| Carcinoma | 15 | 0.163 | 12 | 0.086 | 10 | -0.176 |
| Cardiomyopathies | 15 | 0.393 | 15 | 0.379 | 11 | 0.383 |
| Cataract | 15 | 0.327 | 12 | 0.327 | 9 | 0.314 |
| Cholangiocarcinoma | 48 | 0.213 | 42 | 0.168 | 30 | 0.001 |
| Cholangitis | 10 | 0.217 | 8 | 0.164 | 6 | 0.097 |
| Cholelithiasis | 6 | 0.22 | 6 | 0.171 | 5 | 0.106 |

**Supplementary Table 1.** *(continued)*

| Disease | $\beta = 0$ | | $\beta = 0.2$ | | $\beta = 0.4$ | |
|---|---|---|---|---|---|---|
| | Overlap (%) | Pr-Pr Sim. | Overlap (%) | Pr-Pr Sim. | Overlap (%) | Pr-Pr Sim. |
| Cholelithiasis | 6 | 0.22 | 6 | 0.171 | 5 | 0.106 |
| Cholestasis | 23 | 0.255 | 26 | 0.241 | 32 | 0.239 |
| Chorioamnionitis | 17 | 0.274 | 13 | 0.236 | 11 | 0.194 |
| Colitis | 41 | 0.256 | 36 | 0.262 | 31 | 0.223 |
| Constipation | 19 | 0.334 | 16 | 0.329 | 10 | 0.304 |
| Contracture | 0 | 0.408 | 0 | 0.362 | 0 | 0.329 |
| Cryptorchidism | 43 | 0.35 | 35 | 0.337 | 17 | 0.26 |
| Cystitis | 11 | 0.361 | 12 | 0.362 | 9 | 0.283 |
| Delirium | 10 | 0.378 | 10 | 0.36 | 8 | 0.296 |
| Dermatitis | 17 | 0.311 | 13 | 0.252 | 8 | 0.206 |
| Dermatomyositis | 37 | 0.216 | 30 | 0.224 | 21 | 0.214 |
| Diarrhea | 8 | 0.267 | 10 | 0.283 | 8 | 0.251 |
| Dizziness | 6 | 0.551 | 6 | 0.535 | 3 | 0.433 |
| Dysarthria | 6 | 0.359 | 4 | 0.315 | 2 | 0.277 |
| Dyskinesias | 35 | 0.623 | 34 | 0.61 | 25 | 0.518 |
| Dyslipidemias | 35 | 0.377 | 28 | 0.356 | 14 | 0.26 |
| Dyspnea | 13 | 0.4 | 10 | 0.43 | 9 | 0.425 |
| Eczema | 7 | 0.225 | 6 | 0.215 | 6 | 0.126 |
| Embolism | 3 | 0.246 | 2 | 0.226 | 2 | 0.202 |
| Emphysema | 22 | 0.378 | 21 | 0.33 | 17 | 0.312 |
| Endometriosis | 55 | 0.327 | 46 | 0.314 | 39 | 0.195 |
| Entamoebiasis | 31 | 0.022 | 24 | -0.05 | 12 | -0.11 |
| Enterocolitis | 12 | 0.092 | 13 | 0.1 | 9 | 0.107 |
| Eosinophilia | 9 | 0.24 | 7 | 0.215 | 6 | 0.136 |
| Epilepsy | 17 | 0.457 | 14 | 0.436 | 15 | 0.331 |
| Exanthema | 11 | 0.108 | 10 | 0.092 | 9 | 0.083 |
| Gallstones | 4 | 0.216 | 2 | 0.181 | 1 | 0.101 |
| Gastroenteritis | 3 | -0.081 | 2 | -0.069 | 1 | -0.006 |
| Gastroparesis | 15 | 0.18 | 12 | 0.18 | 9 | 0.186 |
| Glaucoma | 14 | 0.358 | 10 | 0.383 | 8 | 0.33 |

**Supplementary Table 1.** *(continued)*

| Disease | β = 0 | | β = 0.2 | | β = 0.4 | |
|---|---|---|---|---|---|---|
| | Overlap (%) | Pr-Pr Sim. | Overlap (%) | Pr-Pr Sim. | Overlap (%) | Pr-Pr Sim. |
| Glioblastoma | 33 | 0.336 | 25 | 0.301 | 13 | 0.143 |
| Glioma | 35 | 0.223 | 30 | 0.152 | 28 | -0.002 |
| Gliosarcoma | 20 | 0.18 | 18 | 0.148 | 12 | 0.122 |
| Glomerulonephritis | 9 | 0.226 | 7 | 0.206 | 5 | 0.176 |
| Goiter | 3 | 0.009 | 2 | -0.011 | 2 | -0.002 |
| Gout | 39 | 0.327 | 35 | 0.286 | 27 | 0.264 |
| Heartburn | 3 | 0.159 | 3 | 0.147 | 2 | 0.142 |
| Hemangioblastoma | 17 | -0.079 | 13 | -0.1 | 5 | -0.135 |
| Hemangioma | 5 | 0.172 | 5 | 0.174 | 3 | 0.131 |
| Hemorrhage | 4 | 0.263 | 4 | 0.263 | 3 | 0.233 |
| Hemorrhoids | 3 | 0.198 | 3 | 0.188 | 2 | 0.205 |
| Hyperaldosteronism | 6 | 0.429 | 4 | 0.418 | 2 | 0.398 |
| Hyperalgesia | 21 | 0.522 | 20 | 0.51 | 19 | 0.432 |
| Hyperammonemia | 4 | 0.206 | 4 | 0.21 | 3 | 0.205 |
| Hypercholesterolemia | 25 | 0.383 | 24 | 0.377 | 19 | 0.295 |
| Hyperinsulinism | 10 | 0.28 | 10 | 0.271 | 7 | 0.245 |
| Hyperkalemia | 8 | 0.182 | 5 | 0.171 | 5 | 0.183 |
| Hyperlipidemias | 32 | 0.423 | 27 | 0.397 | 13 | 0.277 |
| Hyperparathyroidism | 13 | 0.243 | 12 | 0.245 | 11 | 0.191 |
| Hyperphosphatemia | 7 | 0.413 | 6 | 0.408 | 4 | 0.376 |
| Hyperpigmentation | 11 | 0.234 | 10 | 0.224 | 7 | 0.198 |
| Hyperplasia | 34 | 0.165 | 35 | 0.172 | 31 | 0.198 |
| Hyperprolactinemia | 2 | 0.206 | 1 | 0.175 | 0 | 0.11 |
| Hypersensitivity | 19 | 0.23 | 20 | 0.22 | 24 | 0.189 |
| Hypertension | 3 | 0.553 | 5 | 0.49 | 9 | 0.361 |
| Hyperthyroidism | 18 | 0.04 | 16 | 0.039 | 16 | 0.027 |
| Hypertrophy | 41 | 0.394 | 38 | 0.394 | 25 | 0.32 |
| Hyperuricemia | 22 | 0.317 | 17 | 0.35 | 11 | 0.262 |
| Hypoalbuminemia | 19 | 0.185 | 18 | 0.175 | 14 | 0.124 |
| Hypocalcemia | 5 | 0.232 | 3 | 0.242 | 2 | 0.227 |

**Supplementary Table 1.** *(continued)*

| Disease | $\beta = 0$ | | $\beta = 0.2$ | | $\beta = 0.4$ | |
|---|---|---|---|---|---|---|
| | Overlap (%) | Pr-Pr Sim. | Overlap (%) | Pr-Pr Sim. | Overlap (%) | Pr-Pr Sim. |
| Hypoglycemia | 4 | 0.228 | 3 | 0.205 | 3 | 0.255 |
| Hypogonadism | 5 | 0.213 | 8 | 0.169 | 6 | 0.095 |
| Hypokalemia | 4 | 0.396 | 3 | 0.387 | 2 | 0.339 |
| Hyponatremia | 7 | 0.289 | 6 | 0.264 | 6 | 0.241 |
| Hypoparathyroidism | 2 | 0.217 | 2 | 0.243 | 1 | 0.177 |
| Hypophosphatasia | 16 | -0.133 | 13 | -0.124 | 7 | -0.102 |
| Hypophosphatemia | 8 | 0.254 | 8 | 0.23 | 5 | 0.172 |
| Hypopigmentation | 3 | 0.091 | 3 | 0.103 | 2 | 0.175 |
| Hypopituitarism | 16 | 0.218 | 15 | 0.206 | 13 | 0.196 |
| Hypotension | 6 | 0.52 | 6 | 0.495 | 8 | 0.423 |
| Hypothyroidism | 7 | 0.235 | 3 | 0.208 | 2 | 0.188 |
| Hypotrichosis | 3 | 0.284 | 1 | 0.263 | 0 | 0.207 |
| Hypoxia | 12 | 0.361 | 10 | 0.373 | 8 | 0.332 |
| Keloid | 37 | 0.291 | 34 | 0.268 | 20 | 0.175 |
| Keratitis | 22 | 0.27 | 13 | 0.193 | 7 | 0.138 |
| Keratosis | 38 | 0.152 | 31 | 0.131 | 15 | 0.134 |
| Leiomyoma | 21 | 0.141 | 17 | 0.104 | 12 | -0.002 |
| Leiomyosarcoma | 30 | 0.219 | 28 | 0.126 | 18 | -0.008 |
| Leishmaniasis | 52 | 0.305 | 43 | 0.249 | 28 | 0.165 |
| Leprosy | 3 | 0.089 | 2 | 0.065 | 1 | 0.021 |
| Leukemia | 21 | 0.173 | 16 | 0.111 | 9 | 0.023 |
| Liposarcoma | 14 | 0.145 | 12 | 0.079 | 4 | -0.119 |
| Listeriosis | 30 | 0.274 | 26 | 0.219 | 17 | 0.012 |
| Lymphangioleiomyomatosis | 10 | 0.028 | 8 | -0.014 | 2 | -0.144 |
| Lymphoma | 16 | 0.29 | 17 | 0.224 | 15 | 0.064 |
| Malaria | 7 | 0.241 | 4 | 0.22 | 0 | 0.129 |
| Malnutrition | 8 | 0.078 | 6 | 0.066 | 4 | 0.05 |
| Medulloblastoma | 35 | 0.291 | 28 | 0.224 | 20 | 0.093 |
| Melanoma | 24 | 0.257 | 19 | 0.232 | 15 | 0.145 |
| Meningitis | 3 | 0.169 | 3 | 0.105 | 1 | 0.042 |

**Supplementary Table 1.** *(continued)*

| Disease | $\beta = 0$ | | $\beta = 0.2$ | | $\beta = 0.4$ | |
|---|---|---|---|---|---|---|
| | Overlap (%) | Pr-Pr Sim. | Overlap (%) | Pr-Pr Sim. | Overlap (%) | Pr-Pr Sim. |
| Methemoglobinemia | 16 | 0.196 | 16 | 0.184 | 12 | 0.116 |
| Mucositis | 17 | 0.159 | 18 | 0.125 | 15 | 0.038 |
| Myoclonus | 15 | 0.647 | 12 | 0.627 | 10 | 0.515 |
| Myositis | 2 | 0.261 | 3 | 0.223 | 1 | 0.247 |
| Narcolepsy | 11 | 0.504 | 10 | 0.486 | 5 | 0.352 |
| Neoplasms | 23 | 0.2 | 26 | 0.09 | 24 | 0.006 |
| Nephritis | 9 | 0.219 | 11 | 0.226 | 4 | 0.178 |
| Nephrolithiasis | 4 | 0.415 | 6 | 0.4 | 2 | 0.332 |
| Nephrosis | 6 | 0.328 | 5 | 0.339 | 5 | 0.273 |
| Neuroblastoma | 13 | 0.253 | 11 | 0.267 | 10 | 0.096 |
| Neurofibrosarcoma | 3 | 0.151 | 2 | 0.11 | 2 | -0.035 |
| Neutropenia | 11 | 0.228 | 9 | 0.171 | 8 | 0.089 |
| Ophthalmoplegia | 3 | 0.147 | 3 | 0.166 | 3 | 0.134 |
| Osteoarthritis | 30 | 0.342 | 27 | 0.311 | 27 | 0.152 |
| Osteoporosis | 35 | 0.398 | 33 | 0.372 | 26 | 0.291 |
| Osteosarcoma | 56 | 0.349 | 45 | 0.29 | 36 | 0.163 |
| Pancreatitis | 41 | 0.175 | 40 | 0.159 | 33 | 0.166 |
| Paresis | 1 | 0.339 | 0 | 0.305 | 0 | 0.273 |
| Periodontitis | 17 | 0.315 | 11 | 0.303 | 8 | 0.259 |
| Peritonitis | 16 | 0.346 | 15 | 0.324 | 12 | 0.2 |
| Pheochromocytoma | 14 | 0.278 | 13 | 0.275 | 9 | 0.26 |
| Photophobia | 4 | 0.244 | 2 | 0.225 | 2 | 0.204 |
| Pneumonia | 32 | 0.424 | 32 | 0.363 | 22 | 0.211 |
| Polyneuropathies | 10 | 0.36 | 8 | 0.346 | 3 | 0.303 |
| Polyuria | 5 | 0.396 | 3 | 0.387 | 2 | 0.351 |
| Pre-Eclampsia | 19 | 0.362 | 17 | 0.348 | 15 | 0.341 |
| Prolactinoma | 2 | 0.234 | 1 | 0.208 | 0 | 0.168 |
| Prostatitis | 10 | 0.477 | 9 | 0.421 | 4 | 0.285 |
| Proteinuria | 8 | -0.022 | 10 | 0.016 | 10 | 0.088 |
| Pruritus | 30 | 0.294 | 31 | 0.29 | 22 | 0.238 |

**Supplementary Table 1.** *(continued)*

| Disease | $\beta = 0$ | | $\beta = 0.2$ | | $\beta = 0.4$ | |
|---|---|---|---|---|---|---|
| | Overlap (%) | Pr-Pr Sim. | Overlap (%) | Pr-Pr Sim. | Overlap (%) | Pr-Pr Sim. |
| Psoriasis | 5 | 0.324 | 7 | 0.287 | 10 | 0.165 |
| Pyelonephritis | 1 | 0.472 | 1 | 0.366 | 1 | 0.173 |
| Rhabdomyosarcoma | 18 | 0.292 | 14 | 0.218 | 9 | 0.066 |
| Rhinitis | 30 | 0.473 | 20 | 0.421 | 8 | 0.293 |
| Sarcoidosis | 10 | 0.2 | 8 | 0.194 | 7 | 0.12 |
| Sarcoma | 16 | 0.314 | 11 | 0.265 | 8 | 0.076 |
| Schistosomiasis | 6 | 0.153 | 6 | 0.147 | 5 | 0.105 |
| Schizophrenia | 18 | 0.659 | 18 | 0.573 | 23 | 0.463 |
| Seizures | 6 | 0.608 | 10 | 0.603 | 15 | 0.482 |
| Seminoma | 13 | 0.153 | 8 | 0.107 | 5 | -0.101 |
| Sinusitis | 5 | 0.46 | 4 | 0.379 | 2 | 0.272 |
| Stomatitis | 9 | 0.064 | 7 | 0.095 | 2 | 0.079 |
| Stroke | 17 | 0.399 | 15 | 0.401 | 9 | 0.28 |
| Synovitis | 10 | 0.234 | 7 | 0.218 | 4 | 0.15 |
| Tetany | 3 | 0.162 | 3 | 0.165 | 2 | 0.126 |
| Thrombocytopenia | 14 | 0.252 | 11 | 0.256 | 5 | 0.234 |
| Thromboembolism | 8 | 0.215 | 8 | 0.183 | 5 | 0.141 |
| Thrombosis | 9 | 0.294 | 18 | 0.296 | 18 | 0.255 |
| Toothache | 2 | 0.17 | 2 | 0.155 | 0 | 0.046 |
| Torticollis | 4 | 0.462 | 3 | 0.409 | 1 | 0.263 |
| Tremor | 11 | 0.6 | 9 | 0.59 | 8 | 0.534 |
| Trichuriasis | 18 | 0.161 | 15 | 0.157 | 9 | 0.184 |
| Tuberculosis | 5 | 0.269 | 4 | 0.243 | 3 | 0.193 |
| Urticaria | 26 | 0.329 | 23 | 0.333 | 18 | 0.282 |
| Uveitis | 13 | 0.218 | 12 | 0.195 | 7 | 0.192 |
| Vitiligo | 46 | 0.167 | 39 | 0.15 | 25 | 0.095 |
| Xerostomia | 4 | 0.22 | 4 | 0.214 | 3 | 0.139 |